\documentclass[journal,onecolumn,11pt]{IEEEtran}
\usepackage{booktabs}

\usepackage{amsmath}                
\usepackage{amsthm}                 
\usepackage{amssymb}
\usepackage{thmtools}
\usepackage{kpfonts}
\usepackage[geometry]{ifsym}
\usepackage{dsfont}
\usepackage[mathscr]{euscript}
\usepackage{graphicx}
\usepackage{color}
\usepackage{paralist}
\usepackage{framed}
\usepackage{float}
\usepackage{longtable}
\usepackage{cite}
\usepackage[colorlinks=true, citecolor=magenta, linkcolor=blue]{hyperref}       
\usepackage[font=itshape]{quoting}

\usepackage{listings}
\usepackage{framed} 
\usepackage{nomencl} 
\makenomenclature

\declaretheoremstyle[%
  headfont=\bfseries,%
  headpunct={:},%
  notefont=\normalfont\bfseries,%
  notebraces={--~}{},
    qed=$\blacksquare$,
]{definitionstyle}
\theoremstyle{definition}
\declaretheorem[style=definitionstyle,name=Definition]{defn}
\declaretheorem[style=definitionstyle,name=Theorem]{thm}
\declaretheorem[style=definitionstyle,name=Proof]{prf}

\newcommand{\liinesfig}[3]{\renewcommand{\figurename}{Fig.}\begin{figure}[htb!]\begin{center}\includegraphics[width=3.25in]{#1}
\vspace{-0.1in}\caption{#2}\label{#3}\end{center}\end{figure}}
\newcommand{\liinesbigfig}[3]{\begin{figure*}[h!]\begin{center}\includegraphics[width=6.5in]{#1}
\vspace{-0.15in}\caption{#2}\label{#3}\end{center}\vspace{-0.15in}\end{figure*}}

\begin{document}

\title{A Tensor-Based Formulation of \\ Hetero-functional Graph Theory}
\markboth{Submitted for Publication:  \htmladdnormallink{(DOI)}{https://dx.doi.org/10.1016/j.renene.2017.02.063}}%
{Farid \MakeLowercase{\textit{et al.}}: Bare Demo of IEEEtran.cls for Journals}
\author{Amro~M.~Farid, Dakota~Thompson, Wester Schoonenberg
 \thanks{Amro M. Farid is an Associate Professor of Engineering with the Thayer School of Engineering at Dartmouth and a Visiting Associate Professor with the mechanical engineering department at MIT, Cambridge, MA, USA.  {\tt\small amfarid@mit.edu}}
\thanks{Dakota Thompson is with the Thayer School of Engineering at Dartmouth, Hanover, NH, USA.  {\tt\small dakota.j.thompson.th@dartmouth.edu}}
\thanks{Wester C. Schoonenberg is with the Thayer School of Engineering at Dartmouth, Hanover, NH, USA.  {\tt\small Wester.C.Schoonenberg.TH@dartmouth.edu}}
}

\maketitle

\begin{abstract}
Recently, hetero-functional graph theory (HFGT) has developed as a means to mathematically model the structure of large-scale complex flexible engineering systems.  It does so by fusing concepts from network science and model-based systems engineering (MBSE).  For the former, it utilizes multiple graph-based data structures to support a matrix-based quantitative analysis.  For the latter, HFGT inherits the heterogeneity of conceptual and ontological constructs found in model-based systems engineering including system form, system function, and system concept.  These diverse conceptual constructs indicate multi-dimensional rather than two-dimensional relationships.   This paper provides the first tensor-based treatment of hetero-functional graph theory.  In particular, it addresses the ``system concept" and the hetero-functional adjacency matrix from the perspective of tensors and introduces the hetero-functional incidence tensor as a new data structure.  The tensor-based formulation described in this work makes a stronger tie between HFGT and its ontological foundations in MBSE.  Finally, the tensor-based formulation facilitates several analytical results that provide an understanding of the relationships between HFGT and multi-layer networks.  
\end{abstract}





\section{Introduction}
One defining characteristic of twenty-first century engineering challenges is the breadth of their scope.  The National Academy of Engineering (NAE) has identified 14 ``game-changing goals"\cite{Anonymous-NAE:2019:00}.  
\begin{enumerate}
\item Advance personalized learning
\item Make solar energy economical
\item Enhance virtual reality
\item Reverse-engineer the brain
\item Engineer better medicines
\item Advance health informatics
\item Restore and improve urban infrastructure
\item Secure cyber-space
\item Provide access to clean water
\item Provide energy from fusion
\item Prevent nuclear terror
\item Manage the nitrogen cycle
\item Develop carbon sequestration methods
\item Engineer the tools of scientific discovery
\end{enumerate}
At first glance, each of these aspirational engineering goals is so large and complex in its own right that it might seem entirely intractable.  However, and quite fortunately, the developing consensus across a number of STEM (science, technology, engineering, and mathematics) fields is that each of these goals is characterized by an ``engineering system" that is analyzed and re-synthesized using a meta-problem-solving skill set\cite{Park:2021:ISC-BC10}.  
\begin{defn}
Engineering system\cite{De-Weck:2011:00}:  A class of systems characterized by a high degree of technical complexity, social intricacy, and elaborate processes, aimed at fulfilling important functions in society.
\end{defn}

\begin{table*}[t]
\caption{A Classification of Engineering Systems by Function and Operand\cite{De-Weck:2011:00}}\label{Ta:EngSysTax}
\vspace{-0.2in}
\begin{center}
\begin{tabular}{p{1.0in}p{1.0in}p{1.0in}p{1.0in}p{1.0in}p{1.0in}}\toprule
\textbf{Function/Operand} & \textbf{Living Organisms} & \textbf{Matter} & \textbf{Energy} & \textbf{Information} & \textbf{Money} \\\hline
\textbf{Transform} & Hospital & Blast Furnace & Engine, electric motor & Analytic engine, calculator & Bureau of Printing \& Engraving \\
\textbf{Transport} & Car, Airplane, Train & Truck, train, car, airplane & Electricity grid & Cables, radio, telephone, and internet & Banking Fedwire and Swift transfer systems \\
\textbf{Store} & Farm, Apartment Complex & Warehouse & Battery, flywheel, capacitor & Magnetic tape \& disk, book & U.S. Buillon Repository (Fort Knox) \\
\textbf{Exchange} & Cattle auction, (illegal) human trafficking & eBay trading system & Energy market & World Wide Web, Wikipedia & London Stock Exchange \\
\textbf{Control} & U.S. Constitution \& laws & National Highway Traffic Safety Administration & Nuclear Regulatory Commission & Internet engineering task force & United States Federal Reserve \\\bottomrule
\end{tabular}
\end{center}
\end{table*}

The challenge of \textbf{\emph{convergence}} towards \textbf{\emph{abstract}} and \textbf{\emph{consistent}} methodological foundations for engineering systems is formidable.  Consider the engineering systems taxonomy presented in Table \ref{Ta:EngSysTax}\cite{De-Weck:2011:00}.  It classifies engineering systems by five generic functions that fulfill human needs:  1.) transform 2.) transport 3.) store, 4.) exchange, and 5.) control.  On another axis, it classifies them by their operands:  1.)  living organisms (including people), 2.) matter, 3.) energy, 4.) information, 5.) money.  This classification presents a broad array of application domains that must be consistently treated.   Furthermore, these engineering systems are at various stages of development and will continue to be so for decades, if not centuries.   And so the study of engineering systems must equally support design synthesis, analysis, and re-synthesis while supporting innovation; be it incremental or disruptive.  

\subsection{Background Literature}
Two fields in particular have attempted to traverse this convergence challenge:  systems engineering and network science.   Systems engineering, and more recently model-based systems engineering (MBSE), has developed as a practical and interdisciplinary engineering discipline that enables the successful realization of complex systems from concept, through design, to full implementation\cite{SE-Handbook-Working-Group:2015:00}.  It equips the engineer with methods and tools to handle systems of ever-greater complexity arising from greater interactions within these systems or from the expanding heterogeneity they demonstrate in their structure and function.  Despite its many accomplishments, model-based systems engineering still relies on graphical modeling languages that provide limited quantitative insight (on their own)\cite{Weilkiens:2007:00,Friedenthal:2011:00,Schoonenberg:2019:ISC-BK04}. 

In contrast, network science has developed to quantitatively analyze networks that appear in a wide variety of engineering systems.  And yet, despite its methodological developments in multi-layer networks, network science has often been unable to address the explicit heterogeneity often encountered in engineering systems\cite{Schoonenberg:2019:ISC-BK04,Kivela:2014:00}.   In a recent comprehensive review Kivela et. al \cite{Kivela:2014:00} write:
\begin{quoting}
``The study of multi-layer networks $\ldots$ has become extremely popular.  Most real and engineered systems include multiple subsystems and layers of connectivity and developing a deep understanding of multi-layer systems necessitates generalizing `traditional' graph theory.  Ignoring such information can yield misleading results, so new tools need to be developed.  One can have a lot of fun studying `bigger and better' versions of the diagnostics, models and dynamical processes that we know and presumably love -- and it is very important to do so but the new `degrees of freedom' in multi-layer systems also yield new phenomena that cannot occur in single-layer systems.  Moreover, the increasing availability of empirical data for fundamentally multi-layer systems amidst the current data deluge also makes it possible to develop and validate increasingly general frameworks for the study of networks.  

$\ldots$ Numerous similar ideas have been developed in parallel, and the literature on multi-layer networks has rapidly become extremely messy.  Despite a wealth of antecedent ideas in subjects like sociology and engineering, many aspects of the theory of multi-layer networks remain immature, and the rapid onslaught of papers on various types of multilayer networks necessitates an attempt to unify the various disparate threads and to discern their similarities and differences in as precise a manner as possible.

$\ldots$ [The multi-layer network community] has produced an equally immense explosion of disparate terminology, and the lack of consensus (or even generally accepted) set of terminology and mathematical framework for studying is extremely problematic."
\end{quoting}

In many ways, the parallel developments of the model-based systems engineering and network science communities intellectually converge in \emph{hetero-functional graph theory} (HFGT)\cite{Schoonenberg:2019:ISC-BK04}.   For the former, it utilizes multiple graph-based data structures to support a matrix-based quantitative analysis.  For the latter, HFGT inherits the heterogeneity of conceptual and ontological constructs found in model-based systems engineering including system form, system function, and system concept.  More specifically, the explicit treatment of function and operand facilitates a structural understanding of the diversity of engineering systems found in Table \ref{Ta:EngSysTax}.  Although not named as such originally, the first works on HFGT appeared as early as 2006-2008\cite{Farid:2006:IEM-C02,Farid:2007:IEM-TP00,Farid:2008:IEM-J05,Farid:2008:IEM-J04}.  Since then, HFGT has become multiply established and demonstrated cross-domain applicability\cite{Farid:2015:ISC-J19,Schoonenberg:2019:ISC-BK04}; culminating in the recent consolidating text\cite{Schoonenberg:2019:ISC-BK04}.

The primary benefit of HFGT, relative to multi-layer networks, is the broad extent of its ontological elements and associated mathematical models\cite{Schoonenberg:2019:ISC-BK04}.   In their recent review, Kivela et. al showed that \emph{all} of the reviewed works have exhibited at least one of the following modeling constraints\cite{Kivela:2014:00}:
\begin{enumerate}
\item Alignment of nodes between layers is required \cite{Kivela:2014:00,De-Domenico:2013:00,De-Domenico:2014:00,Yagan:2012:00,Nicosia:2013:00,Bianconi:2013:00,Battiston:2014:00,Horvat:2012:00,Sole-Ribalta:2013:00,Cozzo:2013:00,Sola:2013:00,Pattison:1999:00,Barigozzi:2011:00,Cai:2005:00,Harrer:2012:00,Stroele:2009:00,Li:2012:00,Ng:2011:00,Brodka:2010:00,Brodka:2011:00,Brodka:2012:00,Berlingerio:2011:00,Berlingerio:2013:00,Berlingerio:2013:01,Tang:2012:00,Barrett:2012:00,Kazienko:2011:00,Coscia:2013:00,Kazienko:2011:01,Mucha:2010:00,Carchiolo:2011:00,Bassett:2013:00,Irving:2012:00,Sorrentino:2012:00,Funk:2010:00,Marceau:2011:00,Wei:2012:00,Rocklin:2013:00,Hindes:2013:00,Baxter:2012:01,Gomez-Gardenes:2012:00,Barigozzi:2010:00,Cellai:2013:00,Brummitt:2012:00,Mucha:2010:01,Wasserman:1994:00,Min:2013:00,Lee:2012:02,Min:2014:00,Cozzo:2013:01}
\item Disjointment between layers is required \cite{Kivela:2014:00,Allard:2009:00,Bashan:2013:00,Brummitt:2012:00,Buldyrev:2010:00,Cardillo:2012:00,Dickison:2012:00,Donges:2011:00,Hindes:2013:00,Lazega:2008:00,Leicht:2009:00,Louzada:2013:00,Martin-Hernandez:2013:00,Parshani:2010:00,Sahneh:2013:00,Saumell-Mendiola:2012:00,Sun:2009:00,Vazquez:2006:00,Wang:2013:00,Zhou:2007:00,Zhou:2013:00}
\item Equal number of nodes for all layers is required \cite{Kivela:2014:00,Barigozzi:2011:00,Barrett:2012:00,Bassett:2013:00,Battiston:2014:00,Berlingerio:2011:00,Berlingerio:2013:00,Berlingerio:2013:01,Bianconi:2013:00,Brodka:2010:00,Brodka:2011:00,Brodka:2012:00,Buldyrev:2010:00,Cai:2005:00,Carchiolo:2011:00,Coscia:2013:00,Cozzo:2013:00,De-Domenico:2013:00,De-Domenico:2014:00,Dickison:2012:00,Funk:2010:00,Gao:2011:00,Harrer:2012:00,Hindes:2013:00,Horvat:2012:00,Irving:2012:00,Kazienko:2011:00,Kazienko:2011:01,Lee:2012:00,Li:2012:00,Louzada:2013:00,Marceau:2011:00,Martin-Hernandez:2013:00,Min:2013:00,Min:2014:00,Mucha:2010:00,Ng:2011:00,Nicosia:2013:00,Pattison:1999:00,Rocklin:2013:00,Sola:2013:00,Sole-Ribalta:2013:00,Sorrentino:2012:00,Stroele:2009:00,Tang:2012:00,Wei:2012:00,Yagan:2012:00}
\item Exclusively vertical coupling between all layers is required \cite{Kivela:2014:00,Barigozzi:2011:00,Barrett:2012:00,Bassett:2013:00,Battiston:2014:00,Berlingerio:2011:00,Berlingerio:2013:00,Berlingerio:2013:01,Bianconi:2013:00,Brodka:2010:00,Brodka:2011:00,Brodka:2012:00,Cai:2005:00,Carchiolo:2011:00,Cardillo:2012:00,Coscia:2013:00,Cozzo:2012:00,Cozzo:2013:00,Criado:2012:00,De-Domenico:2013:00,De-Domenico:2014:00,Funk:2010:00,Harrer:2012:00,Hindes:2013:00,Horvat:2012:00,Irving:2012:00,Kazienko:2011:00,Kazienko:2011:01,Lee:2012:00,Li:2012:00,Marceau:2011:00,Min:2013:00,Min:2014:00,Mucha:2010:00,Ng:2011:00,Nicosia:2013:00,Pattison:1999:00,Rocklin:2013:00,Sola:2013:00,Sole-Ribalta:2013:00,Sorrentino:2012:00,Stroele:2009:00,Tang:2012:00,Wei:2012:00,Xu:2011:00,Yagan:2012:00,Yagan:2013:00}
\item Equal couplings between all layers are required\cite{Kivela:2014:00,Barigozzi:2011:00,Barrett:2012:00,Battiston:2014:00,Berlingerio:2011:00,Berlingerio:2013:00,Berlingerio:2013:01,Bianconi:2013:00,Brodka:2010:00,Brodka:2011:00,Brodka:2012:00,Cai:2005:00,Cardillo:2012:00,Coscia:2013:00,Cozzo:2012:00,Cozzo:2013:00,Criado:2012:00,Funk:2010:00,Harrer:2012:00,Hindes:2013:00,Horvat:2012:00,Irving:2012:00,Kazienko:2011:00,Kazienko:2011:01,Lee:2012:00,Li:2012:00,Marceau:2011:00,Min:2013:00,Min:2014:00,Ng:2011:00,Nicosia:2013:00,Pattison:1999:00,Rocklin:2013:00,Sola:2013:00,Sole-Ribalta:2013:00,Sorrentino:2012:00,Stroele:2009:00,Tang:2012:00,Wei:2012:00,Xu:2011:00,Yagan:2012:00,Yagan:2013:00}
\item Node counterparts are coupled between all layers\cite{Barigozzi:2011:00,Barrett:2012:00,Battiston:2014:00,Berlingerio:2011:00,Berlingerio:2013:00,Berlingerio:2013:01,Bianconi:2013:00,Brodka:2010:00,Brodka:2011:00,Brodka:2012:00,Cai:2005:00,Cardillo:2012:00,Coscia:2013:00,Cozzo:2012:00,Criado:2012:00,Funk:2010:00,Harrer:2012:00,Hindes:2013:00,Horvat:2012:00,Irving:2012:00,Kazienko:2011:00,Kazienko:2011:01,Lee:2012:00,Li:2012:00,Marceau:2011:00,Min:2013:00,Min:2014:00,Ng:2011:00,Nicosia:2013:00,Pattison:1999:00,Rocklin:2013:00,Sorrentino:2012:00,Stroele:2009:00,Tang:2012:00,Wei:2012:00,Xu:2011:00,Yagan:2012:00,Yagan:2013:00}
\item Limited number of modelled layers\cite{Allard:2009:00,Bashan:2013:00,Brummitt:2012:00,Buldyrev:2010:00,Cardillo:2012:00,Carley:2001:00,Carley:2007:00,Cozzo:2012:00,Criado:2012:00,Davis:2011:00,Dickison:2012:00,Donges:2011:00,Funk:2010:00,Gao:2011:00,Hindes:2013:00,Lazega:2008:00,Lee:2012:00,Leicht:2009:00,Louzada:2013:00,Marceau:2011:00,Martin-Hernandez:2013:00,Min:2013:00,Min:2014:00,Parshani:2010:00,Sahneh:2013:00,Saumell-Mendiola:2012:00,Sun:2009:00,Sun:2011:00,Sun:2012:00,Sun:2013:00,Tsvetovat:2004:00,Vazquez:2006:00,Wang:2013:00,Wei:2012:00,Xu:2011:00,Yagan:2013:00,Zhou:2007:00,Zhou:2013:00}
\item Limited number of aspects in a layer\cite{Allard:2009:00,Barigozzi:2011:00,Barrett:2012:00,Bashan:2013:00,Bassett:2013:00,Battiston:2014:00,Berlingerio:2011:00,Berlingerio:2013:00,Berlingerio:2013:01,Bianconi:2013:00,Brodka:2010:00,Brodka:2011:00,Brodka:2012:00,Brummitt:2012:00,Buldyrev:2010:00,Cai:2005:00,Carchiolo:2011:00,Cardillo:2012:00,Coscia:2013:00,Cozzo:2013:00,De-Domenico:2013:00,De-Domenico:2014:00,Dickison:2012:00,Donges:2011:00,Funk:2010:00,Harrer:2012:00,Hindes:2013:00,Horvat:2012:00,Irving:2012:00,Kazienko:2011:00,Kazienko:2011:01,Lazega:2008:00,Lee:2012:00,Leicht:2009:00,Li:2012:00,Louzada:2013:00,Marceau:2011:00,Martin-Hernandez:2013:00,Min:2013:00,Min:2014:00,Mucha:2010:00,Ng:2011:00,Nicosia:2013:00,Parshani:2010:00,Pattison:1999:00,Rocklin:2013:00,Sahneh:2013:00,Saumell-Mendiola:2012:00,Sola:2013:00,Sole-Ribalta:2013:00,Sorrentino:2012:00,Stroele:2009:00,Sun:2009:00,Tang:2012:00,Vazquez:2006:00,Wang:2013:00,Wei:2012:00,Yagan:2012:00,Zhou:2007:00,Zhou:2013:00}
\end{enumerate}
To demonstrate the consequences of these modeling limitations, the HFGT text\cite{Schoonenberg:2019:ISC-BK04} developed a very small, but highly heterogeneous, hypothetical test case system that exhibited \emph{all eight} of the modeling limitations identified by Kivela et. al.  Consequently, none of the multi-layer network models identified by Kivela et. al. would be able to model such a hypothetical test case.   In contrast, a complete HFGT analysis of this hypothetical test case was demonstrated in the aforementioned text\cite{Schoonenberg:2019:ISC-BK04}.  
\liinesbigfig{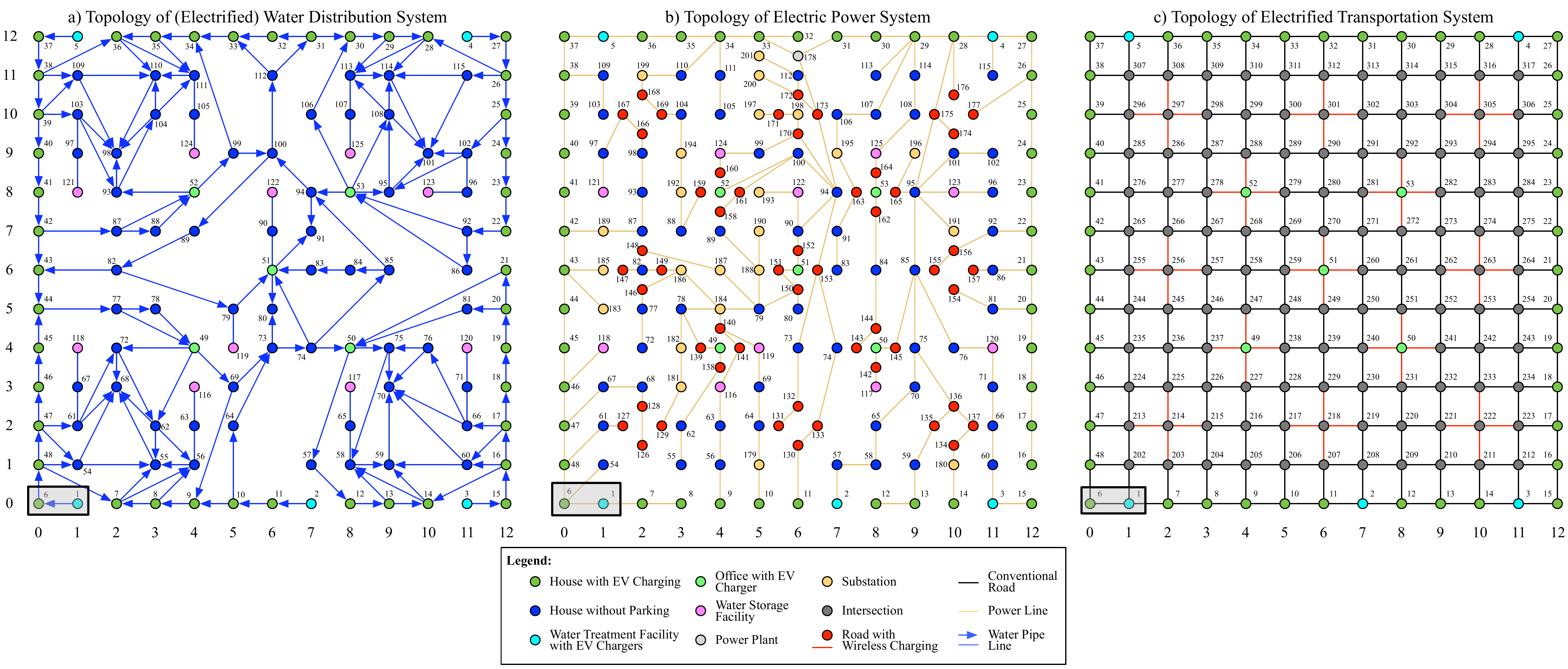}{A Topological Visualizaiton of the Trimetrica Smart City Infrastructure Test Case\cite{Schoonenberg:2019:ISC-BK04}.}{fig:Trimetrica}
The same text provides the even more complex hypothetical smart city infrastructure example  shown in Fig. \ref{fig:Trimetrica}.  It not only includes an electric power system, water distribution system, and electrified transportation system but it also makes very fine distinctions in the functionality of its component elements.

Given the quickly developing \emph{``disparate terminology and the lack of consensus"}, Kivela et. al.'s\cite{Kivela:2014:00} stated goal \emph{``to unify the various disparate threads and to discern their similarities and differences in as precise a manner as possible"} appears imperative.  While many may think that the development of mathematical models is subjective, in reality, ontological science presents a robust methodological foundation.   As briefly explained in Appendix \ref{sec:ontology}, and as detailed elsewhere\cite{Schoonenberg:2019:ISC-BK04,Farid:2016:ISC-BC06,Guizzardi:2007:00}, the process of developing a mathematical model of a given (engineering) system is never direct.  Rather, a specific engineering system (which is an instance of a class of systems) has abstract elements in the mind\footnote{It is likely that modeling abstract elements in the mind is unfamiliar to this journal's readership.  This is purely an issue of nomenclature.  Most physicists and engineers would agree on the indispensable role that \emph{intuition} -- itself a mental model -- has to the development of mathematical models of systems.  For example, the shift from Newtonian mechanics to Einstein's relativity constituted first an expansion in the abstract elements of the mental model and their relationships well before that mental model could be translated into its associated mathematics.  Similarly, the ``disparate terminology and lack of consensus" identified by Kivela et. al \cite{Kivela:2014:00} suggests that a reconciliation of this abstract mental model is required (See Section \ref{sec:ontology}).} that constitute an \emph{abstraction} $\cal A$ (which is an instance of a domain conceptualization ${\cal C}$).  ${\cal C}$ is mapped to a set of primitive mathematical elements called a language $\cal L$, which is in turn instantiated to produce a mathematical model ${\cal M}$. The fidelity of the mathematical model with respect to an abstraction is determined by the four complementary linguistic properties shown in Figure \ref{fig:Ontology}\cite{Guizzardi:2007:00}: soundness, completeness, lucidity, and laconicity \cite{Guizzardi:2005:00} (See Defns. \ref{defn:soundness}, \ref{defn:completeness}, \ref{defn:lucidity}, \ref{defn:laconicity}).   When all four properties are met, the abstraction and the mathematical model have an isomorphic (one-to-one) mapping and faithfully represent each other.  For example, the network science and graph theory literature {\color{blue}assume} an abstract conceptualization of nodes and edges prior to defining their 1-to-1 mathematical counterparts\cite{Schoonenberg:2019:ISC-BK04}.  Consequently, as hetero-functional graph and multi-layer network models of engineering systems are developed, there is a need to reconcile both the abstraction and the mathematical model on the basis of the four criteria identified above (See Appendix \ref{sec:ontology}.).

\liinesfig{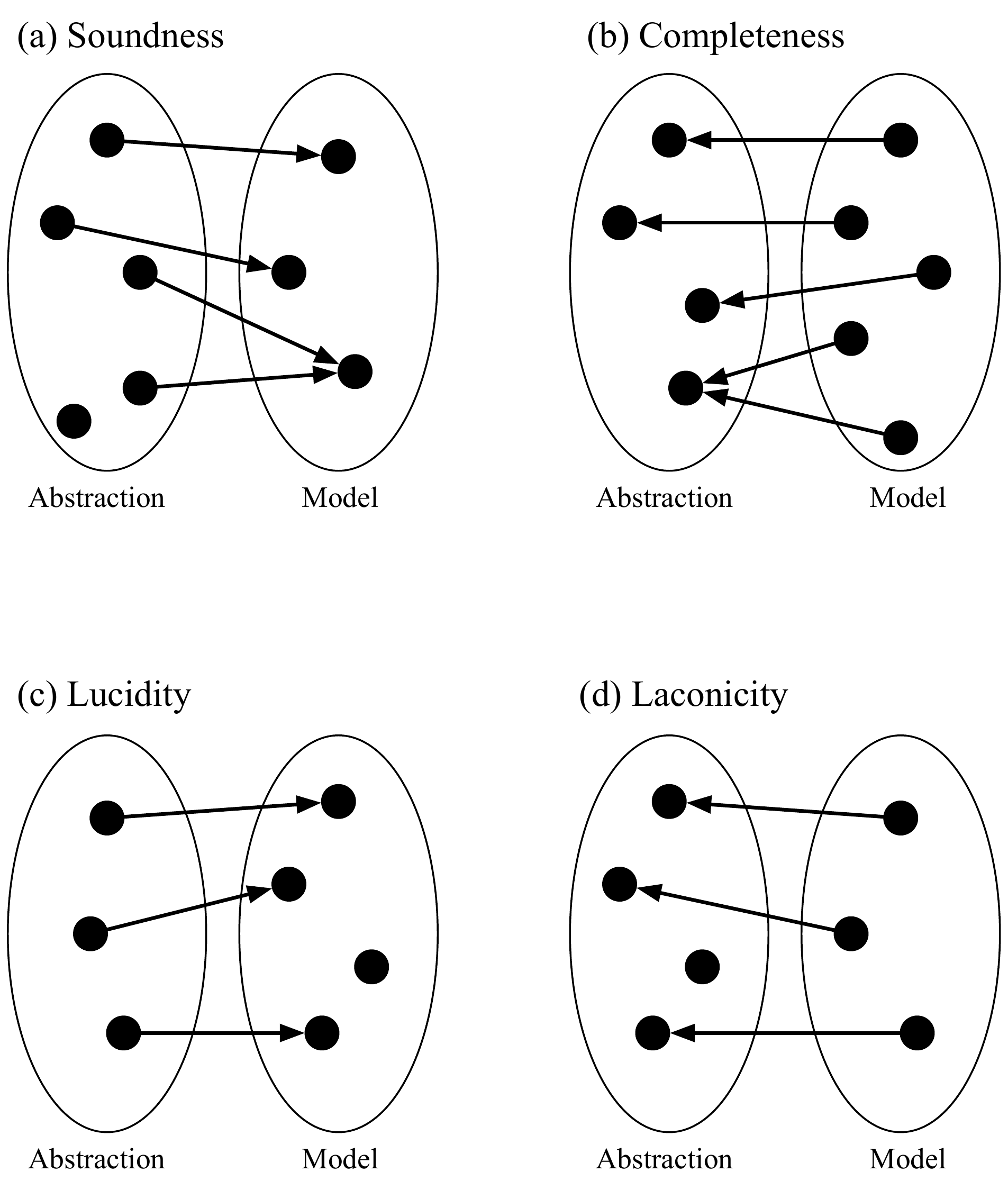}{Graphical Representation of Four Ontological Properties As Mapping Between Abstraction and Model:  \textbf{a} Soundness, \textbf{b} Completeness, \textbf{c} Lucidity, and \textbf{d} Laconicity \cite{Guizzardi:2007:00}.}{fig:Ontology}

The ontological strength of hetero-functional graph theory comes from the ``systems thinking" foundations in the model-based systems engineering literature\cite{Crawley:2015:00,Schoonenberg:2019:ISC-BK04}.   In effect, and very briefly, all systems have a ``subject + verb + operand" form where the system form is the subject, the system function is the verb + operand (i.e. predicate) and the system concept is the mapping of the two to each other.   The key distinguishing feature of HFGT (relative to multi-layer networks) is its introduction of system function.  In that regard, it is more complete than multi-layer networks if system function is accepted as part of an engineering system abstraction.  Another key distinguishing feature of HFGT is the differentiation between elements related to transformation and transportation.   In that regard, it takes great care to not \emph{overload} mathematical modeling elements and preserve lucidity.  

\subsection{Original Contribution}\label{sec:contribution}
This paper provides a tensor-based formulation of several of the most important parts of hetero-functional graph theory.  More specifically, it discusses the system concept, the hetero-functional adjacency matrix, and the hetero-functional incidence tensor.  Whereas the hetero-functional graph theory text\cite{Schoonenberg:2019:ISC-BK04} is a comprehensive discussion of the subject, the treatment is based entirely on two-dimensional matrices.   The tensor-based formulation described in this work makes a stronger tie between HFGT and its ontological foundations in MBSE.  Furthermore, the tensor-based treatment developed here reveals patterns of underlying structure in engineering systems that are less apparent in a matrix-based treatment.  Finally, the tensor-based formulation facilitates an understanding of the relationships between HFGT and multi-layer networks (``despite its disparate terminology and lack of consensus").  In so doing, this tensor-based treatment is likely to advance Kivela et. al's goal to discern the similarities and differences between these mathematical models in as precise a manner as possible.  

\subsection{Paper Outline}
The rest of the paper is organized as follows.  Section \ref{sec:As} discusses the system concept as an allocation of system function to system form.   Section \ref{sec:Arho} discusses the hetero-functional adjacency matrix emphasizing the relationships between system capabilities (i.e. structural degrees of freedom as defined therein).    Section \ref{sec:Mrho}, then, discusses the hetero-functional incidence tensor which describes the relationships between system capabilities, operands, and physical locations in space (i.e. system buffers as defined later).  Section \ref{sec:discussion} goes on to discuss this tensor-based formulation from the perspective of layers and network descriptors.  Section \ref{sec:conclusion} brings the work to a close and offers directions for future work.  {\color{blue}Given the multi-disciplinary nature of this work, several appendices are provided to support the work with background material and avoid breaking the logical flow of the main article.  Appendix \ref{sec:ontology} provides the fundamental definitions of ontological science that were used to motivate this work's original contribution.  Appendix \ref{app:notations} describes the notation conventions used throughout this work.  The paper assumes that the reader is well grounded in graph theory and network science as it is found in any one of a number of excellent texts\cite{Barabasi:2016:00,Newman:2009:00}.  The paper does not assume prior exposure to hetero-functional graph theory.   It's most critical definitions are tersely introduced in the body of the work upon first mention.   More detailed classifications of these concepts are compiled in Appendix \ref{sec:hfgt_dfn} for convenience.  Given the theoretical treatment provided here, the interested reader is referred to the hetero-functional graph theory text\cite{Schoonenberg:2019:ISC-BK04} for further explanation of these well-established concepts and concrete examples.  Furthermore, several recent works have made illustrative comparisons between (formal) graphs and hetero-functional graphs\cite{Thompson:2020:SPG-C68,Thompson:2021:SPG-J46}.  Finally, this work makes extensive use of set, Boolean, matrix, and tensor operations; all of which are defined unambiguously in Appendices \ref{sec:setDefn}, \ref{sec:boolDefn}, \ref{sec:matDefn}, and \ref{sec:tensorDefn} respectively.}

\section{The System Concept}\label{sec:As}
At a high-level, the system concept $A_S$ describes the allocation of system function to system form as the central question of engineering design.  {\color{blue}First, Subsection \ref{sec:JS} provides introductory definitions of system resources, processes, and knowledge base that serve as prerequisite knowledge for the remaining subsections.   Next, Subsection \ref{sec:JH} introduces the transportation knowledge base as a third-order tensor.  Next, Subsection \ref{sec:JHbar} introduces the refined transportation knowledge base as a fourth-order tensor.  The tensor-based formulations in these two subsection directly support the original contribution mentioned in Section \ref{sec:contribution}.  Finally, in order to support the following section, Subsection \ref{sec:AS} concludes the section with a discussion of existence and availability of system capabilities as part of the system concept.}  

{\color{blue}\subsection{System Resources, Processes, and Knowledge Base}\label{sec:JS}}
This dichotomy of form and function is repeatedly emphasized in the fields of engineering design and systems engineering\cite{Crawley:2015:00,Buede:2009:00,Kossiakoff:2003:00,Farid:2016:ISC-BK03}.  More specifically, the allocation of system processes to system resources is captured in the ``design equation"\cite{Farid:2007:IEM-TP00,Farid:2016:ISC-BC06}: 
\begin{equation}\label{Eq:designequation}
P=J_S\odot R
\end{equation}

\noindent where $R$ is set of system resources, $P$ is the set of system processes, $J_S$ is the system knowledge base, and $\odot$ is matrix Boolean multiplication (Defn. \ref{defn:boolMult}).  

\begin{defn}[System Resource]\cite{SE-Handbook-Working-Group:2015:00}
An asset or object $r_v \in R$ that is utilized during the execution of a process.  
\end{defn}

\begin{defn}[System Process\cite{Hoyle:1998:00,SE-Handbook-Working-Group:2015:00}]\label{def:CH4:process}
An activity $p \in P$ that transforms a predefined set of input operands into a predefined set of outputs. 
\end{defn}

\begin{defn}[System Operand]\cite{SE-Handbook-Working-Group:2015:00}
An asset or object $l_i \in L$ that is operated on or consumed during the execution of a process.  
\end{defn}

\begin{defn}[System Knowledge Base\cite{Farid:2006:IEM-C02,Farid:2007:IEM-TP00,Farid:2008:IEM-J05,Farid:2008:IEM-J04,Farid:2015:ISC-J19,Farid:2016:ISC-BC06}]\label{def:systemKB}
A binary matrix $J_S$ of size $\sigma(P)\times\sigma(R)$ whose element $J_S(w,v)\in\{0,1\}$ is equal to one when action $e_{wv} \in {\cal E}_S$ (in the SysML sense) exists as a system process $p_w \in P$ being executed by a resource $r_v \in R$.  The $\sigma()$ notation gives the size of a set.   
\end{defn} 
\noindent In other words, the system knowledge base forms a bipartite graph between the set of system processes and the set of system resources\cite{Farid:2015:ISC-J19}.  

Hetero-functional graph theory further recognizes that there are inherent differences within the set of resources as well as within the set of processes.   {\color{blue}Therefore, classifications of these sets of resources and sets of processes are introduced and defined in Appendix \ref{sec:hfgt_dfn}.}  $R=M \cup B \cup H$ where $M$ is the set of transformation resources (Defn. \ref{defn:M}), $B$ is the set of independent buffers (Defn. \ref{defn:B}), and $H$ is the set of transportation resources (Defn. \ref{defn:H}).   Furthermore, the set of buffers $B_S=M \cup B$ (Defn. \ref{defn:BS}) is introduced for later discussion.  Similarly, $P = P_\mu \cup P_{\bar{\eta}}$ where $P_\mu$ is the set of transformation processes (Defn. \ref{defn:Pmu}) and $P_{\bar{\eta}}$ is the set of refined transportation processes (Defn. \ref{defn:PHref}).  The latter, in turn, is determined from the Cartesian product (\Cross) (Defn. \ref{defn:CartesProd}) of the set of transportation processes $P_\eta$ (Defn. \ref{defn:PH}) and the set of holding processes $P_\gamma$ (Defn. \ref{defn:Pgamma}).  
\begin{equation}\label{eq:CH4:refinedtransportationprocesses}
    P_{\bar{\eta}} = P_{\gamma} \mbox{\Cross} P_{\eta}
\end{equation}

\liinesbigfig{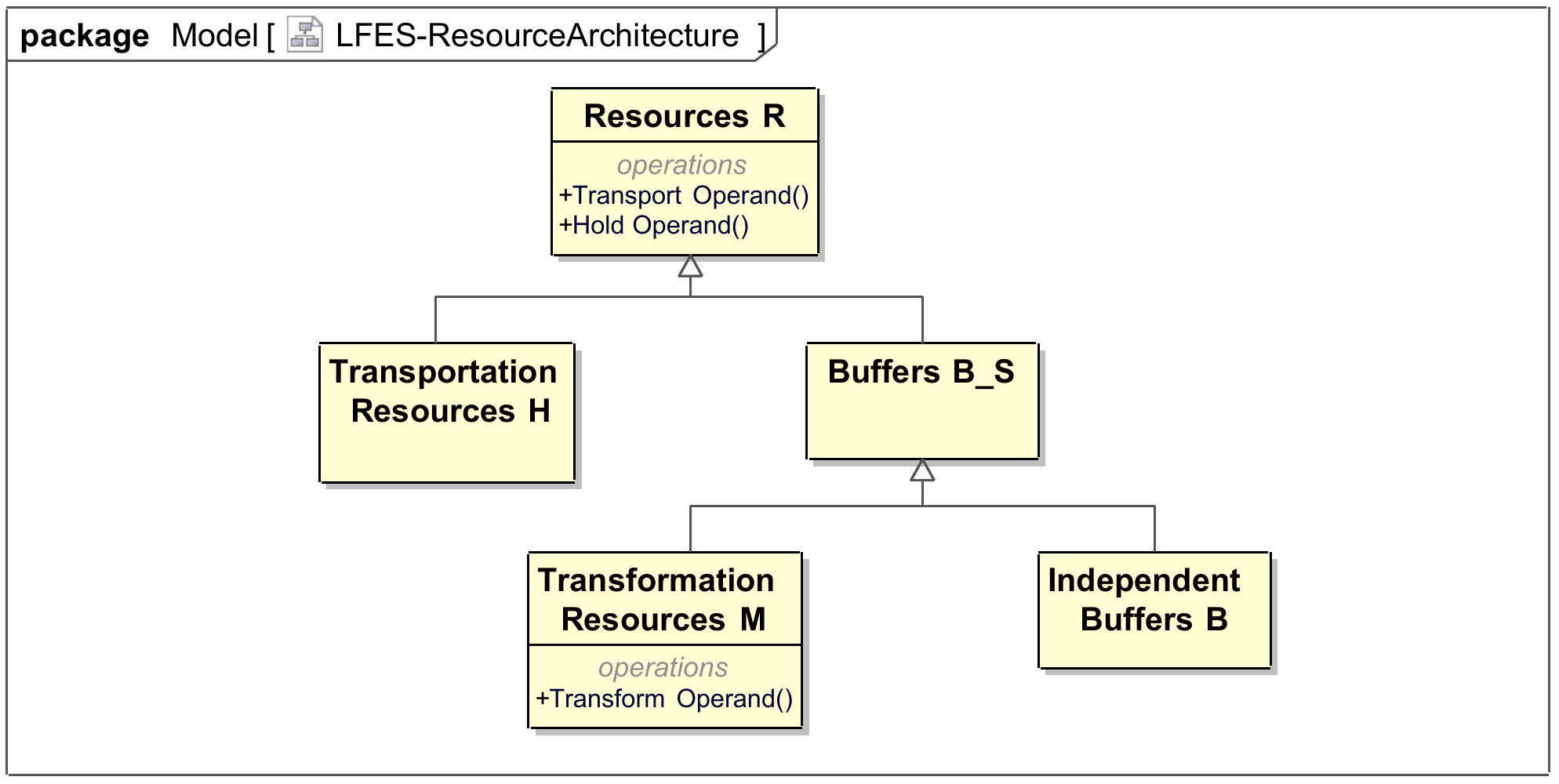}{The Hetero-functional Graph Theory Meta-Architecture drawn using the Systems Markup Language (SysML).  It consists of three types of resources $R = M \cup B \cup H$ that are capable of two types of process $P_{\bar{\eta}} = P_{\gamma} \mbox{\Cross} P_{\eta}$\cite{Schoonenberg:2019:ISC-BK04}.}{fig:HFGT-Meta}

This taxonomy of resources, processes, and their allocation is organized in the HFGT meta-architecture shown in Figure \ref{fig:HFGT-Meta}.  The taxonomy of resources $R$ and processes $P$  originates from the field of production systems where transformation processes are viewed as ``value-adding", holding processes support the design of fixtures, and transportation processes are cost-minimized.  Furthermore, their existence is necessitated by their distinct roles in the structural relationships found in hetero-functional graphs.  Consequently, subsets of the design equation \ref{Eq:designequation} can be written to emphasize the relationships between the constitutent classes of processes and resources\cite{Farid:2006:IEM-C02,Farid:2007:IEM-TP00,Farid:2008:IEM-J05,Farid:2008:IEM-J04,Farid:2015:ISC-J19}.  
\begin{align} \label{eq:designequationM}
    P_\mu &= J_M\odot M \\ \label{eq:designequationG}
    P_\gamma &= J_\gamma \odot R \\ \label{eq:designequationH}
    P_\eta &= J_H\odot R \\ \label{eq:designequationHG}
    P_{\bar{\eta}} &= J_{\bar{H}} \odot R 
\end{align}
where $J_M$ is the transformation knowledge base,  $J_\gamma$ is the holding knowledge base, $J_H$ is the transportation knowledge base, and $J_{\bar{H}}$ is the refined transportation knowledge base \cite{Farid:2007:IEM-TP00,Farid:2013:IEM-C30,Farid:2015:ISC-J19,Farid:2015:IEM-J23,Farid:2016:ETS-J27,Farid:2016:ETS-BC05}.  {\color{blue}The original system knowledge base $J_S$ is straightforwardly reconstructed from these smaller knowledge bases\cite{Farid:2006:IEM-C02,Farid:2007:IEM-TP00,Farid:2008:IEM-J05,Farid:2008:IEM-J04,Farid:2015:ISC-J19}:  
\begin{equation}\label{eq:JS}
J_{S}=
\left[\begin{array}{ccc}
J_{M}&|&\mathbf{0}\\\hline
&J_{\bar{H}}&\\
\end{array}\right]
\end{equation}}

{\color{blue}\subsection{The Transportation Knowledge Base Tensor}\label{sec:JH}}
The transportation knowledge base $J_H$ is best understood as a \emph{matricized} 3$^{rd}$-order tensor ${\cal J}_H$ where the element ${\cal J}_H(y_1,y_2,v)=1$ when the transportation process $p_u \in P_\eta $ defined by the origin $b_{s_{y_1}} \in B_{S}$ and the destination $b_{s_{y_2}} \in B_{S} $ is executed by the resource $r_v \in R$.    
\begin{align}
J_H&={\cal F}_M\left({\cal J}_H,[2,1],[3]\right) \\
{\cal J}_H&={\cal F}_M^{-1}\left(J_H,[\sigma(B_S),\sigma(B_S),\sigma(R)],[2,1],[3]\right)
\end{align}
where ${\cal F}_M$ and ${\cal F}_M^{-1}$ are the matricization and tensorization functions (Defns. \ref{defn:matricization} and \ref{defn:tensorization}) respectively.  Here, ${\cal F}_M()$ serves to vectorize the dimensions of the origin and destination buffers into the single dimension of transportation processes.  {\color{blue}Appendix \ref{sec:tensorDefn}, more generally, introduces the reader to tensor-based operations.}  

The ${\cal J}_H$ tensor reveals that the transportation knowledge base is closely tied to the classical understanding of a graph $A_{B_S}$ where point elements of form called nodes, herein taken to be the set of buffers $B_S$, are connected by line elements of form called edges.  Such a graph in hetero-functional graph theory (and model-based systems engineering) is called a formal graph\cite{Crawley:2015:00} because all of its elements describe the system form and any statement of function is entirely \emph{implicit}.  
\begin{alignat}{3}
A_{B_S}(y_1,y_2) &=\bigvee_v^{\sigma(R)}{\cal J}_H(y_1,y_2,v) &&= \bigvee_v^{\sigma(R)}J_H(u,v) \qquad \forall y_1,y_2 \in \{1,\ldots, \sigma(B_S)\} , u = \sigma(B_S)(y_1 -1) + y_2, v \in \{1,\ldots, \sigma(R)\}   \\
A_{B_S} &= {\cal J}_H \odot_3 \mathds{1}^{\sigma(R)} &&=vec^{-1}\left(J_H\odot\mathds{1}^{\sigma(R)},[\sigma(B_S),\sigma(B_S)]\right)^T\\
A_{B_S}^{TV} &= \left({\cal J}_H \odot_3 \mathds{1}^{\sigma(R)}\right)^{TV} &&=J_H\odot\mathds{1}^{\sigma(R)}
\end{alignat}
where the $\bigvee$ notation is the Boolean analogue of the $\sum$ notation (Defn. \ref{defn:bigvee}), $\odot _n$ is the n-mode Boolean matrix product (Defn. \ref{defn:nBoolModeProduct}), $vec^{-1}()$ is inverse vectorization (Defn. \ref{defn:vec-1}) and $()^V$ is shorthand for vectorization (Defn. \ref{defn:vec}).  {\color{blue}Appendix \ref{sec:boolDefn}, more generally, introduces the reader to Boolean operations.}  Furthermore, the notation $\mathds{1}^n$ is used to indicate a ones-vector of length n.  The transportation system knowledge base $J_H$ replaces the edges of the formal graph $A_{B_S}$ with an \emph{explicit} description of function in the transportation processes $P_\eta$.  The multi-column nature of the transportation knowledge base $J_H$ contains more information than the formal graph $A_{B_S}$ and allows potentially many resources to execute any given transportation process.  Consequently, the OR operation across the rows of $J_H$ (or the third dimension of ${\cal J}_H$) is sufficient to reconstruct the formal graph $A_{B_S}$.  In short, a single column transportation knowledge base is mathematically equivalent to a vectorized formal graph $A_{B_S}$.  

{\color{blue}\subsection{The Refined Transportation Knowledge Base Tensor}\label{sec:JHbar}}
Similarly, the refined transportation knowledge base is best understood as a matricized 4$^{th}$ order tensor  ${\cal J}_{\bar{H}}$ where the element ${\cal J}_H(g,y_1,y_2,v)=1$ when the refined transportation process $p_\varphi \in P_{\bar{\eta}} $ defined by the holding process $p_{\gamma g} \in P_\gamma$, the origin $b_{s_{y_1}} \in B_{S}$ and the destination $b_{s_{y_2}} \in B_{S}$ is executed by the resource $r_v \in R$.    
\begin{align}
J_{\bar{H}}&={\cal F}_M\left({\cal J}_{\bar{H}},[3,2,1],[4]\right) \\
{\cal J}_{\bar{H}}&={\cal F}_M^{-1}\left(J_{\bar{H}},[\sigma(P_\gamma),\sigma(B_S),\sigma(B_S),\sigma(R)],[3,2,1],[4]\right)
\end{align}

The $J_{\bar{H}}$ tensor reveals that the refined transportation knowledge base is closely tied to the classical understanding of a multi-commodity flow network ${\cal A}_{L B_S}$\cite{Hu:1963:00,Okamura:1983:00,Ahuja:1988:00}.  Mathematically, it is a 3$^{rd}$-order tensor whose element ${\cal A}_{L B_S}(i,y_1,y_2)=1$ when operand $l_i \in L$ is transported from buffer $b_{s_{y_1}}$ to $b_{s_{y_2}}$.  Again, the multi-commodity flow network ${\cal A}_{L B_S}$ is purely a description of system form and any statement of function is entirely \emph{implicit}.  In the special case\footnote{By Defn. \ref{defn:Pgamma}, holding processes are distinguished by three criteria: 1.) different operands, 2.) how they hold those operands, and 3.) if they change the state of the operand.  The special case mentioned above is restricted to only the first of these three conditions.} of a system where the set of operands $L$ maps 1-to-1 the set of holding processes $P_\gamma$ (i.e. $i=g$):    
\begin{alignat}{3}
{\cal A}_{L B_S}(i,y_1,y_2)&=\bigvee_v^{\sigma(R)}{\cal J}_{\bar{H}}(g,y_1,y_2,v) && \quad \forall g \in \{1,\ldots, \sigma(P_\gamma)\} ,  y_1,y_2 \in \{1,\ldots, \sigma(B_S)\} ,  v \in \{1,\ldots, \sigma(R)\}\\
&= \bigvee_v^{\sigma(R)}J_{\bar{H}}(\varphi,v) && \quad \forall \varphi=\sigma^2(B_S)(g-1)+\sigma(B_S)(y_1-1)+y_2, v \in \{1,\ldots, \sigma(R)\} \\
{\cal A}_{L B_S}&= {\cal J}_{\bar{H}} \odot_4 \mathds{1}^{\sigma(R)} &&=vec^{-1}\left(J_{\bar{H}}\odot\mathds{1}^{\sigma(R)},\left[\sigma(B_S),\sigma(B_S),\sigma(P_\gamma)\right]\right)^T\\
{\cal A}_{L B_S}^{TV}&= \left({\cal J}_{\bar{H}} \odot_4 \mathds{1}^{\sigma(R)}\right)^{TV} &&=J_{\bar{H}}\odot\mathds{1}^{\sigma(R)}
\end{alignat}
The refined transportation system knowledge base $J_{\bar{H}}$ replaces the operands and edges of the multi-commodity flow network $A_{MP}$ with an \emph{explicit} description of function in the holding processes $P_\gamma$ and transportation processes $P_\eta$.  The multi-column nature of the refined transportation knowledge base $J_{\bar{H}}$  contains more information than the multi-commodity flow network ${\cal A}_{L B_S}$ and allows potentially many resources to execute any given refined transportation process.  Consequently, the OR operation across the rows of $J_{\bar{H}}$ (or the fourth dimension of ${\cal J}_{\bar{H}}$)  is sufficient to reconstruct the multi-commodity flow network ${\cal A}_{L B_S}$.  In short, a single column of the refined transportation knowledge base is mathematically equivalent to a vectorized multi-commodity flow network.  

The transformation, holding, transportation and refined transportation knowledge bases ($J_M$, $J_\gamma$, $J_H$ and $J_{\bar{H}}$) readily serve to reconstruct the system knowledge base $J_S$.  First, the refined transportation knowledge base is the Khatri-Rao product of the holding and transportation knowledge bases.  
\begin{align}\label{eq:JHbar}
J_{\bar{H}}(\sigma(P_\eta)(g-1)+u,v) &=J_\gamma(g,v) \cdot J_H(u,v) \qquad \forall g \in \{1,\ldots, \sigma(P_\gamma)\} ,  u = \sigma(B_S)(y_1 -1) + y_2,  v \in \{1,\ldots, \sigma(R)\}\\
J_{\bar{H}} &=J_\gamma \circledast J_H\\
&= \left[J_\gamma \otimes \mathds{1}^{\sigma(P_\eta)}\right] \cdot \left[\mathds{1}^{\sigma(P_\gamma)} \otimes J_H \right]
\end{align}
where $\cdot$ is the Hadamard (or scalar) product (Defn. \ref{defn:hadamard}), $\circledast$ is the Khatri-Rao product (Defn. \ref{defn:khatri-rao}) and $\otimes$ is the Kronecker product (Defn. \ref{defn:kron}).

{\color{blue}\subsection{Existence, Availability, and Concept of System Capabilities}\label{sec:AS}}
Hetero-functional graph theory also differentiates between the \emph{existence} and the \emph{availability} of physical capabilities in the system\cite{Farid:2007:IEM-TP00,Farid:2015:IEM-J23}.  While the former is described by the system knowledge base the latter is captured by the system constraints matrix (which is assumed to evolve in time).  

\begin{defn}[System Constraints Matrix\cite{Farid:2006:IEM-C02,Farid:2007:IEM-TP00,Farid:2008:IEM-J05,Farid:2008:IEM-J04,Farid:2015:ISC-J19,Farid:2016:ISC-BC06}]\label{def:systemKS}  A binary matrix $K_S$ of size $\sigma(P)\times\sigma(R)$ whose element $K_S(w,v)\in\{0,1\}$ is equal to one when a constraint eliminates event $e_{wv}$ from the event set.  
\end{defn}
\noindent The system constraints matrix is constructed analogously to the system knowledge base\cite{Farid:2006:IEM-C02,Farid:2007:IEM-TP00,Farid:2008:IEM-J05,Farid:2008:IEM-J04,Farid:2015:ISC-J19}.  
\begin{equation}\label{eq:KS}
K_{S}=
\left[\begin{array}{ccc}
K_{M}&|&\mathbf{0}\\\hline
&K_{\bar{H}}&\\
\end{array}\right]
\end{equation}
\noindent In this regard, the system constraints matrix has a similar meaning to graph percolation \cite{Callaway:2000:00,Newman:2003:00} and temporal networks \cite{Holme:2012:00}.  

Once the system knowledge base $J_S$ and the system constraints matrix $K_S$ have been constructed, the system concept $A_S$ follows straightforwardly.  
\begin{defn}[System Concept\cite{Farid:2006:IEM-C02,Farid:2007:IEM-TP00,Farid:2008:IEM-J05,Farid:2008:IEM-J04,Farid:2015:ISC-J19,Farid:2016:ISC-BC06}]\label{def:systemConcept}
A binary matrix $A_S$ of size $\sigma(P)\times\sigma(R)$ whose element $A_S(w,v)\in\{0,1\}$ is equal to one when action $e_{wv} \in {\cal E}_S$ (in the SysML sense) is available as a system process $p_w \in P$ being executed by a resource $r_v \in R$.
\begin{equation}
A_S=J_S\ominus K_S=J_S\cdot \bar{K}_S
\end{equation}
where $\ominus$ is Boolean subtraction (Defn. \ref{defn:boolSub}) and $\overline{K}_S=NOT(K_S)$.  
\end{defn}

Every filled element of the system concept indicates a \emph{system capability} (Defn. \ref{defn:capability}) of the form:  ``Resource $r_v$ does process $p_w$".   The system constraints matrix limits the availability of capabilities in the system knowledge base to create the system concept  $A_S$.  The system capabilities are quantified by the structural degrees of freedom.

\begin{defn}[Structural Degrees of Freedom\cite{Farid:2006:IEM-C02,Farid:2007:IEM-TP00,Farid:2008:IEM-J05,Farid:2008:IEM-J04,Farid:2015:ISC-J19,Farid:2016:ISC-BC06}]\label{def:DOFS}
The set of independent actions ${\cal E}_S$ that completely defines the instantiated processes in a large flexible engineering system.  Their number is given by:
\begin{align}\label{Eq:DOFS1}
DOF_S=\sigma({\cal E}_S)&=\sum_w^{\sigma(P)}\sum_v^{\sigma(R)}\left[ J_S\ominus K_S \right](w,v)\\
&=\sum_w^{\sigma(P)}\sum_v^{\sigma(R)}A_S(w,v) \label{Eq:DOFS2} \\
&= \langle J_S, \bar{K}_S \rangle_F 
\end{align}
\end{defn}

\noindent  As has been discussed extensively in prior publications, the term structural degrees of freedom is best viewed as a generalization of kinematic degrees of freedom (or generalized coordinates)\cite{Farid:2006:IEM-C02,Farid:2007:IEM-TP00,Farid:2008:IEM-J05,Farid:2008:IEM-J04,Farid:2015:ISC-J19,Farid:2014:ISC-C37}.  Note that the transformation degrees of freedom $DOF_M$ and the refined transportation degrees of freedom $DOF_H$ are calculated similarly\cite{Farid:2006:IEM-C02,Farid:2007:IEM-TP00,Farid:2008:IEM-J05,Farid:2008:IEM-J04}:
\begin{align}\label{eq:DOFM:ch4}
    DOF_M &= \sum_j^{\sigma(P_\mu)}\sum_k^{\sigma(M)}\left[ J_M\ominus K_M \right](j,k)\\ \label{eq:DOFH:ch4}
    DOF_H &= \sum_\varphi^{\sigma(P_{\bar{\eta}})}\sum_v^{\sigma(R)}\left[ J_{\bar{H}}\ominus K_{\bar{H}} \right](u,v)
\end{align}

\section{Hetero-functional Adjacency Matrix}\label{sec:Arho}
{\color{blue}This section serves provides a tensor-based formulation of the hetero-functional adjacency matrix.  First, Subsection \ref{sec:Arho2} introduces this matrix as pairwise sequences of system capabilities.  Next, Subsection \ref{sec:Jrho} provides a tensor-based formulation of the system sequence knowledge base.  Next, Subsection \ref{sec:Krho} provides a tensor-based formulation of the system sequence constraints.  Both of these subsections directly support the paper's original contribution.  Finally, Subsection \ref{sec:DOFrho} concludes the section with a discussion of sequence-dependent degrees of freedom.  

\subsection{Pairwise Sequences of System Capabilities}\label{sec:Arho2}}
Once the system's physical capabilities (or structural degrees of freedom have been defined), the hetero-functional adjacency matrix $A_\rho$ is introduced to represent their pairwise sequences \cite{Farid:2015:ISC-J19,Farid:2015:SPG-J17,Viswanath:2013:ETS-J08,Farid:2016:ETS-J27,Schoonenberg:2017:IEM-J34}.

\begin{defn}[Hetero-functional Adjacency Matrix\cite{Farid:2015:ISC-J19,Farid:2015:SPG-J17,Viswanath:2013:ETS-J08,Farid:2016:ETS-J27,Schoonenberg:2017:IEM-J34}]\label{def:HFAM}
A square binary matrix $A_\rho$ of size $\sigma(R)\sigma(P)\times\sigma(R)\sigma(P)$ whose element $J_\rho(\chi_1,\chi_2)\in \{0,1\}$ is equal to one when string $z_{\chi_1,\chi_2}=e_{w_1v_1}e_{w_2v_2} \in {\cal Z}$ is available and exists, where index $\chi_i \in \left[1, \dots ,\sigma(R)\sigma(P)\right]$.
\end{defn}
\noindent In other words, the hetero-functional adjacency matrix corresponds to a hetero-functional graph $G = \{{\cal E}_S, {\cal Z} \}$ with structural degrees of freedom (i.e. capabilities) ${\cal E}_S$ as nodes and feasible sequences   ${\cal Z}$ as edges.  

Much like the system concept $A_S$, the hetero-functional adjacency matrix $A_\rho$ arises from a Boolean difference\cite{Farid:2015:ISC-J19,Farid:2015:SPG-J17,Viswanath:2013:ETS-J08,Farid:2016:ETS-J27,Schoonenberg:2017:IEM-J34}.   
\begin{equation}\label{eq:Arho}
	A_{\rho} = J_{\rho} \ominus K_\rho
\end{equation}
where $J_\rho$ is the system sequence knowledge base and $K_\rho$ is the system sequence constraints matrix.  
\begin{defn}[System Sequence Knowledge Base \cite{Farid:2015:ISC-J19,Farid:2015:SPG-J17,Viswanath:2013:ETS-J08,Farid:2016:ETS-J27,Schoonenberg:2017:IEM-J34}]\label{def:SSKB}
A square binary matrix $J_\rho$ of size $\sigma(R)\sigma(P)\times\sigma(R)\sigma(P)$ whose element $J_\rho(\chi_1,\chi_2)\in \{0,1\}$ is equal to one when string $z_{\chi_1,\chi_2}=e_{w_1v_1}e_{w_2v_2} \in {\cal Z}$ exists, where index $\chi_i \in \left[1, \dots ,\sigma(R)\sigma(P)\right]$.
\end{defn}

\begin{defn}[System Sequence Constraints Matrix\cite{Farid:2015:ISC-J19,Farid:2015:SPG-J17,Viswanath:2013:ETS-J08,Farid:2016:ETS-J27,Schoonenberg:2017:IEM-J34}]\label{def:SSCM}  A square binary constraints matrix $K_{\rho}$ of size $\sigma(R)\sigma(P)\times\sigma(R)\sigma(P)$ whose elements $K(\chi_1,\chi_2)\in \{0,1\}$ are equal to one when string $z_{\chi_1 \chi_2}=e_{w_1v_1}e_{w_2v_2} \in {\cal Z}$ is eliminated. 
\end{defn}

The definitions of the system sequence knowledge base $J_\rho$ and the system sequence constraints matrix $K_\rho$ feature a translation of indices from $e_{w_1v_1}e_{w_2v_2}$ to $z_{\chi_1 \chi_2}$.  This fact suggests that these matrices have their associated 4$^{th}$ order tensors  ${\cal J}_\rho$, ${\cal K}_\rho$ and ${\cal A}_\rho$.  
\begin{align}
J_\rho&={\cal F}_M\left({\cal J}_\rho,[1,2],[3,4]\right)\\
K_\rho&={\cal F}_M\left({\cal K}_\rho,[1,2],[3,4]\right)\\
A_\rho&={\cal F}_M\left({\cal A}_\rho,[1,2],[3,4]\right)
\end{align}

{\color{blue}\subsection{The System Sequence Knowledge Base Tensor}\label{sec:Jrho}
The system sequence knowledge base} $J_\rho$ and its tensor-equivalent ${\cal J}_\rho$ create all the \emph{potential} sequences of the capabilities in $A_S$.  
\begin{align}\label{Eq:RheoProdKB}
{\cal J}_\rho(w_1,v_1,w_2,v_2)&=A_S(w_1,v_1)\cdot A_S(w_2,v_2) \quad \forall w_1,w_2 \in \{1\ldots \sigma(P)\} , v_1,v_2 \in \{1\ldots \sigma(R)\} \\
{J}_\rho(\chi_1,\chi_2)&=A_S^V(\chi_1)\cdot A^V_S(\chi_2) \qquad\quad\;\;  \forall \chi_1,\chi_2 \in \{1\ldots \sigma(R)\sigma(P)\}\\
J_\rho&=A_S^VA_S^{VT}\\
J_\rho&=\left[J_S\cdot\bar{K}_S\right]^V\left[J_S\cdot\bar{K}_S\right]^{VT}\\
{\cal J}_\rho&={\cal F}_M^{-1}\left(J_\rho,[\sigma(P),\sigma(R),\sigma(P),\sigma(R)],[1,2],[3,4]\right)
\end{align}

{\color{blue}\subsection{The System Sequence Constraints Tensor}\label{sec:Krho}}
Of these potential sequences of capabilities, the system sequence constraints matrix $K_\rho$ serves to eliminate the infeasible pairs. The feasibility arises from five types of constraints:
\begin{enumerate}[I:]
\item $P_\mu P_\mu$.  Two transformation processes that follow each other must occur at the same transformation resource.  $m_1=m_2$.  
\item $P_\mu P_{\bar{\eta}}$.  A refined transportation process that follows a transformation process must have an origin equivalent to the transformation resource at which the transformation process was executed.  $m_1-1=(u_1-1)/\sigma(B_S)$ where $/$ indicates integer division.  
\item $P_{\bar{\eta}} P_\mu$.  A refined transportation process that precedes a transformation process must have a destination equivalent to the transformation resource at which the transformation process was executed.  $m_2-1=  (u_1-1)\%\sigma(B_S)$ where $\%$ indicates the modulus.  
\item $P_{\bar{\eta}} P_{\bar{\eta}}$. A refined transportation process that follows another must have an origin equivalent to the destination of the other.  $(u_1-1) \% \sigma(B_S) = (u_2-1)/\sigma(B_S)$
\item $PP$.  The type of operand of one process must be equivalent to the type of output of another process.  In other words, the ordered pair of processes $P_{w_1}P_{w_2}$ is feasible if and only if $A_P(w_1,w_2)=1$ where $A_P$ is the adjacency matrix that corresponds to a \emph{functional graph} in which pairs of system processes are connected.  
\end{enumerate}
In previous hetero-functional graph theory works, the system sequence constraints matrix $K_\rho$ was calculated straightforwardly using for FOR loops to loop over the indices $\chi_1$ and $\chi_2$ and checking the presence of the five feasibility constraints identified above.   

Here, an alternate approach based upon tensors is provided for insight into the underlying mathematical structure.  For convenience, $\overline{K}_\rho=NOT(K_\rho)$ captures the set of all \emph{feasibility} conditions that pertain to valid sequences of system capabilities.  This set requires that \emph{any} of the first four constraints above \emph{and} the last constraint be satisfied.   

\begin{equation}
\overline{K}_\rho=\left( \overline{K}_{\rho I} \oplus \overline{K}_{\rho II} \oplus \overline{K}_{\rho III} \oplus \overline{K}_{\rho IV}  \right)\cdot \overline{K}_{\rho V}
\end{equation}
where $\oplus$ is Boolean addition (Defn. \ref{defn:boolAdd}) and  $\overline{K}_{\rho I}, \overline{K}_{\rho II}, \overline{K}_{\rho III}, \overline{K}_{\rho IV}, \overline{K}_{\rho V}$ are the matrix implementations of the five types of feasibility constraints identified above.  Their calculation is most readily achieved through their associated 4$^{th}$-order tensors.  

{\color{blue}\vspace{0.1in}\noindent\textbf{Type I Constraints:}}  For the Type I constraint, $\overline{\cal K}_{\rho I}$ is constructed from a sum of 4$^{th}$-order outer products (Defn. \ref{defn:outerProduct}) of elementary basis vectors.  
\begin{align}
\overline{\cal K}_{\rho I} &=
 \sum_{w_1=1}^{\sigma(P_\mu)}
 \sum_{v_1=1}^{\sigma(M)}
 \sum_{w_2=1} ^{\sigma(P_\mu)}
 \sum_{v_2=v_1}^{v_1} 
 e_{w_1}^{\sigma(P)} 
 \circ 
 e_{v_1}^{\sigma(R)} 
 \circ 
 e_{w_2}^{\sigma(P)} 
 \circ e_{v_2}^{\sigma(R)} 
 \end{align}
where the $e_i^n$ notation places the value 1 on the $i^{th}$ element of a vector of length $n$.  $\overline{K}_{\rho I}$ is calculated straightforwardly by matricizing both sides and evaluating the sums. 
 \begin{align}
\overline{K}_{\rho I} &= 
 \sum_{w_1=1}^{\sigma(P_\mu)}
 \sum_{v_1=1}^{\sigma(M)}
 \sum_{w_2=1} ^{\sigma(P_\mu)}
 \sum_{v_2=v_1}^{v_1} 
\left(
e_{v_1}^{\sigma(R)}
\otimes
e_{w_1}^{\sigma(P)} 
\right) 
\otimes 
\left(
e_{v_2}^{\sigma(R)} 
\otimes 
e_{w_2}^{\sigma(P)}
\right)^T \\ 
\overline{K}_{\rho I} &= 
\sum_{v_1=1}^{\sigma(M)} 
\left(
e_{v_1}^{\sigma(R)}
\otimes 
\begin{bmatrix}
\mathds{1}^{\sigma(P_\mu)}\\
\mathbf{0}^{\sigma(P_{\bar{\eta}})}
\end{bmatrix}
\right)
\left(
e_{v_1}^{\sigma(R)} \otimes 
\begin{bmatrix}
\mathds{1}^{\sigma(P_\mu)}\\
\mathbf{0}^{\sigma(P_{\bar{\eta}})}
\end{bmatrix}
\right)^T 
\end{align}

{\color{blue}\vspace{0.1in}\noindent\textbf{Type II Constraints:}}
Similarly, for the Type II constraint:  
\begin{align}\label{eq:KrhoII}
\overline{\cal K}_{\rho II} &=
 \sum_{w_1=1}^{\sigma(P_\mu)}
 \sum_{v_1=1}^{\sigma(M)}
 \sum_{y_1=v_1}^{v_1}
 \sum_{v_2=1}^{\sigma(R)} 
 e_{w_1}^{\sigma(P)} 
 \circ 
 e_{v_1}^{\sigma(R)} 
 \circ 
 \begin{bmatrix}
 \mathbf{0}^{\sigma(P_\mu)}\\
X_{y_1}^{\sigma(P_{\bar{\eta}})}
\end{bmatrix} 
 \circ 
 e_{v_2}^{\sigma(R)} 
 \end{align}
 Here, the $X_{y_1}^{\sigma(P_{\bar{H}})}$ vector has a value of 1 wherever a refined transportation process $p_{w_2}$ originates at the transformation resource $m_{v_1}$.  Drawing on the discussion of the 3rd-order tensor ${\cal J}_{\bar{H}}$ in Section \ref{sec:As}, $X_{y_1}^{\sigma(P_{\bar{H}})}$, itself, is expressed as a vectorized sum of 3$^{rd}$-order outer products.  
\begin{align}
X_{y_1}^{\sigma(P_{\bar{H}})}= 
\sum_{g=1}^{\sigma(P_\gamma)}
\sum_{y_2=1}^{\sigma(B_S)} 
\left(
e_g^{\sigma(P_\gamma)} 
\circ 
e_{y_1}^{\sigma(B_S)} 
\circ 
e_{y_2}^{\sigma(B_S)}
\right)^{TV}= 
\left(
\mathds{1}^{\sigma(P_\gamma)} 
\otimes 
e_{y_1}^{\sigma(B_S)} 
\otimes 
\mathds{1}^{\sigma(B_S)}
\right)
\end{align}
$\overline{K}_{\rho II}$ is then calculated straightforwardly by matricizing both sides of Eq. \ref{eq:KrhoII} and evaluating the sums. 
\begin{align}
\overline{K}_{\rho II} &= 
 \sum_{w_1=1}^{\sigma(P_\mu)}
 \sum_{v_1=1}^{\sigma(M)}
 \sum_{y_1=v_1}^{v_1}
 \sum_{v_2=1}^{\sigma(R)} 
\left(
e_{v_1}^{\sigma(R)} 
\otimes 
e_{w_1}^{\sigma(P)} 
\right) 
\otimes 
\left(
e_{v_2}^{\sigma(R)} 
\otimes 
\begin{bmatrix}
\mathbf{0}^{\sigma(P_\mu)}\\
 X_{y_1}^{\sigma(P_{\bar{H}})}
\end{bmatrix} \right)^T \\ 
\overline{K}_{\rho II} &= 
\sum_{v_1=1}^{\sigma(M)}
\left(e_{v_1}^{\sigma(R)} 
\otimes 
\begin{bmatrix}
\mathds{1}^{\sigma(P_\mu)} \\
\mathbf{0}^{P_{\bar{\eta}}}
\end{bmatrix}
\right) 
\left(
\mathds{1}^{\sigma(R)}
\otimes
\begin{bmatrix}
\mathbf{0}^{\sigma(P_\mu)}\\
\mathds{1}^{\sigma(P_\gamma)} \otimes e_{v_1}^{\sigma(B_S)} \otimes \mathds{1}^{\sigma(B_S)}
\end{bmatrix}
\right)^T
\end{align}

{\color{blue}\vspace{0.1in}\noindent\textbf{Type III Constraints:}}
Similarly, for the Type III constraint:  
\begin{align}\label{eq:KrhoIII}
\overline{\cal K}_{\rho III} &=
 \sum_{y_2=v_2}^{v_2}
 \sum_{v_1=1}^{\sigma(R)}
 \sum_{w_2=1}^{\sigma(P_\mu)}
 \sum_{v_2=1}^{\sigma(M)} 
 \begin{bmatrix}
 \mathbf{0}^{\sigma(P_\mu)}\\
 X_{y_2}^{\sigma(P_{\bar{\eta}})}
\end{bmatrix} 
 \circ 
 e_{v_1}^{\sigma(R)} 
 \circ 
 e_{w_2}^{\sigma(P)} 
\circ 
e_{v_2}^{\sigma(R)} 
\end{align}
 Here, the $ X_{y_2}^{\sigma(P_{\bar{H}})}$ vector has a value of 1 wherever a refined transportation process $p_{w_1}$ terminates at the transformation resource $m_{v_2}$.  $ X_{y_2}^{\sigma(P_{\bar{H}})}$, itself, is expressed as a vectorized sum of 3$^{rd}$-order outer products.  
\begin{align}
X_{y_2}^{\sigma(P_{\bar{H}})}= 
\sum_{g=1}^{\sigma(P_\gamma)}
\sum_{y_1=1}^{\sigma(B_S)} 
\left(
e_g^{\sigma(P_\gamma)} 
\circ 
e_{y_1}^{\sigma(B_S)} 
\circ 
e_{y_2}^{\sigma(B_S)}
\right)^{TV}= 
\left(
\mathds{1}^{\sigma(P_\gamma)} 
\otimes 
\mathds{1}^{\sigma(B_S)} 
\otimes 
e_{y_2}^{\sigma(B_S)}
\right)
\end{align}
$\overline{K}_{\rho III}$ is then calculated straightforwardly by matricizing both sides of Eq. \ref{eq:KrhoIII} and evaluating the sums. 
\begin{align}
\bar{K}_{\rho III} &=
 \sum_{y_2=v_2}^{v_2}
 \sum_{v_1=1}^{\sigma(R)}
 \sum_{w_2=1}^{\sigma(P_\mu)}
 \sum_{v_2=1}^{\sigma(M)} 
\left(
e_{v_1}^{\sigma(R)}  
\otimes  
\begin{bmatrix}
\mathbf{0}^{\sigma(P_\mu)}\\
X_{y_2}^{\sigma(P_{\bar{H}})}
\end{bmatrix}
\right)
\otimes
\left(
e_{v_2}^{\sigma(R)} 
\otimes
e_{w_2}^{\sigma(P)} 
\right)^T\\
\overline{K}_{\rho III} &=
\sum_{v_2=1}^{\sigma(M)} 
\left(
\mathds{1}^{\sigma(R)}  
\otimes  
\begin{bmatrix}
\mathbf{0}^{\sigma(P_\mu)}\\
\mathds{1}^{\sigma(P_\gamma)} 
\otimes 
\mathds{1}^{\sigma(B_S)} 
\otimes 
e_{v_2}^{\sigma(B_S)}
\end{bmatrix}
\right)
\left(
e_{v_2}^{\sigma(R)} 
\otimes
\begin{bmatrix}
\mathds{1}^{\sigma(P_\mu)} \\
\mathbf{0}^{P_{\bar{\eta}}}
\end{bmatrix}
\right)^T
\end{align}

{\color{blue}\vspace{0.1in}\noindent\textbf{Type IV Constraints:}}
Similarly, for the Type IV constraint:  
\begin{align}\label{eq:KrhoIV}
\overline{\cal K}_{\rho IV} &=
 \sum_{y_2=1}^{\sigma{(B_S)}}
 \sum_{v_1=1}^{\sigma(R)}
 \sum_{y_1=y_2}^{y_2}
 \sum_{v_2=1}^{\sigma(R)} 
 \begin{bmatrix}
 \mathbf{0}^{\sigma(P_\mu)}\\
 X_{y_2}^{\sigma(P_{\bar{H}})}
\end{bmatrix} 
 \circ 
 e_{v_1}^{\sigma(R)} 
 \circ 
 \begin{bmatrix}
 \mathbf{0}^{\sigma(P_\mu)}\\
 X_{y_1}^{\sigma(P_{\bar{H}})}
\end{bmatrix} 
\circ 
e_{v_2}^{\sigma(R)} 
\end{align}
$\overline{K}_{\rho IV}$ is then calculated straightforwardly by matricizing both sides of Eq. \ref{eq:KrhoIV} and evaluating the sums. 
\begin{align}
\overline{K}_{\rho IV} &=
 \sum_{y_2=1}^{\sigma{(B_S)}}
 \sum_{v_1=1}^{\sigma(R)}
 \sum_{y_1=y_2}^{y_2}
 \sum_{v_2=1}^{\sigma(R)} 
\left(
e_{v_1}^{\sigma(R)} 
\otimes
 \begin{bmatrix}
 \mathbf{0}^{\sigma(P_\mu)}\\
 X_{y_2}^{\sigma(P_{\bar{H}})}
\end{bmatrix} 
\right)
\otimes
\left(
e_{v_2}^{\sigma(R)} 
\otimes
 \begin{bmatrix}
 \mathbf{0}^{\sigma(P_\mu)}\\
 X_{y_1}^{\sigma(P_{\bar{H}})}
\end{bmatrix} 
\right)^T\\
\overline{K}_{\rho IV} &=
 \sum_{y_2=1}^{\sigma{(B_S)}}
 \left(
\mathds{1}^{\sigma(R)} 
\otimes
 \begin{bmatrix}
 \mathbf{0}^{\sigma(P_\mu)}\\
\mathds{1}^{\sigma(P_\gamma)} 
\otimes 
\mathds{1}^{\sigma(B_S)} 
\otimes 
e_{y_2}^{\sigma(B_S)} 
\end{bmatrix} 
\right)
\left(
\mathds{1}^{\sigma(R)} 
\otimes
 \begin{bmatrix}
 \mathbf{0}^{\sigma(P_\mu)}\\
 \mathds{1}^{\sigma(P_\gamma)} 
\otimes 
e_{y_2}^{\sigma(B_S)} 
\otimes 
\mathds{1}^{\sigma(B_S)}
\end{bmatrix} 
\right)^T
\end{align}

{\color{blue}\vspace{0.1in}\noindent\textbf{Type V Constraints:}}
The Type V constraint must make use of the functional graph adjacency matrix $A_P$.  Consequently, the fourth-order tensor ${\cal K}_{\rho V}$ is calculated first on a scalar basis using the Kronecker delta function $\delta_i$ (Defn. \ref{defn:kronDelta}) and then is matricized to $K_{\rho V}$.    
\begin{align}
{\cal \overline{K}}_{\rho V}(w_1,v_1,w_2,v_2)&=
\delta_{v_1v_1} \cdot \delta_{v_2v_2} \cdot A_P(w_1,w_2) 
\qquad \forall w_1,w_2 \in \{1, \ldots, \sigma(P)\},
\; v_1,v_2 \in  \{1, \ldots, \sigma(R)\}
\\
\overline{K}_{\rho V}&=
\sum_{v_1=1}^{\sigma(R)}
\sum_{v_2=1}^{\sigma(R)}
\left(
e_{v_1}^{\sigma(R)}
\otimes
e_{v_2}^{\sigma(R)T}
\right)
\otimes
A_P\\
\overline{K}_{\rho V}&=
\left(
\mathds{1}^{\sigma(R)}
\otimes
\mathds{1}^{\sigma(R)T}
\right)
\otimes
A_P
\end{align}

{\color{blue}\subsection{Sequence-Dependent Degrees of Freedom}\label{sec:DOFrho}}
Once the system sequence knowledge base and constraints matrix have been calculated, the number of sequence-dependent degrees of freedom follow straightforwardly.   
\begin{defn}[Sequence-Dependent Degrees of Freedom \cite{Farid:2015:ISC-J19,Farid:2015:SPG-J17,Viswanath:2013:ETS-J08,Farid:2016:ETS-J27,Schoonenberg:2017:IEM-J34}]\label{def:DOFrho}  The set of independent pairs of actions $z_{\chi_1\chi_2}=e_{w_1v_1}e_{w_2v_2} \in {\cal Z}$ of length 2 that completely describe the system language. The number is given by:
\begin{align}\label{Eq:DOFRho}
DOF_{\rho} = \sigma({\cal Z}) &= \sum_{\chi_1}^{\sigma(R)\sigma(P)} \sum_{\chi_2}^{\sigma(R)\sigma(P)} [J_\rho\ominus K_\rho](\chi_1,\chi_2)\\ \label{Eq:DOFRhoAR}
&=\sum_{\chi_1}^{\sigma(R)\sigma(P)}\sum_{\chi_2}^{\sigma(R)\sigma(P)}[A_\rho](\chi_1,\chi_2)
\end{align}
\end{defn}
\nomenclature[E]{$DOF_\rho$}{Sequence-Dependent Degrees of Freedom}

For systems of substantial size, the size of the hetero-functional adjacency matrix may be challenging to process computationally. However, the matrix is generally very sparse.  Therefore, projection operators are used to eliminate the sparsity by projecting the matrix onto a one's vector\cite{Farid:2016:ETS-J27,Schoonenberg:2017:IEM-J34}.  This is demonstrated below for $J_S^V$ and $A_\rho$:
\begin{align}
    \mathds{P}_S J_S^V &= \mathds{1}^{\sigma({\cal E}_S)} \\ \label{eq:projHFAM}
    \mathds{P}_S A_\rho \mathds{P}_S^T &= \widetilde{A}_{\rho}
\end{align}
\nomenclature[E]{$\mathds{P}_S$}{A (non-unique) projection matrix for the vectorized knowledge base}
\nomenclature[E]{$\widetilde{A}_{\rho}$}{Hetero-functional Adjacency Matrix after elimination of row and column sparsity}

\noindent where $\mathds{P}_S$ is a (non-unique) projection matrix for the vectorized system knowledge base and the hetero-functional adjacency matrix \cite{Farid:2016:ETS-J27,Schoonenberg:2017:IEM-J34}.  Note that the number of sequence dependent degrees of freedom for the projected hetero-functional adjacency matrix can be calculated as:
\begin{equation}
DOF_{\rho} = \sigma({\cal Z}) = \sum_{\psi_1}^{\sigma({\cal E}_S)} \sum_{\psi_2}^{\sigma({\cal E}_S)} [\widetilde{A}_\rho](\psi_1,\psi_2)
\end{equation}
where $\psi \in \left[1, \dots ,\sigma({\cal E}_S)\right]$. 
\nomenclature[D]{$\psi$}{Index of the elements in the set of structural degrees of freedom ${\cal E}_S$}

\section{Hetero-functional Incidence Tensor}\label{sec:Mrho}
{\color{blue}This section serves to introduce the hetero-functional incidence tensor as part of the paper's original contribution.  Subsection \ref{sec:Mrho3} describes the tensor in third-order form.  Subsection \ref{sec:Mrho4} then elaborates why it sometimes useful to present this tensor in fourth-order form.  Finally, Subsection \ref{sec:Mrho2} shows how matricizing the heter-functional incidence tensor (into second-order form) can serve to reconstruct the hetero-functional adjacency matrix.

\subsection{Third Order Form}\label{sec:Mrho3}}
To complement the concept of a hetero-functional adjacency matrix $A_\rho$ and its associated tensor ${\cal A}_\rho$,  the hetero-functional incidence tensor $\widetilde{\cal M}_\rho$ describes the structural relationships between the physical capabilities (i.e. structural degrees of freedom) ${\cal E}_S$, the system operands $L$, and the system buffers $B_S$.   
 
\begin{equation}
\widetilde{\cal M}_\rho=\widetilde{\cal M}_\rho^+-\widetilde{\cal M}_\rho^-
\end{equation}
\begin{defn}[The Negative 3$^{rd}$ Order Hetero-functional Incidence Tensor $\widetilde{\cal M}_\rho^-$]
The negative hetero-functional incidence tensor $\widetilde{\cal M_\rho}^- \in \{0,1\}^{\sigma(L)\times \sigma(B_S) \times \sigma({\cal E}_S)}$  is a third-order tensor whose element $\widetilde{\cal M}_\rho^{-}(i,y,\psi)=1$ when the system capability ${\epsilon}_\psi \in {\cal E}_S$ pulls operand $l_i \in L$ from buffer $b_{s_y} \in B_S$.
\end{defn} 
\begin{defn}[The Positive  3$^{rd}$ Order Hetero-functional Incidence Tensor $\widetilde{\cal M}_\rho^+$]
The positive hetero-functional incidence tensor $\widetilde{\cal M}_\rho^+ \in \{0,1\}^{\sigma(L)\times \sigma(B_S) \times \sigma({\cal E}_S)}$  is a third-order tensor whose element $\widetilde{\cal M}_\rho^{+}(i,y,\psi)=1$ when the system capability ${\epsilon}_\psi \in {\cal E}_S$ injects operand $l_i \in L$ into buffer $b_{s_y} \in B_S$.
\end{defn} 

{\color{blue}The calculation of these two tensors depends on the definition of two more matrices which further depend on the hetero-functional graph theory definitions in Appendix \ref{sec:hfgt_dfn}.}
\begin{defn}[The Negative Process-Operand Incidence Matrix $M_{LP}^-$]\label{def:negativeprocess-operandIM}
A binary incidence matrix $M_{L P}^{-} \in \{0,1\}^{\sigma(L)\times \sigma(P)}$ whose element $M_{L P}^{-}(i,w)=1$ when the system process $p_w \in P$ pulls operand $l_i \in L$ as an input.  It is further decomposed into the negative transformation process-operand incidence matrix $M_{L P_\mu}^-$ (Defn. \ref{def:negxFormProcOper}) and the negative refined transformation process-operand incidence matrix $M_{L P_{\bar{\eta}}}^-$ (Defn. \ref{def:negxPortRefProcOper}) which by definition is in turn calculated from the negative holding process-operand incidence matrix $M_{LP_{\gamma}}^-$ (Defn. \ref{defn:negHoldProcOper}).
\begin{equation}
M_{L P}^-=
\begin{bmatrix}
M_{L P_\mu}^{-} & M_{L P_{\bar{\eta}}}^-
\end{bmatrix}
=
\begin{bmatrix}
M_{L P_\mu}^{-} & M_{LP_{\gamma}}^- \otimes \mathds{1}^{\sigma(P_\eta)T}
\end{bmatrix}
\end{equation}
\end{defn} 

\begin{defn}[The Positive Process-Operand Incidence Matrix $M_{LP}^+$]
A binary incidence matrix $M_{LP}^{+} \in \{0,1\}^{\sigma(L)\times \sigma(P)}$ whose element $M_{L P}^{+}(i,w)=1$ when the system process $p_w \in P$ injects operand $l_i \in L$ as an output.  It is further decomposed into the positive transformation process-operand incidence matrix $M_{LP_\mu}^+$ (Defn. \ref{def:posxFormProcOper}) and the positive refined transformation process-operand incidence matrix $M_{LP_{\bar{\eta}}}^+$ (Defn. \ref{def:posxPortRefProcOper}) which, by definition, is, in turn, calculated from the positive holding process-operand incidence matrix $M_{LP_{\gamma}}^+$ (Defn. \ref{defn:posHoldProcOper})
\begin{equation}
M_{LP}^+=
\begin{bmatrix}
M_{LP_\mu}^{+} & M_{LP_{\bar{\eta}}}^+
\end{bmatrix}
=
\begin{bmatrix}
M_{LP_\mu}^{+} & M_{LP_{\gamma}}^+ \otimes \mathds{1}^{\sigma(P_\eta)T}
\end{bmatrix}
\end{equation}
\end{defn}

With the definitions of these incidence matrices in place, the calculation of the negative and positive hetero-functional incidence tensors $\widetilde{\cal M}_\rho^-$ and $\widetilde{\cal M}_\rho^+$ follows straightforwardly as a third-order outer product.  For $\widetilde{\cal M}_\rho^-$:
\begin{align}\label{Eq:MrhoMinus}
\widetilde{\cal M}_\rho^-=\sum_{i=1}^{\sigma(L)}\sum_{y_1=1}^{\sigma(B_S)} e_i^{\sigma(L)} \circ e_{y_1}^{\sigma(B_S)} \circ \mathds{P}_S \left(\left(X^{-}_{i y_1} \right)^V\right)
\end{align}
where 
\begin{equation}
X^{-}_{i y_1}=
\left[\begin{array}{c}
M_{LP_\mu}^{-T}e_{i}^{\sigma(L)}e_{y_1}^{\sigma(M)T} \quad | \quad  \mathbf{0}\\\hline
M_{LP_{\gamma}}^{-T}e_{i}^{\sigma(L)} \otimes \left(e_{y_1}^{\sigma(B_S)} \otimes \mathds{1}^{\sigma(B_S)}\right) \otimes \mathds{1}^{\sigma(R)T}\\
\end{array}\right]
\end{equation}
The $X^{-}_{i y_1}$ matrix is equivalent in size to the system concept $A_S$.  It has a value of one in all elements where the associated process both withdraws input operand $l_i$ and originates at the buffer $b_{s_{y_1}}$.  Consequently, when $X^{-}_{i y_1}$ is vectorized and then projected with $\mathds{P}_S$, the result is a vector with a value of one only where the associated system capabilities meet these criteria.  

For $\widetilde{\cal M}_\rho^+$:
\begin{align}\label{Eq:MrhoPlus}
\widetilde{\cal M}_\rho^+=\sum_{i=1}^{\sigma(L)}\sum_{y_2=1}^{\sigma(B_S)} e_i^{\sigma(L)} \circ e_{y_2}^{\sigma(B_S)} \circ \mathds{P}_S \left(\left(X^{+}_{i y_2} \right)^V\right)
\end{align}
where 
\begin{equation}
X^{+}_{i y_2}=
\left[\begin{array}{c}
M_{LP_\mu}^{+T}e_{i}^{\sigma(L)}e_{y_2}^{\sigma(M)T} \quad | \quad  \mathbf{0}\\\hline
M_{LP_{\gamma}}^{+T}e_{i}^{\sigma(L)} \otimes \left(\mathds{1}^{\sigma(B_S)} \otimes e_{y_2}^{\sigma(B_S)}\right) \otimes \mathds{1}^{\sigma(R)T}\\
\end{array}\right]
\end{equation}
The $X^{+}_{i y_2}$ matrix is equivalent in size to the system concept $A_S$.  It also has a value of one in all elements where the associated process both injects output operand $l_i$ and terminates at the buffer $b_{s_{y_2}}$.  Consequently, when $X^{+}_{i y_2}$ is vectorized and then projected with $\mathds{P}_S$, the result is a vector with a value of one only where the associated system capabilities meet these criteria. 

It is important to note that the definitions of the 3$^{rd}$ order hetero-functional incidence tensors $\widetilde{\cal M}_\rho^-$, and $\widetilde{\cal M}_\rho^+$ are provided in projected form as indicated by the presence of the projection operator $\mathds{P}_S$ in Equations \ref{Eq:MrhoMinus} and \ref{Eq:MrhoPlus} respectively.  It is often useful to use the un-projected form of these tensors.  
\begin{align}\label{Eq:MrhoMinusSparse}
{\cal M}_\rho^-=\sum_{i=1}^{\sigma(L)}\sum_{y_1=1}^{\sigma(B_S)} e_i^{\sigma(L)} \circ e_{y_1}^{\sigma(B_S)} \circ \left(X^{-}_{i y_1} \right)^V
\end{align}
\begin{align}\label{Eq:MrhoPlusSparse}
{\cal M}_\rho^+=\sum_{i=1}^{\sigma(L)}\sum_{y_2=1}^{\sigma(B_S)} e_i^{\sigma(L)} \circ e_{y_2}^{\sigma(B_S)} \circ \left(X^{+}_{i y_2} \right)^V
\end{align}

{\color{blue}\subsection{Fourth Order Form}\label{sec:Mrho4}}
The third dimension of these unprojected 3$^{rd}$ order hetero-functional incidence tensors can then be split into two dimensions to create 4$^{th}$ order hetero-functional incidence tensors.  
 \begin{align}
 {\cal M}_{PR}^+&=vec^{-1}\left({\cal M}_\rho^+,[\sigma(P),\sigma(R)],3\right)\\
  {\cal M}_{PR}^-&=vec^{-1}\left({\cal M}_\rho^-,[\sigma(P),\sigma(R)],3\right)
 \end{align}
These fourth order tensors describe the structural relationships between the system processes $P$, the physical resources $R$ that realize them, the system operands $L$ that are consumed and injected in the process, and the system buffers $B_S$ from which these are operands are sent and the system buffers $B_S$ to which these operands are received.  They are used in the following section as part of the discussion on layers. 
\begin{equation}
{\cal M}_{PR}={\cal M}_{PR}^+ - {\cal M}_{PR}^-
\end{equation}
\begin{defn}[The Negative 4$^{th}$ Order Hetero-functional Incidence Tensor ${\cal M}_{PR}^-$]
The negative 4$^{th}$ Order hetero-functional incidence tensor ${\cal M}_{PR}^- \in \{0,1\}^{\sigma(L)\times \sigma(B_S) \times \sigma(P) \times \sigma(R)}$  has element ${\cal M}_{PR}^{-}(i,y,w,v)=1$ when the system process $p_w \in P$ realized by resource $r_v \in R$ pulls operand $l_i \in L$ from buffer $b_{s_y} \in B_S$.
\end{defn} 
\begin{defn}[The Positive 4$^{th}$ Order Hetero-functional Incidence Tensor ${\cal M}_{PR}^-$]
The positive 4$^{th}$ Order hetero-functional incidence tensor ${\cal M}_{PR}^- \in \{0,1\}^{\sigma(L)\times \sigma(B_S) \times \sigma(P) \times \sigma(R)}$  has element ${\cal M}_{PR}^{-}(i,y,w,v)=1$ when the system process $p_w \in P$ realized by resource $r_v \in R$ injects operand $l_i \in L$ into buffer $b_{s_y} \in B_S$.
\end{defn} 

Furthermore, the negative and positive 4$^{th}$ hetero-functional incidence tensors can be used to demonstrate a direct relationship to the system concept $A_S$.  
\begin{align}
{\cal M}_{PR}^{-}(i,y,w,v) =& X^{-}_{i y_1}(w,v) \cdot A_S(w,v) \\\label{Eq:MPR1-AS}
{\cal M}_{PR}^{-} =& \sum_i^{\sigma(L)}\sum_{y_1}^{\sigma(B_S)} X^{-}_{i y_1} \cdot A_S\\
{\cal M}_{PR}^{+}(i,y,w,v) =& X^{+}_{i y_2}(w,v) \cdot A_S(w,v) \\\label{Eq:MPR2-AS}
{\cal M}_{PR}^{+} =& \sum_i^{\sigma(L)}\sum_{y_2}^{\sigma(B_S)} X^{+}_{i y_2} \cdot A_S
\end{align}
Equations \ref{Eq:MPR1-AS} and \ref{Eq:MPR2-AS} show that the 4$^{th}$ order hetero-functional incidence tensors contain three types of information:
\begin{enumerate}
\item the mapping of system processes to system resources in the system concept $A_S$,
\item the mapping of processes to their operands in $M_{LP}^-$ and $M_{LP}^+$, 
\item the implicit knowledge that by definition transformation processes occur at a stationary buffer, and that transportation processes are defined by their origin and destination buffers.  
\end{enumerate}
In other words, the hetero-functional incidence tensor is a complete descption of a system's allocated architecture.  

{\color{blue}\subsection{Second Order Form}\label{sec:Mrho2}}
Returning back to the third-order hetero-functional incidence tensor $\widetilde{M}_\rho$, it and and its positive and negative components $\widetilde{M}_\rho^+, \widetilde{M}_\rho^-$, can also be easily matricized.  
\begin{align}
M_\rho&={\cal F}_M\left({\cal M}_\rho,[1,2],[3]\right)\\
M_\rho^-&={\cal F}_M\left({\cal M}_\rho^-,[1,2],[3]\right)\\
M_\rho^+&={\cal F}_M\left({\cal M}_\rho^+,[1,2],[3]\right)
\end{align}
The resulting matrices have a size of $\sigma(L)\sigma(B_S) \times \sigma(\cal{E_S})$ which have a corresponding physical intuition.   Each buffer $b_{s_{y}}$ has $\sigma(L)$ copies to reflect a place (i.e. bin) for each operand at that buffer.  Each of these places then forms a bipartite graph with the system's physical capabilities.   Consequently, and as expected, the hetero-functional adajacency matrix $A_\rho$ can be calculated as a matrix product of the positive and negative hetero-functional incidence matrices $M_\rho^+$ and $M_\rho^+$.  
 \begin{equation}\label{eq:bigresult}
 A_\rho = M_\rho^{+T}\odot M_\rho^- = M_\rho^{+T}M_\rho^-
 \end{equation}
 Such a product systematically enforces all five of the feasibility constraints identified in Section \ref{sec:Arho}.  Furthermore, the Boolean and real matrix products are interchangeable because each process is associated with exactly one origin-destination pair.    
 
\section{Discussion}\label{sec:discussion}
{\color{black}Given the discussion on multi-layer networks in the introduction, it is worthwhile reconciling the gap in terminology between multi-layer networks and hetero-functional graph theory.  First, the concept of layers in hetero-functional graphs is discussed.  Second, an ontological comparison of layers in hetero-functional graphs and multi-layer networks is provided.  Third, a discussion of network descriptors in the context of layers is provided.  Given the ``disparate terminology and the lack of consensus" in the multi-layer network literature, the discussion uses the multi-layer description provided De Dominico et. al\cite{De-Domenico:2013:00}.}

\subsection{Layers in Hetero-functional Graphs}

\begin{defn}[Layer]\cite{Thompson:2022:ISC-C80}
A layer $G_{\lambda} = \{{\cal E}_{S\lambda}, Z_{S\lambda} \}$ of a hetero-functional graph $G = \{{\cal E}_{S}, Z_{S}\}$ is a subset of a hetero-functional graph, $G_{\lambda} \subseteq G$, for which a predefined layer selection (or classification) criterion applies.  A set of layers in a hetero-functional graph adhere to a classification scheme composed of a number of selection criteria.
\end{defn}
\noindent Note that this definition of a layer is particularly flexible because it depends on the nature of the classification scheme and its associated selection criteria.  Nevertheless, and as discussed later, it is important to choose a classification scheme that leads to a set of mutually exclusive layers that are also collectively exhaustive of the hetero-functional graph as a whole.

To select out specific subsets of capabilities (or structural degrees of freedom), HFGT has used the concept of ``selector matrices" of various types\cite{Schoonenberg:2019:ISC-BK04,Farid:2017:IEM-J13}.  Here a layer selector matrix is defined. 
\begin{defn}\cite{Thompson:2022:ISC-C80}
Layer Selector Matrix: A binary matrix $\Lambda_\lambda$ of size $\sigma(P)\times \sigma(R)$ whose element $\Lambda_\lambda(w,v)=1$ when the capability $e_{wv} \subset E_{S\lambda}$.   
\end{defn}

From this definition, the calculation of a hetero-functional graph layer follows straightforwardly.  First, a layer projection operator $\mathds{P}_\lambda$ is calculated\cite{Thompson:2022:ISC-C80}: 
\begin{align}
    \mathds{P}_\lambda \Lambda_\lambda^V &= \mathds{1}^{\sigma({\cal E}_S)} \label{eq:projHFAMLayer}
\end{align}
 Next, the negative and positive hetero-functional incidence tensors $\widetilde{\cal M}_{\rho\lambda}^-$ and $\widetilde{\cal M}_{\rho\lambda}^+$ for a given layer $\lambda$ are calculated straightforwardly\cite{Thompson:2022:ISC-C80}.

\begin{align}\label{Eq:MrhoMinusLambda}
\widetilde{\cal M}_{\rho\lambda}^-=\sum_{i=1}^{\sigma(L)}\sum_{y_1=1}^{\sigma(B_S)} e_i^{\sigma(L)} \circ e_{y_1}^{\sigma(B_S)} \circ \mathds{P}_\lambda \left(\left(X^{-}_{i y_1} \right)^V\right) = \widetilde{\cal M}_{\rho}^- \odot_3 \mathds{P}_\lambda
\end{align}
\begin{align}\label{Eq:MrhoPlusLambda}
\widetilde{\cal M}_{\rho\lambda}^+=\sum_{i=1}^{\sigma(L)}\sum_{y_2=1}^{\sigma(B_S)} e_i^{\sigma(L)} \circ e_{y_2}^{\sigma(B_S)} \circ \mathds{P}_S \left(\left(X^{+}_{i y_2} \right)^V\right) = \widetilde{\cal M}_{\rho}^+ \odot_3 \mathds{P}_\lambda
\end{align}
From there, the positive and negative hetero-functional incidence tensors for a given layer can be matricized and the adjacency matrix of the associated layer $\widetilde{A}_{\rho \lambda}$ follows straightforwardly\cite{Thompson:2022:ISC-C80}. 
\begin{align}
\widetilde{M}_{\rho \lambda}^+ &= {\cal F}_M (\widetilde{\cal M}_{\rho \lambda}^+, [1,2],[3]) \\
\widetilde{M}_{\rho \lambda}^- &= {\cal F}_M (\widetilde{\cal M}_{\rho \lambda}^-, [1,2],[3]) \\
\widetilde{A}_{\rho \lambda} &= \widetilde{M}_{\rho \lambda}^{+ T} \odot \widetilde{M}_{\rho \lambda}^{-}
\end{align}

This approach of separating a hetero-functional graph into its constituent layers is quite generic because the layer selector matrix $\Lambda_\lambda$ can admit a wide variety of classification schemes.  Three classification schemes are discussed here:  
\begin{enumerate}
    \item An Input Operand Set Layer
    \item An Output Operand Set Layer
    \item A Dynamic Device Model Layer
\end{enumerate}

\liinesfig{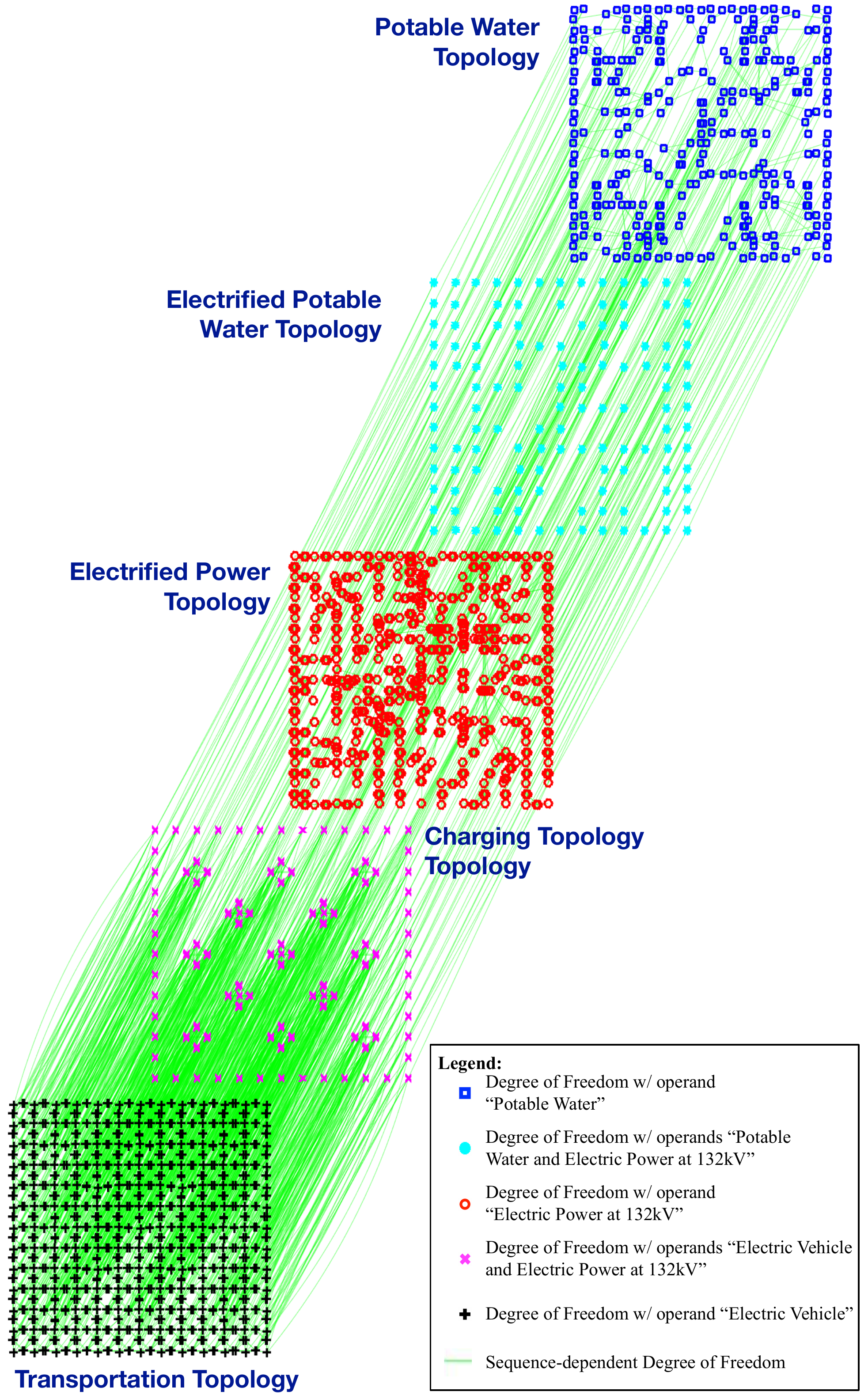}{The Trimetric Smart City Infrastructure Test Case Visualized as Five Layers Defined by Input Operand Sets:  The Potable Water Topology, The Electrified Potable Water Topology, the Electric Power Topology, the Charging Topology, and the Transportation Topology}{fig:HFGT}

\begin{defn}\label{defn:InputOperandSetLayer}
Input Operand Set Layer:  A hetero-functional graph layer for which all of the node-capabilities have a common set of input operands $L_\lambda \subseteq L$.  
\end{defn}
\noindent This definition of an Operand Set Layer was used in the HFGT text\cite{Schoonenberg:2019:ISC-BK04} to partition the Trimetrica test case (first mentioned in Figure \ref{fig:Trimetrica}) into the multi-layer depiction in Figure \ref{fig:HFGT}.  In this classification scheme, any system contains up to 2$^{\sigma(L)}$ possible layers. For completeness, an index $\lambda_D \in \{1, \dots, 2^{\sigma(L)}\}$ is used to denote a given layer.  In reality, however, the vast majority of physical systems exhibit far fewer than 2$^{\sigma(L)}$ layers.  Consequently, it is often useful to simply assign an index $\lambda$ to each layer and create a 1-1 mapping function (i.e. lookup table) $f_\lambda$ back to the $\lambda_D$ index.
\begin{equation}
f_\lambda: \lambda \rightarrow \lambda_D
\end{equation}
The utility of the $\lambda_d$ index (stated as a base 10 number) becomes apparent when it is converted into a binary (base 2) number $\lambda_v \in \{0,1\}^{\sigma(L)}$ which may be used equivalently as a binary vector of the same length.    
\begin{equation}
\lambda_v = bin(\lambda_D) 
\end{equation}
The resulting binary vector $\lambda_v$ has the useful property that $\lambda_v (l_i) = 1$, iff operand $l_i \in L_\lambda$.  Consequently, a given value of $\lambda_v$ serves to select from $L$ the operands that pertain to layer $\lambda$.  The associated layer selector matrix follows straightforwardly:
\begin{equation}\label{Eq:Lambda1selector}
\Lambda_\lambda(w,v)= 
\left\{
\begin{array}{cc}
1 & \mbox{if} \quad \lambda_v = M_{LP}^-(:,w) \quad \forall r_v \in R\\
0 & \mbox{otherwise} \; 
\end{array}
\right.
\end{equation}
It is also worth noting that the layer selector matrix $\Lambda$ above is effectively a third order tensor whose value $\Lambda(\lambda,w,v)=1$ when the capability $e_{wv}$ is part of layer $\lambda$.  

One advantage of a classification scheme based on \textbf{\emph{sets}} of input operands is that they lead to the generation of a mutually exclusive and collectively exhaustive set of layers.  Because no process (and consequently capability) has two sets of input operands, it can only exist in a single layer (mutual exclusivity).  In the meantime, the presence of $2^{\sigma(L)}$ assures that all capabilities fall into (exactly) one layer (exhaustivity).  It is worth noting that a classification scheme based on \textbf{\emph{individual}} operands would not yield these properties.  For example, a water pump consumes electricity and water as input operands.  Consequently, it would have a problematic existence in both the ``water layer" as well as the ``electricity layer".  In contrast, a classification scheme based on operand sets creates an ``electricity-water" layer.  

Analogously to Defn. \ref{defn:InputOperandSetLayer}, an output-operand set layer can be defined and its associated layer selector matrix calculated.  
\begin{defn}
Output Operand Set Layer:  A hetero-functional graph layer for which all of the node-capabilities have a common set of output operands $L_\lambda \subseteq L$.  
\end{defn}
\begin{equation}\label{Eq:Lambda2selector}
\Lambda_\lambda(w,v)= 
\left\{
\begin{array}{cc}
1 & \mbox{if} \quad \lambda_v = M_{LP}^+(:,w) \quad \forall r_v \in R\\
0 & \mbox{otherwise} \; 
\end{array}
\right.
\end{equation}

The third classification scheme is required when developing dynamic equations of motion from the structural information of a hetero-functional graph\cite{Farid:2015:SPG-J17,Schoonenberg:2022:ISC-J50}.  Every process is said to have a ``dynamic device model" that is usually described as a set of differential, algebraic, or differential-algebraic equations\cite{Farid:2015:SPG-J17,Schoonenberg:2022:ISC-J50}.  The simplest of these are the constitutive laws of basic dynamic system elements (e.g. resistors, capacitors, and inductors).  Some processes, although distinct, may have device models with the same functional form.  For example, two resistors at different places in an electrical system have the same constitutive (Ohm's) law, but have different transportation processes because their origin and destinations are different.  Consequently, layers that distinguish on the basis of dynamic device model (i.e. constitutive law) are necessary.  
\begin{defn}
Dynamic Device Model Layer:  A hetero-functional graph layer for which all of the node-capabilities have a dynamic device model with the same functional form.  
\end{defn}
\noindent In such a case, the layer selector matrix $\Lambda_\lambda$ straightforwardly maps capabilities to their layer and dynamic device model interchangeably.  A sufficient number of layers need to be created to account for all of the different types of dynamic device models in the system.  This classification scheme may be viewed as a generalization of the well-known literature on ``linear-graphs"\cite{Rowell:1997:00} and ``bond graphs"\cite{Karnopp:1990:00}.

\subsection{Finding Commonality between Multilayer Networks and Hetero-functional Graphs}
The above discussion of layers in a hetero-functional graph inspires a comparison with multi-layer networks.  The multi-layer adjacency tensor (${\cal A}_{MLN}$) defined by De Dominico et. al.\cite{De-Domenico:2013:00} is chosen to facilitate the discussion.  This fourth order tensor has elements ${\cal A}_{MLN}(\alpha_1,\alpha_2,\beta_1,\beta_2)$ where the indices $\alpha_1,\alpha_2$ denote ``vertices" and $\beta_1,\beta_2$ denote ``layers".  De Dominico et. al write that this multilayer adjacency tensor is a \cite{De-Domenico:2013:00}: \emph{``\dots very general object that can be used to represent a wealth of complicated relationships among nodes."}  The challenge in reconciling the multi-layer adjacency tensor ${\cal A}_{MLN}$ and the hetero-functional adjacency tensor ${\cal A}_{\rho}$ is an ontological one.  Referring back to the ontological discussion in the introduction and more specifically Figure \ref{fig:Ontology} reveals that the underlying abstract conceptual elements (in the mind) to which these two mathematical models refer may not be the same.  

Consider the following interpretation of ${\cal A}_{MLN}(\alpha_1,\alpha_2,\beta_1,\beta_2)={\cal A}_{B_Sl_1}(y_1,y_2,i_1,i_2)$ where the multi-layer network's vertices are equated to the buffers $B_S$ and the layers are equated to the operands $L$.  This interpretation would well describe the departure of an operand $l_{i_1}$ from buffer $b_{sy_1}$ and arriving as $l_{i_2}$ at $b_{sy_2}$.  The equivalence of vertices to buffers is effectively a consensus view in the literature.  In contrast, the concept of a ``layer" in a multi-layer network (as motivated in the introduction) remains relatively unclear.  The equivalence of layers to operands warrants further attention.  
\begin{thm}
The mathematical model $A_{B_Sl_1}$ is neither lucid nor complete with respect to the system processes $P$ (as an abstraction). 
\end{thm}
\begin{prf} 
By contradiction. Assume that ${\cal A}_{B_Sl_1}$ is both lucid and complete network model with respect to system processes $P$.  Consider an operand $l_1$ that departs $b_{s1}$, undergoes process $p_1$, and arrives as $l_1$ at $b_{s2}$.  Now consider the same operand $l_1$ that departs $b_{s1}$, undergoes process $p_2$, and arrives as $l_1$ at $b_{s2}$.  Both of these scenarios would be denoted by ${\cal A}_{B_Sl_1}(1,2,1,1)=1$.  Consequently, this modeling element is overloaded and as such violates the ontological property of lucidity.  Furthermore, because ${\cal A}_{B_Sl_1}$ makes no mention of the concept of system processes, then it violates the completeness property as well.  
\end{prf}

The counter-example provided in the proof above is not simply a theoretical abstraction but rather quite practical.  For several decades, the field of mechanical engineering has used ``linear graphs"\cite{Rowell:1997:00} to derive the equations of motion of dynamic systems with multi-domain physics.  Consider the RLC circuit shown in Figure \ref{fig:RLC} and its associated linear graph.  As parallel elements, the inductor and capacitor both transfer electrical power (as an operand) between the same pair of nodes.  However, the constitutive law (as a physical process) of a capacitor is distinct from that of the inductor.  Consequently, the interpretation ${\cal A}_{B_Sl_1}$ of a multi-layer network is inadequate even for this very simple counter-example\footnote{Although electric power systems and circuits have served as a rich application domain for graph theory and network science, these approaches usually parameterize the circuit components homogeneously as a fixed-value impedance/admittance at constant frequency.  When the constant frequency assumption is relaxed, the diversity of constitutive laws for resistors, capacitors, and inductors must be explicitly considered.}.  

\liinesbigfig{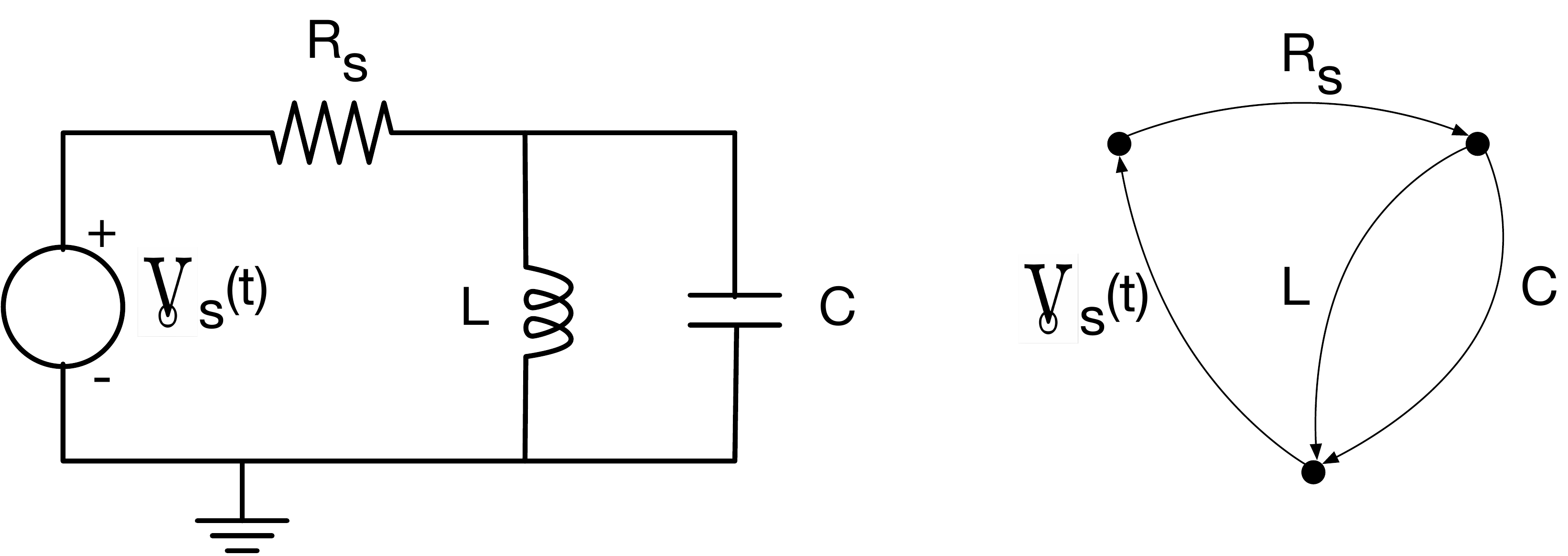}{A Simple RLC Circuit shown as a circuit diagram on left and as a linear graph model on right.  Each resistor, capacitor and inductor can be said to be part of its own layer by virtue of their distinct constitutive laws.}{fig:RLC}

Another possible interpretation of a multi-layer network is ${\cal A}_{MLN}(\alpha_1,\alpha_2,\beta_1,\beta_2)$ = ${\cal A}_{B_Sl_2}(y_1,y_2,w_1,w_2)$ where the multi-layer network's vertices are equated to the buffers $B_S$ and the layers are equated to the processes $P$.  This interpretation would well describe the execution of a process $p_{w_1}$ that is realized by buffer $b_{sy_1}$ followed by a process $p_{w_2}$ that is realized by buffer $b_{sy_2}$.  The equivalence of layers to processes warrants further attention as well.  
\begin{thm}
The mathematical model $A_{B_Sl_2}$ is neither lucid nor complete with respect to the system's transportation resources $H$ (as an abstraction). 
\end{thm}
\begin{prf} 
By contradiction. Assume that ${\cal A}_{B_Sl_2}$ is both a lucid and complete network model with respect to system's transportation resources $H$.  Consider transportation process $p$ between a buffer $b_{s1}$ and a distinct buffer $b_{s2}$.  If such a transportation process were realized by any buffer $b_s \in B_S$, then by definition it would no longer be a buffer but rather a transportation resource.  Consequently,  ${\cal A}_{B_Sl_2}$ is not complete with respect the system's transportation resources $H$.  Now consider a process $p_{1}$ that is realized by buffer $b_{s1}$ followed by a process $p_{2}$ that is realized by a distinct buffer $b_{s2}$.  This is denoted by ${\cal A}_{B_Sl_2}(1,2,1,2)=1$.  Given the distinctness of $b_{s1}$ and $b_{s2}$, a transportation process must have happened in between $p_1$ and $p_2$ although it is not explicitly stated by the mathematical statement  ${\cal A}_{B_Sl_2}(1,2,1,2)=1$.  Such a transportation process, although well-defined by its origin and destination could have been realized by any one of a number of transportation resources.  Consequently, the modeling element is overloaded and as such violates the property of lucidity.  The lack of an explicit description of transportation processes or resources limits the utility of this type of multi-layer network model.  
\end{prf}

It is worth noting that the first multi-layer network interpretation ${\cal A}_{B_Sl_1}$ can be derived directly from the positive and negative hetero-functional incidence matrices\cite{Thompson:2022:ISC-C80}.  
\begin{align}
 {\cal A}_{B_SL_1}(y_1,y_2,i_1,i_2) =&  \bigvee_\psi {\cal M}_\rho^-(i,y_1,\psi)\cdot{\cal M}_\rho^+(i,y_2,\psi) \\
{\cal A}_{B_Sl_1} =& {\cal F}_M^{-1}\left(M_\rho^{-T}\odot M_\rho^+, [\sigma(B_S),\sigma(B_S),\sigma(L),\sigma(L)],[1,3],[2,4]\right)
\end{align}
When $M_\rho^{-T}$ and $M_\rho^+$ are multiplied so that the capabilities ${\cal E}_s$ are the inner dimension, the result is an adjacency matrix that when tensorized becomes ${\cal A}_{B_Sl_1}$.  In effect, ${A}_{B_Sl_1}$ (in matricized form) is the dual adjacency matrix\cite{Anonymous:2021:02} of the hetero-functional adjacency matrix $A_\rho$.  The presence of this matrix multiplication obfuscates (i.e. creates a lack of lucidity) as to whether one capability or another occurred when expressing the adjacency tensor element ${\cal A}_{B_Sl_1}(y_1,y_2,i_1,i_2)$.  In contrast, the matrix multiplication in Eq. \ref{eq:bigresult} does not cause the same problem.  When two capabilities succeed one another, the information associated with their physical feasibility in terms of intermediate buffers and their functional feasibility in terms of intermediate operands remains intact.  In other words, given the sequence of capabilities $e_{w_1v_1}e_{w_2v_2}$, one can immediately deduce the exchanged operands in $L_\lambda \subseteq L$ and the intermediate buffer $b_s$.  In the case of the exchanged operands, one simply needs to intersect the output-operand set of the first process with the input operand set of the second process.  In the case of the intermediate buffer, one checks if either or both of the resources are buffers.  If not, then two transportation processes followed one another and the intermediate buffer is deduced by Eq. \ref{eq:indexingtransport}.  In short, the hetero-functional adjacency matrix (or tensor) unambigously describes the sequence of two subject+verb+operand sentences whereas neither of the above interpretations of a multi-layer network do.  

\subsection{Network Descriptors}\label{sec:networkdescriptors}
In light of the commonalities and differences between hetero-functional graphs and (formal) multilayer networks, this section discusses the meaning of network descriptors in the context of hetero-functional graphs.  In this regard, the hetero-functional adjacency matrix is an adjacency matrix like any other.  Consequently, network descriptors can be calculated straightforwardly.  Furthermore, network descriptors can be applied to subsets of the graph so as to conduct a layer-by-layer analysis.  Nevertheless, that the nodes in a hetero-functional graph represent whole-sentence-capabilities means that network descriptors have the potential to provide new found meanings over formal graphs based on exclusively formal elements. 

\subsubsection{Degree Centrality} Degree centrality measures the number of edges attached to a vertex. Since a hetero-functional graph is a directed graph, there is a need to distinguish between the in-degree centrality, which measures the number of edges going into vertex, and the out-degree centrality, which measures the number of edges going out of a vertex \cite{Thompson:2020:SPG-C68}.  In the context of hetero-functional graph theory, the in-degree centrality of a vertex calculates the number of capabilities that potentially proceed the capability related to the vertex. The out-degree centrality calculates the number of capabilities that potentially succeed the vertex's capability. The higher the degree centrality of a capability, the more connected that capability is to the other capabilities in the hetero-functional graph.  It is important to recognize that because transportation capabilities receive nodes in a hetero-functional graph, they can become the most central node.  In contrast, the degree centrality of a formal graph could not reach such a conclusion because the function of transportation is tied to formal edges rather than formal nodes.  

\subsubsection{Closeness Centrality} Closeness centrality measures the average shortest path from one vertex to every other reachable vertex in the graph. In a hetero-functional graph, the meaning of closeness centrality shows how a disruption has the potential to propagate through the graph across all different types of operands \cite{Thompson:2020:SPG-C68}. This metric is especially valuable for the resilience studies of interdependent systems, where the propagation of disruption across multiple disciplines is often poorly understood. 

\subsubsection{Eigenvector Centrality} Eigenvector centrality calculates the importance of a node relative to the other nodes in the network \cite{Bonacich:2007:00}. It also includes the eigenvector centrality of the node's direct neighbors \cite{De-Domenico:2013:00}. The eigenvector centrality is specifically designed for the weighting of the in-degree of nodes in a directed network. The \emph{Katz centrality}, on the other hand, provides an approach to study the relative importance of nodes based on the out-degree \cite{De-Domenico:2013:00}. 

\subsubsection{Clustering Coefficients} Clustering coefficients describe how strongly the nodes of a network cluster together. This is performed by searching for ``triangles" or ``circles" of nodes in a network. In a directed network, these circles can appear in multiple distinct combinations of directed connections. Each of these combinations needs to be measured and counted differently. Fagiolo discussed this taxonomy and accompanying clustering coefficients \cite{Fagiolo:2007:00}. These clustering coefficients for directed networks can be directly applied to hetero-functional graphs and show which capabilities are strongly clustered together.  The definition of layers in hetero-functional graphs allows for a consistent definition and calculation of clustering coefficients within and across layers for different types of systems. When investigating a system, the clustering coefficient may show clusters of capabilities that were not yet recognized as heavily interdependent. Such information can be used to revise control structures such that clusters of capabilities are controlled by the same entity for efficiency. 

\subsubsection{Modularity} Modularity serves as a measure to study if a network can be decomposed in disjoint sets. In the hetero-functional graph theory literature, much has been published about modularity as it was a prime motivation towards the inception of the theory \cite{Farid:2007:IEM-TP00, Farid:2008:IEM-J06}.  Hetero-functional graph theory introduces the concept of the \emph{Degree-of-Freedom-based Design Structure Matrix} (or: the capability DSM) that does not only encompass the hetero-functional adjacency matrix, but extends the concept to the other elements of hetero-functional graph theory: the service model and the control model. The hetero-functional graph design structure matrix has the ability to visualize the couplings between the subsystems of an engineering system and to classify those interfaces. Note that the capability DSM can also be applied to just the hetero-functional adjacency matrix. Furthermore, the capability DSM applies to the concept of layers in a hetero-functional graph. To study the interfaces between layers, the capability DSM can adopt layers as subsystems and classify the interfaces between the layers as mentioned previously.  In conclusion, hetero-functional graphs are described by flat adjacency matrices, regardless of the number of layers in the analysis. Consequently, conventional graph theoretic network descriptors can be applied. The main difference in definition between the conventional graph theoretic application and the hetero-functional graph theoretic application is the result of the difference in the \emph{definition of the fundamental modeling elements}, the nodes and edges, in a hetero-functional graph.

\section{Conclusions and Future Work}\label{sec:conclusion}
This paper has provided a tensor-based formulation of several of the most important parts of hetero-functional graph theory.  More specifically, it discussed the system concept showing it as a generalization of formal graphs and multi-commodity networks.  It also discussed the hetero-functional adjacency matrix and its tensor-based closed form calculation.  It also discussed the hetero-functional incidence tensor and related it back to the hetero-functional adjacency matrix.   The tensor-based formulation described in this work makes a stronger tie between HFGT and its ontological foundations in MBSE.  Finally, the tensor-based formulation facilitates an understanding of the relationships between HFGT and multi-layer networks ``despite its disparate terminology and lack of consensus".  In so doing, this tensor-based treatment is likely to advance Kivela et. al's goal to discern the similarities and differences between these mathematical models in as precise a manner as possible.  

\bibliographystyle{IEEEtran}
\bibliography{LIINESLibrary,LIINESPublications}

\begin{thebibliography}{100}
\providecommand{\url}[1]{#1}
\csname url@samestyle\endcsname
\providecommand{\newblock}{\relax}
\providecommand{\bibinfo}[2]{#2}
\providecommand{\BIBentrySTDinterwordspacing}{\spaceskip=0pt\relax}
\providecommand{\BIBentryALTinterwordstretchfactor}{4}
\providecommand{\BIBentryALTinterwordspacing}{\spaceskip=\fontdimen2\font plus
\BIBentryALTinterwordstretchfactor\fontdimen3\font minus
  \fontdimen4\font\relax}
\providecommand{\BIBforeignlanguage}[2]{{%
\expandafter\ifx\csname l@#1\endcsname\relax
\typeout{** WARNING: IEEEtran.bst: No hyphenation pattern has been}%
\typeout{** loaded for the language `#1'. Using the pattern for}%
\typeout{** the default language instead.}%
\else
\language=\csname l@#1\endcsname
\fi
#2}}
\providecommand{\BIBdecl}{\relax}
\BIBdecl

\bibitem{Anonymous-NAE:2019:00}
\BIBentryALTinterwordspacing
Anonymous-NAE, ``Nae grand challenges for engineering,'' {National Academy of
  Engineering}, Tech. Rep., 2019. [Online]. Available:
  \url{http://www.engineeringchallenges.org/challenges.aspx}
\BIBentrySTDinterwordspacing

\bibitem{Park:2021:ISC-BC10}
\BIBentryALTinterwordspacing
G.-J. Park and A.~M. Farid, ``Design of large engineering systems,'' in
  \emph{Design Engineering and Science}, N.~P. Suh, M.~Cavique, and J.~Foley,
  Eds.\hskip 1em plus 0.5em minus 0.4em\relax Berlin, Heidelberg: Springer,
  2021, pp. 367--415. [Online]. Available:
  \url{https://doi.org/10.1007/978-3-030-49232-8_14}
\BIBentrySTDinterwordspacing

\bibitem{De-Weck:2011:00}
\BIBentryALTinterwordspacing
O.~L. {De Weck}, D.~Roos, and C.~L. Magee, \emph{{Engineering systems: meeting
  human needs in a complex technological world}}.\hskip 1em plus 0.5em minus
  0.4em\relax Cambridge, Mass.: MIT Press, 2011. [Online]. Available:
  \url{http://www.knovel.com/knovel2/Toc.jsp?BookID=4611
  http://mitpress-ebooks.mit.edu/product/engineering-systems}
\BIBentrySTDinterwordspacing

\bibitem{SE-Handbook-Working-Group:2015:00}
{SE Handbook Working Group}, \emph{Systems Engineering Handbook: A Guide for
  System Life Cycle Processes and Activities}.\hskip 1em plus 0.5em minus
  0.4em\relax International Council on Systems Engineering (INCOSE), 2015.

\bibitem{Weilkiens:2007:00}
T.~Weilkiens, \emph{{Systems engineering with SysML/UML modeling, analysis,
  design}}.\hskip 1em plus 0.5em minus 0.4em\relax Burlington, Mass.: Morgan
  Kaufmann, 2007.

\bibitem{Friedenthal:2011:00}
S.~Friedenthal, A.~Moore, and R.~Steiner, \emph{{A Practical Guide to SysML:
  The Systems Modeling Language}}, 2nd~ed.\hskip 1em plus 0.5em minus
  0.4em\relax Burlington, MA: Morgan Kaufmann, 2011.

\bibitem{Schoonenberg:2019:ISC-BK04}
\BIBentryALTinterwordspacing
W.~C. Schoonenberg, I.~S. Khayal, and A.~M. Farid, \emph{{A Hetero-functional
  Graph Theory for Modeling Interdependent Smart City Infrastructure}}.\hskip
  1em plus 0.5em minus 0.4em\relax Berlin, Heidelberg: Springer, 2019.
  [Online]. Available: \url{http://dx.doi.org/10.1007/978-3-319-99301-0}
\BIBentrySTDinterwordspacing

\bibitem{Kivela:2014:00}
M.~Kivel{\"a}, A.~Arenas, M.~Barthelemy, J.~P. Gleeson, Y.~Moreno, and M.~A.
  Porter, ``Multilayer networks,'' \emph{Journal of complex networks}, vol.~2,
  no.~3, pp. 203--271, 2014.

\bibitem{Farid:2006:IEM-C02}
\BIBentryALTinterwordspacing
A.~M. Farid and D.~C. McFarlane, ``{A Development of Degrees of Freedom for
  Manufacturing Systems},'' in \emph{IMS'2006: 5th International Symposium on
  Intelligent Manufacturing Systems: Agents and Virtual Worlds}, Sakarya,
  Turkey, 2006, pp. 1--6. [Online]. Available:
  \url{http://engineering.dartmouth.edu/liines/resources/Conferences/IEM-C02.pdf}
\BIBentrySTDinterwordspacing

\bibitem{Farid:2007:IEM-TP00}
\BIBentryALTinterwordspacing
A.~M. Farid, ``{Reconfigurability Measurement in Automated Manufacturing
  Systems},'' Ph.D. Dissertation, University of Cambridge Engineering
  Department Institute for Manufacturing, 2007. [Online]. Available:
  \url{http://engineering.dartmouth.edu/liines/resources/Theses/IEM-TP00.pdf}
\BIBentrySTDinterwordspacing

\bibitem{Farid:2008:IEM-J05}
\BIBentryALTinterwordspacing
A.~M. Farid and D.~C. McFarlane, ``{Production degrees of freedom as
  manufacturing system reconfiguration potential measures},'' \emph{Proceedings
  of the Institution of Mechanical Engineers, Part B (Journal of Engineering
  Manufacture) -- invited paper}, vol. 222, no. B10, pp. 1301--1314, 2008.
  [Online]. Available: \url{http://dx.doi.org/10.1243/09544054JEM1056}
\BIBentrySTDinterwordspacing

\bibitem{Farid:2008:IEM-J04}
\BIBentryALTinterwordspacing
A.~M. Farid, ``{Product Degrees of Freedom as Manufacturing System
  Reconfiguration Potential Measures},'' \emph{International Transactions on
  Systems Science and Applications -- invited paper}, vol.~4, no.~3, pp.
  227--242, 2008. [Online]. Available:
  \url{http://engineering.dartmouth.edu/liines/resources/Journals/IEM-J04.pdf}
\BIBentrySTDinterwordspacing

\bibitem{Farid:2015:ISC-J19}
\BIBentryALTinterwordspacing
------, ``{Static Resilience of Large Flexible Engineering Systems: Axiomatic
  Design Model and Measures},'' \emph{IEEE Systems Journal}, vol.~PP, no.~99,
  pp. 1--12, 2015. [Online]. Available:
  \url{http://dx.doi.org/10.1109/JSYST.2015.2428284}
\BIBentrySTDinterwordspacing

\bibitem{De-Domenico:2013:00}
M.~De~Domenico, A.~Sol{\'e}-Ribalta, E.~Cozzo, M.~Kivel{\"a}, Y.~Moreno, M.~A.
  Porter, S.~G{\'o}mez, and A.~Arenas, ``Mathematical formulation of multilayer
  networks,'' \emph{Physical Review X}, vol.~3, no.~4, p. 041022, 2013.

\bibitem{De-Domenico:2014:00}
M.~De~Domenico, A.~Sol{\'e}-Ribalta, S.~G{\'o}mez, and A.~Arenas,
  ``Navigability of interconnected networks under random failures,''
  \emph{Proceedings of the National Academy of Sciences}, vol. 111, no.~23, pp.
  8351--8356, 2014.

\bibitem{Yagan:2012:00}
\BIBentryALTinterwordspacing
O.~Ya{\u g}an and V.~Gligor, ``Analysis of complex contagions in random
  multiplex networks,'' \emph{Phys. Rev. E}, vol.~86, p. 036103, Sep 2012.
  [Online]. Available: \url{http://link.aps.org/doi/10.1103/PhysRevE.86.036103}
\BIBentrySTDinterwordspacing

\bibitem{Nicosia:2013:00}
V.~Nicosia, G.~Bianconi, V.~Latora, and M.~Barthelemy, ``Growing multiplex
  networks,'' \emph{Physical review letters}, vol. 111, no.~5, p. 058701, 2013.

\bibitem{Bianconi:2013:00}
G.~Bianconi, ``Statistical mechanics of multiplex networks: Entropy and
  overlap,'' \emph{Physical Review E}, vol.~87, no.~6, p. 062806, 2013.

\bibitem{Battiston:2014:00}
F.~Battiston, V.~Nicosia, and V.~Latora, ``Structural measures for multiplex
  networks,'' \emph{Physical Review E}, vol.~89, no.~3, p. 032804, 2014.

\bibitem{Horvat:2012:00}
E.-A. Horv{\'a}t and K.~A. Zweig, ``One-mode projection of multiplex bipartite
  graphs,'' in \emph{Proceedings of the 2012 International Conference on
  Advances in Social Networks Analysis and Mining (ASONAM 2012)}.\hskip 1em
  plus 0.5em minus 0.4em\relax IEEE Computer Society, 2012, pp. 599--606.

\bibitem{Sole-Ribalta:2013:00}
A.~Sole-Ribalta, M.~De~Domenico, N.~E. Kouvaris, A.~D{\'\i}az-Guilera,
  S.~G{\'o}mez, and A.~Arenas, ``Spectral properties of the laplacian of
  multiplex networks,'' \emph{Physical Review E}, vol.~88, no.~3, p. 032807,
  2013.

\bibitem{Cozzo:2013:00}
E.~Cozzo, R.~A. Banos, S.~Meloni, and Y.~Moreno, ``Contact-based social
  contagion in multiplex networks,'' \emph{Physical Review E}, vol.~88, no.~5,
  p. 050801, 2013.

\bibitem{Sola:2013:00}
L.~Sol{\'a}, M.~Romance, R.~Criado, J.~Flores, A.~Garc{\'\i}a~del Amo, and
  S.~Boccaletti, ``Eigenvector centrality of nodes in multiplex networks,''
  \emph{Chaos: An Interdisciplinary Journal of Nonlinear Science}, vol.~23,
  no.~3, p. 033131, 2013.

\bibitem{Pattison:1999:00}
P.~Pattison and S.~Wasserman, ``Logit models and logistic regressions for
  social networks: Ii. multivariate relations,'' \emph{British Journal of
  Mathematical and Statistical Psychology}, vol.~52, no.~2, pp. 169--193, 1999.

\bibitem{Barigozzi:2011:00}
M.~Barigozzi, G.~Fagiolo, and G.~Mangioni, ``Identifying the community
  structure of the international-trade multi-network,'' \emph{Physica A:
  statistical mechanics and its applications}, vol. 390, no.~11, pp.
  2051--2066, 2011.

\bibitem{Cai:2005:00}
D.~Cai, Z.~Shao, X.~He, X.~Yan, and J.~Han, ``Community mining from
  multi-relational networks,'' in \emph{European Conference on Principles of
  Data Mining and Knowledge Discovery}.\hskip 1em plus 0.5em minus 0.4em\relax
  Springer, 2005, pp. 445--452.

\bibitem{Harrer:2012:00}
A.~Harrer and A.~Schmidt, ``An approach for the blockmodeling in
  multi-relational networks,'' in \emph{Advances in Social Networks Analysis
  and Mining (ASONAM), 2012 IEEE/ACM International Conference on}.\hskip 1em
  plus 0.5em minus 0.4em\relax IEEE, 2012, pp. 591--598.

\bibitem{Stroele:2009:00}
V.~Stroele, J.~Oliveira, G.~Zimbrao, and J.~M. Souza, ``Mining and analyzing
  multirelational social networks,'' in \emph{Computational Science and
  Engineering, 2009. CSE'09. International Conference on}, vol.~4.\hskip 1em
  plus 0.5em minus 0.4em\relax IEEE, 2009, pp. 711--716.

\bibitem{Li:2012:00}
\BIBentryALTinterwordspacing
W.~Li, A.~Bashan, S.~V. Buldyrev, H.~E. Stanley, and S.~Havlin, ``Cascading
  failures in interdependent lattice networks: The critical role of the length
  of dependency links,'' \emph{Phys. Rev. Lett.}, vol. 108, p. 228702, May
  2012. [Online]. Available:
  \url{http://link.aps.org/doi/10.1103/PhysRevLett.108.228702}
\BIBentrySTDinterwordspacing

\bibitem{Ng:2011:00}
M.~K.-P. Ng, X.~Li, and Y.~Ye, ``Multirank: co-ranking for objects and
  relations in multi-relational data,'' in \emph{Proceedings of the 17th ACM
  SIGKDD international conference on Knowledge discovery and data
  mining}.\hskip 1em plus 0.5em minus 0.4em\relax ACM, 2011, pp. 1217--1225.

\bibitem{Brodka:2010:00}
P.~Br{\'o}dka, K.~Musial, and P.~Kazienko, ``A method for group extraction in
  complex social networks,'' \emph{Knowledge Management, Information Systems,
  E-Learning, and Sustainability Research}, pp. 238--247, 2010.

\bibitem{Brodka:2011:00}
P.~Brodka, P.~Stawiak, and P.~Kazienko, ``Shortest path discovery in the
  multi-layered social network,'' in \emph{Advances in Social Networks Analysis
  and Mining (ASONAM), 2011 International Conference on}.\hskip 1em plus 0.5em
  minus 0.4em\relax IEEE, 2011, pp. 497--501.

\bibitem{Brodka:2012:00}
P.~Br{\'o}dka, P.~Kazienko, K.~Musia{\l}, and K.~Skibicki, ``Analysis of
  neighbourhoods in multi-layered dynamic social networks,''
  \emph{International Journal of Computational Intelligence Systems}, vol.~5,
  no.~3, pp. 582--596, 2012.

\bibitem{Berlingerio:2011:00}
M.~Berlingerio, M.~Coscia, F.~Giannotti, A.~Monreale, and D.~Pedreschi, ``The
  pursuit of hubbiness: analysis of hubs in large multidimensional networks,''
  \emph{Journal of Computational Science}, vol.~2, no.~3, pp. 223--237, 2011.

\bibitem{Berlingerio:2013:00}
M.~Berlingerio, F.~Pinelli, and F.~Calabrese, ``Abacus: frequent pattern
  mining-based community discovery in multidimensional networks,'' \emph{Data
  Mining and Knowledge Discovery}, vol.~27, no.~3, pp. 294--320, 2013.

\bibitem{Berlingerio:2013:01}
M.~Berlingerio, M.~Coscia, F.~Giannotti, A.~Monreale, and D.~Pedreschi,
  ``Multidimensional networks: foundations of structural analysis,''
  \emph{World Wide Web}, vol.~16, no. 5-6, pp. 567--593, 2013.

\bibitem{Tang:2012:00}
L.~Tang, X.~Wang, and H.~Liu, ``Community detection via heterogeneous
  interaction analysis,'' \emph{Data mining and knowledge discovery}, vol.~25,
  no.~1, pp. 1--33, 2012.

\bibitem{Barrett:2012:00}
\BIBentryALTinterwordspacing
C.~Barrett, K.~Channakeshava, F.~Huang, J.~Kim, A.~Marathe, M.~V. Marathe,
  G.~Pei, S.~Saha, B.~S.~P. Subbiah, and A.~K.~S. Vullikanti, ``Human initiated
  cascading failures in societal infrastructures,'' \emph{PLoS ONE}, vol.~7,
  no.~10, pp. 1--20, 10 2012. [Online]. Available:
  \url{http://dx.doi.org/10.1371%2Fjournal.pone.0045406}
\BIBentrySTDinterwordspacing

\bibitem{Kazienko:2011:00}
P.~Kazienko, K.~Musial, and T.~Kajdanowicz, ``Multidimensional social network
  in the social recommender system,'' \emph{IEEE Transactions on Systems, Man,
  and Cybernetics-Part A: Systems and Humans}, vol.~41, no.~4, pp. 746--759,
  2011.

\bibitem{Coscia:2013:00}
M.~Coscia, G.~Rossetti, D.~Pennacchioli, D.~Ceccarelli, and F.~Giannotti,
  ````you know because i know'': A multidimensional network approach to human
  resources problem,'' in \emph{Advances in Social Networks Analysis and Mining
  (ASONAM), 2013 IEEE/ACM International Conference on}.\hskip 1em plus 0.5em
  minus 0.4em\relax IEEE, 2013, pp. 434--441.

\bibitem{Kazienko:2011:01}
P.~Kazienko, K.~Musial, E.~Kukla, T.~Kajdanowicz, and P.~Br{\'o}dka,
  ``Multidimensional social network: model and analysis,'' \emph{Computational
  Collective Intelligence. Technologies and Applications}, pp. 378--387, 2011.

\bibitem{Mucha:2010:00}
P.~J. Mucha, T.~Richardson, K.~Macon, M.~A. Porter, and J.-P. Onnela,
  ``Community structure in time-dependent, multiscale, and multiplex
  networks,'' \emph{science}, vol. 328, no. 5980, pp. 876--878, 2010.

\bibitem{Carchiolo:2011:00}
V.~Carchiolo, A.~Longheu, M.~Malgeri, and G.~Mangioni, ``Communities unfolding
  in multislice networks,'' in \emph{Complex Networks}.\hskip 1em plus 0.5em
  minus 0.4em\relax Springer, 2011, pp. 187--195.

\bibitem{Bassett:2013:00}
D.~S. Bassett, M.~A. Porter, N.~F. Wymbs, S.~T. Grafton, J.~M. Carlson, and
  P.~J. Mucha, ``Robust detection of dynamic community structure in networks,''
  \emph{Chaos: An Interdisciplinary Journal of Nonlinear Science}, vol.~23,
  no.~1, p. 013142, 2013.

\bibitem{Irving:2012:00}
D.~Irving and F.~Sorrentino, ``Synchronization of dynamical hypernetworks:
  Dimensionality reduction through simultaneous block-diagonalization of
  matrices,'' \emph{Physical Review E}, vol.~86, no.~5, p. 056102, 2012.

\bibitem{Sorrentino:2012:00}
F.~Sorrentino, ``Synchronization of hypernetworks of coupled dynamical
  systems,'' \emph{New Journal of Physics}, vol.~14, no.~3, p. 033035, 2012.

\bibitem{Funk:2010:00}
S.~Funk and V.~A. Jansen, ``Interacting epidemics on overlay networks,''
  \emph{Physical Review E}, vol.~81, no.~3, p. 036118, 2010.

\bibitem{Marceau:2011:00}
V.~Marceau, P.-A. No{\"e}l, L.~H{\'e}bert-Dufresne, A.~Allard, and L.~J.
  Dub{\'e}, ``Modeling the dynamical interaction between epidemics on overlay
  networks,'' \emph{Physical Review E}, vol.~84, no.~2, p. 026105, 2011.

\bibitem{Wei:2012:00}
X.~Wei, N.~Valler, B.~A. Prakash, I.~Neamtiu, M.~Faloutsos, and C.~Faloutsos,
  ``Competing memes propagation on networks: a case study of composite
  networks,'' \emph{ACM SIGCOMM Computer Communication Review}, vol.~42, no.~5,
  pp. 5--12, 2012.

\bibitem{Rocklin:2013:00}
M.~Rocklin and A.~Pinar, ``On clustering on graphs with multiple edge types,''
  \emph{Internet Mathematics}, vol.~9, no.~1, pp. 82--112, 2013.

\bibitem{Hindes:2013:00}
J.~Hindes, S.~Singh, C.~R. Myers, and D.~J. Schneider, ``Epidemic fronts in
  complex networks with metapopulation structure,'' \emph{Physical Review E},
  vol.~88, no.~1, p. 012809, 2013.

\bibitem{Baxter:2012:01}
G.~Baxter, S.~Dorogovtsev, A.~Goltsev, and J.~Mendes, ``Avalanche collapse of
  interdependent networks,'' \emph{Physical review letters}, vol. 109, no.~24,
  p. 248701, 2012.

\bibitem{Gomez-Gardenes:2012:00}
J.~G{\'o}mez-Garde{\~n}es, I.~Reinares, A.~Arenas, and L.~M. Flor{\'\i}a,
  ``Evolution of cooperation in multiplex networks,'' \emph{Scientific
  reports}, vol.~2, 2012.

\bibitem{Barigozzi:2010:00}
\BIBentryALTinterwordspacing
M.~Barigozzi, G.~Fagiolo, and D.~Garlaschelli, ``Multinetwork of international
  trade: A commodity-specific analysis,'' \emph{Phys. Rev. E}, vol.~81, p.
  046104, Apr 2010. [Online]. Available:
  \url{http://link.aps.org/doi/10.1103/PhysRevE.81.046104}
\BIBentrySTDinterwordspacing

\bibitem{Cellai:2013:00}
\BIBentryALTinterwordspacing
D.~Cellai, E.~L\'opez, J.~Zhou, J.~P. Gleeson, and G.~Bianconi, ``Percolation
  in multiplex networks with overlap,'' \emph{Phys. Rev. E}, vol.~88, p.
  052811, Nov 2013. [Online]. Available:
  \url{http://link.aps.org/doi/10.1103/PhysRevE.88.052811}
\BIBentrySTDinterwordspacing

\bibitem{Brummitt:2012:00}
\BIBentryALTinterwordspacing
C.~D. Brummitt, K.-M. Lee, and K.-I. Goh, ``Multiplexity-facilitated cascades
  in networks,'' \emph{Phys. Rev. E}, vol.~85, p. 045102, Apr 2012. [Online].
  Available: \url{http://link.aps.org/doi/10.1103/PhysRevE.85.045102}
\BIBentrySTDinterwordspacing

\bibitem{Mucha:2010:01}
\BIBentryALTinterwordspacing
P.~J. Mucha, T.~Richardson, K.~Macon, M.~A. Porter, and J.-P. Onnela,
  ``Community structure in time-dependent, multiscale, and multiplex
  networks,'' \emph{Science}, vol. 328, no. 5980, pp. 876--878, 2010. [Online].
  Available: \url{http://science.sciencemag.org/content/328/5980/876}
\BIBentrySTDinterwordspacing

\bibitem{Wasserman:1994:00}
S.~Wasserman and K.~Faust, \emph{Social network analysis: Methods and
  applications}.\hskip 1em plus 0.5em minus 0.4em\relax Cambridge university
  press, 1994, vol.~8.

\bibitem{Min:2013:00}
B.~Min and K.~Goh, ``Layer-crossing overhead and information spreading in
  multiplex social networks,'' \emph{seed}, vol.~21, no. T22, p. T12, 2013.

\bibitem{Lee:2012:02}
K.-M. Lee, J.~Y. Kim, W.-k. Cho, K.-I. Goh, and I.~Kim, ``Correlated
  multiplexity and connectivity of multiplex random networks,'' \emph{New
  Journal of Physics}, vol.~14, no.~3, p. 033027, 2012.

\bibitem{Min:2014:00}
B.~Min, S.~Do~Yi, K.-M. Lee, and K.-I. Goh, ``Network robustness of multiplex
  networks with interlayer degree correlations,'' \emph{Physical Review E},
  vol.~89, no.~4, p. 042811, 2014.

\bibitem{Cozzo:2013:01}
E.~Cozzo, R.~A. Banos, S.~Meloni, and Y.~Moreno, ``Contact-based social
  contagion in multiplex networks,'' \emph{Physical Review E}, vol.~88, no.~5,
  p. 050801, 2013.

\bibitem{Allard:2009:00}
\BIBentryALTinterwordspacing
A.~Allard, P.-A. No\"el, L.~J. Dub\'e, and B.~Pourbohloul, ``Heterogeneous bond
  percolation on multitype networks with an application to epidemic dynamics,''
  \emph{Phys. Rev. E}, vol.~79, p. 036113, Mar 2009. [Online]. Available:
  \url{http://link.aps.org/doi/10.1103/PhysRevE.79.036113}
\BIBentrySTDinterwordspacing

\bibitem{Bashan:2013:00}
\BIBentryALTinterwordspacing
A.~Bashan, Y.~Berezin, S.~V. Buldyrev, and S.~Havlin, ``The extreme
  vulnerability of interdependent spatially embedded networks,'' \emph{Nat
  Phys}, vol.~9, no.~10, pp. 667--672, 10 2013. [Online]. Available:
  \url{http://dx.doi.org/10.1038/nphys2727}
\BIBentrySTDinterwordspacing

\bibitem{Buldyrev:2010:00}
S.~V. Buldyrev, R.~Parshani, G.~Paul, H.~E. Stanley, and S.~Havlin,
  ``Catastrophic cascade of failures in interdependent networks,''
  \emph{Nature}, vol. 464, no. 7291, p. 1025, 2010.

\bibitem{Cardillo:2012:00}
A.~Cardillo, M.~Zanin, J.~G{\'o}mez-Gardenes, M.~Romance, A.~J.~G. del Amo, and
  S.~Boccaletti, ``Modeling the multi-layer nature of the european air
  transport network: Resilience and passengers re-scheduling under random
  failures,'' \emph{arXiv preprint arXiv:1211.6839}, 2012.

\bibitem{Dickison:2012:00}
M.~Dickison, S.~Havlin, and H.~E. Stanley, ``Epidemics on interconnected
  networks,'' \emph{Physical Review E}, vol.~85, no.~6, p. 066109, 2012.

\bibitem{Donges:2011:00}
J.~F. Donges, H.~C. Schultz, N.~Marwan, Y.~Zou, and J.~Kurths, ``Investigating
  the topology of interacting networks,'' \emph{The European Physical Journal
  B}, vol.~84, no.~4, pp. 635--651, 2011.

\bibitem{Lazega:2008:00}
E.~Lazega, M.-T. Jourda, L.~Mounier, and R.~Stofer, ``Catching up with big fish
  in the big pond? multi-level network analysis through linked design,''
  \emph{Social Networks}, vol.~30, no.~2, pp. 159--176, 2008.

\bibitem{Leicht:2009:00}
E.~A. {Leicht} and R.~M. {D'Souza}, ``{Percolation on interacting networks},''
  \emph{ArXiv e-prints}, Jul. 2009.

\bibitem{Louzada:2013:00}
V.~Louzada, N.~Ara{\'u}jo, J.~Andrade~Jr, and H.~Herrmann, ``Breathing
  synchronization in interconnected networks,'' \emph{arXiv preprint
  arXiv:1304.5177}, 2013.

\bibitem{Martin-Hernandez:2013:00}
J.~Martin-Hernandez, H.~Wang, P.~Van~Mieghem, and G.~D'Agostino, ``On
  synchronization of interdependent networks,'' \emph{arXiv preprint
  arXiv:1304.4731}, 2013.

\bibitem{Parshani:2010:00}
\BIBentryALTinterwordspacing
R.~Parshani, S.~V. Buldyrev, and S.~Havlin, ``Interdependent networks: Reducing
  the coupling strength leads to a change from a first to second order
  percolation transition,'' \emph{Phys. Rev. Lett.}, vol. 105, p. 048701, Jul
  2010. [Online]. Available:
  \url{http://link.aps.org/doi/10.1103/PhysRevLett.105.048701}
\BIBentrySTDinterwordspacing

\bibitem{Sahneh:2013:00}
F.~D. Sahneh, C.~Scoglio, and F.~N. Chowdhury, ``Effect of coupling on the
  epidemic threshold in interconnected complex networks: A spectral analysis,''
  in \emph{American Control Conference (ACC), 2013}.\hskip 1em plus 0.5em minus
  0.4em\relax IEEE, 2013, pp. 2307--2312.

\bibitem{Saumell-Mendiola:2012:00}
\BIBentryALTinterwordspacing
A.~Saumell-Mendiola, M.~A. Serrano, and M.~Bogu\~n\'a, ``Epidemic spreading on
  interconnected networks,'' \emph{Phys. Rev. E}, vol.~86, p. 026106, Aug 2012.
  [Online]. Available: \url{http://link.aps.org/doi/10.1103/PhysRevE.86.026106}
\BIBentrySTDinterwordspacing

\bibitem{Sun:2009:00}
Y.~Sun, Y.~Yu, and J.~Han, ``Ranking-based clustering of heterogeneous
  information networks with star network schema,'' in \emph{Proceedings of the
  15th ACM SIGKDD international conference on Knowledge discovery and data
  mining}.\hskip 1em plus 0.5em minus 0.4em\relax ACM, 2009, pp. 797--806.

\bibitem{Vazquez:2006:00}
A.~Vazquez, ``Spreading dynamics on heterogeneous populations: multitype
  network approach,'' \emph{Physical Review E}, vol.~74, no.~6, p. 066114,
  2006.

\bibitem{Wang:2013:00}
C.~Wang, Z.~Lu, and Y.~Qiao, ``A consideration of the wind power benefits in
  day-ahead scheduling of wind-coal intensive power systems,'' \emph{IEEE
  Trans. Power Syst.}, vol.~28, no.~1, pp. 236--245, Feb 2013.

\bibitem{Zhou:2007:00}
\BIBentryALTinterwordspacing
J.~Zhou, L.~Xiang, and Z.~Liu, ``{Global synchronization in general complex
  delayed dynamical networks and its applications},'' \emph{Physica A:
  Statistical Mechanics and its Applications}, vol. 385, no.~2, pp. 729--742,
  Nov. 2007. [Online]. Available:
  \url{http://linkinghub.elsevier.com/retrieve/pii/S0378437107007637}
\BIBentrySTDinterwordspacing

\bibitem{Zhou:2013:00}
D.~Zhou, J.~Gao, H.~E. Stanley, and S.~Havlin, ``Percolation of partially
  interdependent scale-free networks,'' \emph{Physical Review E}, vol.~87,
  no.~5, p. 052812, 2013.

\bibitem{Gao:2011:00}
L.~Gao, J.~Yang, H.~Zhang, B.~Zhang, and D.~Qin, ``Flowinfra: A fault-resilient
  scalable infrastructure for network-wide flow level measurement,'' \emph{2011
  13th Asia-Pacific Network Operations and Management Symposium}, p. KICS KNOM;
  IEICE ICM, Sep 2011.

\bibitem{Lee:2012:00}
K.-M. Lee, J.~Y. Kim, W.-k. Cho, K.-I. Goh, and I.~Kim, ``Correlated
  multiplexity and connectivity of multiplex random networks,'' \emph{New
  Journal of Physics}, vol.~14, no.~3, p. 033027, 2012.

\bibitem{Cozzo:2012:00}
\BIBentryALTinterwordspacing
E.~Cozzo, A.~Arenas, and Y.~Moreno, ``Stability of boolean multilevel
  networks,'' \emph{Phys. Rev. E}, vol.~86, p. 036115, Sep 2012. [Online].
  Available: \url{http://link.aps.org/doi/10.1103/PhysRevE.86.036115}
\BIBentrySTDinterwordspacing

\bibitem{Criado:2012:00}
R.~Criado, J.~Flores, A.~Garc{\'\i}a~del Amo, J.~G{\'o}mez-Garde{\~n}es, and
  M.~Romance, ``A mathematical model for networks with structures in the
  mesoscale,'' \emph{International Journal of Computer Mathematics}, vol.~89,
  no.~3, pp. 291--309, 2012.

\bibitem{Xu:2011:00}
Y.~Xu and W.~Liu, ``{Novel Multiagent Based Load Restoration Algorithm for
  Microgrids},'' \emph{Smart Grid, IEEE Transactions on}, vol.~2, no.~1, pp.
  152--161, 2011.

\bibitem{Yagan:2013:00}
O.~Yagan, D.~Qian, J.~Zhang, and D.~Cochran, ``Conjoining speeds up information
  diffusion in overlaying social-physical networks,'' \emph{IEEE Journal on
  Selected Areas in Communications}, vol.~31, no.~6, pp. 1038--1048, 2013.

\bibitem{Carley:2001:00}
K.~M. Carley and V.~Hill, ``Structural change and learning within
  organizations,'' \emph{Dynamics of organizations: Computational modeling and
  organizational theories}, pp. 63--92, 2001.

\bibitem{Carley:2007:00}
K.~M. Carley, J.~Diesner, J.~Reminga, and M.~Tsvetovat, ``Toward an
  interoperable dynamic network analysis toolkit,'' \emph{Decision Support
  Systems}, vol.~43, no.~4, pp. 1324--1347, 2007.

\bibitem{Davis:2011:00}
D.~Davis, R.~Lichtenwalter, and N.~V. Chawla, ``Multi-relational link
  prediction in heterogeneous information networks,'' in \emph{Advances in
  Social Networks Analysis and Mining (ASONAM), 2011 International Conference
  on}.\hskip 1em plus 0.5em minus 0.4em\relax IEEE, 2011, pp. 281--288.

\bibitem{Sun:2011:00}
Y.~Sun, J.~Han, X.~Yan, P.~S. Yu, and T.~Wu, ``Pathsim: Meta path-based top-k
  similarity search in heterogeneous information networks,'' \emph{Proceedings
  of the VLDB Endowment}, vol.~4, no.~11, pp. 992--1003, 2011.

\bibitem{Sun:2012:00}
Y.~Sun, ``Mining heterogeneous information networks,'' Ph.D. dissertation,
  University of Illinois at Urbana-Champaign, 2012.

\bibitem{Sun:2013:00}
W.-Q. Sun, C.-M. Wang, P.~Song, and Y.~Zhang, ``Flexible load shedding strategy
  considering real-time dynamic thermal line rating,'' \emph{IET Generation,
  Transmission \& Distribution}, vol.~7, no.~2, pp. 130--137, Feb 2013.

\bibitem{Tsvetovat:2004:00}
\BIBentryALTinterwordspacing
M.~Tsvetovat, J.~Reminga, and K.~M. Carley, ``Dynetml: Interchange format for
  rich social network data,'' \emph{SSRN}, 2004. [Online]. Available:
  \url{http://dx.doi.org/10.2139/ssrn.2729286}
\BIBentrySTDinterwordspacing

\bibitem{Farid:2016:ISC-BC06}
\BIBentryALTinterwordspacing
A.~M. Farid, ``An engineering systems introduction to axiomatic design,'' in
  \emph{Axiomatic Design in Large Systems: Complex Products, Buildings \&
  Manufacturing Systems}, A.~M. Farid and N.~P. Suh, Eds.\hskip 1em plus 0.5em
  minus 0.4em\relax Berlin, Heidelberg: Springer, 2016, ch.~1, pp. 1--47.
  [Online]. Available: \url{http://dx.doi.org/10.1007/978-3-319-32388-6}
\BIBentrySTDinterwordspacing

\bibitem{Guizzardi:2007:00}
G.~Guizzardi, ``On ontology, ontologies, conceptualizations, modeling
  languages, and (meta) models,'' \emph{Frontiers in artificial intelligence
  and applications}, vol. 155, p.~18, 2007.

\bibitem{Guizzardi:2005:00}
------, \emph{Ontological foundations for structural conceptual models}.\hskip
  1em plus 0.5em minus 0.4em\relax CTIT, Centre for Telematics and Information
  Technology, 2005.

\bibitem{Crawley:2015:00}
E.~Crawley, B.~Cameron, and D.~Selva, \emph{System Architecture: Strategy and
  Product Development for Complex Systems}.\hskip 1em plus 0.5em minus
  0.4em\relax Upper Saddle River, N.J.: Prentice Hall Press, 2015.

\bibitem{Barabasi:2016:00}
A.-L. Barab{\'a}si \emph{et~al.}, \emph{Network science}.\hskip 1em plus 0.5em
  minus 0.4em\relax Cambridge university press, 2016.

\bibitem{Newman:2009:00}
\BIBentryALTinterwordspacing
M.~Newman, \emph{{Networks: An Introduction}}.\hskip 1em plus 0.5em minus
  0.4em\relax Oxford, United Kingdom: Oxford University Press, 2009. [Online].
  Available: \url{http://books.google.ae/books?id=LrFaU4XCsUoC}
\BIBentrySTDinterwordspacing

\bibitem{Thompson:2020:SPG-C68}
\BIBentryALTinterwordspacing
D.~Thompson, W.~C. Schoonenberg, and A.~M. Farid, ``{A Hetero-functional Graph
  Analysis of Electric Power System Structural Resilience},'' in \emph{IEEE
  Innovative Smart Grid Technologies Conference North America}, Washington, DC,
  United states, 2020, pp. 1--5. [Online]. Available:
  \url{http://dx.doi.org/10.1109/ISGT45199.2020.9087732}
\BIBentrySTDinterwordspacing

\bibitem{Thompson:2021:SPG-J46}
\BIBentryALTinterwordspacing
------, ``{A Hetero-functional Graph Resilience Analysis of the Future American
  Electric Power System},'' \emph{IEEE Access}, vol.~9, pp. 68\,837--68\,848,
  2021. [Online]. Available: \url{https://doi.org/10.1109/ACCESS.2021.3077856}
\BIBentrySTDinterwordspacing

\bibitem{Buede:2009:00}
D.~M. Buede, \emph{{The engineering design of systems: models and methods}},
  2nd~ed.\hskip 1em plus 0.5em minus 0.4em\relax Hoboken, N.J.: John Wiley \&
  Sons, 2009.

\bibitem{Kossiakoff:2003:00}
\BIBentryALTinterwordspacing
A.~Kossiakoff, W.~N. Sweet, and {Knovel (Firm)}, \emph{{Systems engineering
  principles and practice}}.\hskip 1em plus 0.5em minus 0.4em\relax Hoboken,
  N.J.: Wiley-Interscience, 2003. [Online]. Available:
  \url{http://www.knovel.com/knovel2/Toc.jsp?BookID=1430}
\BIBentrySTDinterwordspacing

\bibitem{Farid:2016:ISC-BK03}
\BIBentryALTinterwordspacing
A.~M. Farid and N.~P. Suh, \emph{Axiomatic Design in Large Systems: Complex
  Products, Buildings and Manufacturing Systems}.\hskip 1em plus 0.5em minus
  0.4em\relax Berlin, Heidelberg: Springer, 2016. [Online]. Available:
  \url{http://dx.doi.org/10.1007/978-3-319-32388-6}
\BIBentrySTDinterwordspacing

\bibitem{Hoyle:1998:00}
\BIBentryALTinterwordspacing
D.~Hoyle, \emph{{ISO 9000 pocket guide}}.\hskip 1em plus 0.5em minus
  0.4em\relax Oxford ; Boston: Butterworth-Heinemann, 1998. [Online].
  Available: \url{http://www.loc.gov/catdir/toc/els033/99163006.html}
\BIBentrySTDinterwordspacing

\bibitem{Farid:2013:IEM-C30}
\BIBentryALTinterwordspacing
A.~M. Farid, ``{An Axiomatic Design Approach to Non-Assembled Production Path
  Enumeration in Reconfigurable Manufacturing Systems},'' in \emph{2013 IEEE
  International Conference on Systems Man and Cybernetics}, Manchester, UK,
  2013, pp. 1--8. [Online]. Available:
  \url{http://dx.doi.org/10.1109/SMC.2013.659}
\BIBentrySTDinterwordspacing

\bibitem{Farid:2015:IEM-J23}
\BIBentryALTinterwordspacing
A.~M. Farid and L.~Ribeiro, ``{An Axiomatic Design of a Multi-Agent
  Reconfigurable Mechatronic System Architecture},'' \emph{IEEE Transactions on
  Industrial Informatics}, vol.~11, no.~5, pp. 1142--1155, 2015. [Online].
  Available: \url{http://dx.doi.org/10.1109/TII.2015.2470528}
\BIBentrySTDinterwordspacing

\bibitem{Farid:2016:ETS-J27}
\BIBentryALTinterwordspacing
A.~M. Farid, ``{A Hybrid Dynamic System Model for Multi-Modal Transportation
  Electrification},'' \emph{IEEE Transactions on Control System Technology},
  vol.~PP, no.~99, pp. 1--12, 2016. [Online]. Available:
  \url{http://dx.doi.org/10.1109/TCST.2016.2579602}
\BIBentrySTDinterwordspacing

\bibitem{Farid:2016:ETS-BC05}
\BIBentryALTinterwordspacing
------, ``Electrified transportation system performance: Conventional vs.
  online electric vehicles,'' in \emph{The On-line Electric Vehicle: Wireless
  Electric Ground Transportation Systems}, N.~P. Suh and D.~H. Cho, Eds.\hskip
  1em plus 0.5em minus 0.4em\relax Berlin, Heidelberg: Springer, 2017, ch.~20,
  pp. 279--313. [Online]. Available:
  \url{http://engineering.dartmouth.edu/liines/resources/Books/TES-BC05.pdf}
\BIBentrySTDinterwordspacing

\bibitem{Hu:1963:00}
T.~C. Hu, ``Multi-commodity network flows,'' \emph{Operations research},
  vol.~11, no.~3, pp. 344--360, 1963.

\bibitem{Okamura:1983:00}
H.~Okamura, ``Multicommodity flows in graphs,'' \emph{Discrete Applied
  Mathematics}, vol.~6, no.~1, pp. 55--62, 1983.

\bibitem{Ahuja:1988:00}
R.~K. Ahuja, T.~L. Magnanti, and J.~B. Orlin, \emph{Network flows: Theory,
  Algorithms, and Applications}.\hskip 1em plus 0.5em minus 0.4em\relax
  Cambridge, Mass.: Alfred P. Sloan School of Management,
  Massachusetts~{\ldots}, 1988.

\bibitem{Callaway:2000:00}
\BIBentryALTinterwordspacing
D.~S. Callaway, M.~E.~J. Newman, S.~H. Strogatz, and D.~J. Watts, ``Network
  robustness and fragility: Percolation on random graphs,'' \emph{Phys. Rev.
  Lett.}, vol.~85, pp. 5468--5471, Dec 2000. [Online]. Available:
  \url{http://link.aps.org/doi/10.1103/PhysRevLett.85.5468}
\BIBentrySTDinterwordspacing

\bibitem{Newman:2003:00}
M.~E. Newman, ``The structure and function of complex networks,'' \emph{SIAM
  review}, vol.~45, no.~2, pp. 167--256, 2003.

\bibitem{Holme:2012:00}
P.~Holme and J.~Saram{\"a}ki, ``Temporal networks,'' \emph{Physics reports},
  vol. 519, no.~3, pp. 97--125, 2012.

\bibitem{Farid:2014:ISC-C37}
\BIBentryALTinterwordspacing
A.~M. Farid, ``{Static Resilience of Large Flexible Engineering Systems: Part I
  -- Axiomatic Design Model},'' in \emph{4th International Engineering Systems
  Symposium}.\hskip 1em plus 0.5em minus 0.4em\relax Hoboken, N.J.: Stevens
  Institute of Technology, 2014, pp. 1--8. [Online]. Available:
  \url{http://engineering.dartmouth.edu/liines/resources/Conferences/IES-C37.pdf}
\BIBentrySTDinterwordspacing

\bibitem{Farid:2015:SPG-J17}
\BIBentryALTinterwordspacing
------, ``{Multi-Agent System Design Principles for Resilient Coordination and
  Control of Future Power Systems},'' \emph{Intelligent Industrial Systems},
  vol.~1, no.~3, pp. 255--269, 2015. [Online]. Available:
  \url{http://dx.doi.org/10.1007/s40903-015-0013-x}
\BIBentrySTDinterwordspacing

\bibitem{Viswanath:2013:ETS-J08}
\BIBentryALTinterwordspacing
A.~Viswanath, E.~E.~S. Baca, and A.~M. Farid, ``{An Axiomatic Design Approach
  to Passenger Itinerary Enumeration in Reconfigurable Transportation
  Systems},'' \emph{IEEE Transactions on Intelligent Transportation Systems},
  vol.~15, no.~3, pp. 915 -- 924, 2014. [Online]. Available:
  \url{http://dx.doi.org/10.1109/TITS.2013.2293340}
\BIBentrySTDinterwordspacing

\bibitem{Schoonenberg:2017:IEM-J34}
\BIBentryALTinterwordspacing
W.~C. Schoonenberg and A.~M. Farid, ``{A Dynamic Model for the Energy
  Management of Microgrid-Enabled Production Systems},'' \emph{Journal of
  Cleaner Production}, vol.~1, no.~1, pp. 1--10, 2017. [Online]. Available:
  \url{https://dx.doi.org/10.1016/j.jclepro.2017.06.119}
\BIBentrySTDinterwordspacing

\bibitem{Thompson:2022:ISC-C80}
D.~Thompson and A.~M. Farid, ``Reconciling formal, multi-layer, and
  hetero-functional graphs with the hetero-functional incidence tensor,'' in
  \emph{IEEE Systems of Systems Engineering Conference}, Rochester, NY, 2022,
  pp. 1--6.

\bibitem{Farid:2017:IEM-J13}
\BIBentryALTinterwordspacing
A.~M. Farid, ``{Measures of Reconfigurability and Its Key Characteristics in
  Intelligent Manufacturing Systems},'' \emph{Journal of Intelligent
  Manufacturing}, vol.~28, no.~2, pp. 353--369, 2017. [Online]. Available:
  \url{http://dx.doi.org/10.1007/s10845-014-0983-7}
\BIBentrySTDinterwordspacing

\bibitem{Schoonenberg:2022:ISC-J50}
\BIBentryALTinterwordspacing
W.~C. Schoonenberg and A.~M. Farid, ``{Hetero-functional Network Minimum Cost
  Flow Optimization},'' \emph{Sustainable Energy Grids and Networks (in
  press)}, vol.~31, no. 100749, pp. 1--18, 2022. [Online]. Available:
  \url{https://arxiv.org/abs/2104.00504}
\BIBentrySTDinterwordspacing

\bibitem{Rowell:1997:00}
D.~Rowell and D.~N. Wormley, \emph{{System dynamics: an introduction}}.\hskip
  1em plus 0.5em minus 0.4em\relax Upper Saddle River, NJ: Prentice Hall, 1997.

\bibitem{Karnopp:1990:00}
\BIBentryALTinterwordspacing
D.~Karnopp, D.~L. Margolis, and R.~C. Rosenberg, \emph{{System dynamics: a
  unified approach}}, 2nd~ed.\hskip 1em plus 0.5em minus 0.4em\relax New York:
  Wiley, 1990. [Online]. Available:
  \url{http://www.loc.gov/catdir/enhancements/fy0650/90012110-t.html}
\BIBentrySTDinterwordspacing

\bibitem{Anonymous:2021:02}
\BIBentryALTinterwordspacing
Anonymous, ``Dual graph,'' Wikipedia, Tech. Rep., 2021. [Online]. Available:
  \url{https://en.wikipedia.org/wiki/Dual_graph}
\BIBentrySTDinterwordspacing

\bibitem{Bonacich:2007:00}
P.~Bonacich, ``Some unique properties of eigenvector centrality,'' \emph{Social
  networks}, vol.~29, no.~4, pp. 555--564, 2007.

\bibitem{Fagiolo:2007:00}
G.~Fagiolo, ``Clustering in complex directed networks,'' \emph{Physical Review
  E}, vol.~76, no.~2, p. 026107, 2007.

\bibitem{Farid:2008:IEM-J06}
\BIBentryALTinterwordspacing
A.~M. Farid, ``{Facilitating ease of system reconfiguration through measures of
  manufacturing modularity},'' \emph{Proceedings of the Institution of
  Mechanical Engineers, Part B (Journal of Engineering Manufacture) -- invited
  paper}, vol. 222, no. B10, pp. 1275--1288, 2008. [Online]. Available:
  \url{http://dx.doi.org/10.1243/09544054JEM1055}
\BIBentrySTDinterwordspacing

\bibitem{Anonymous:2021:00}
\BIBentryALTinterwordspacing
Anonymous, ``Cartesian product,'' Wikipedia, Tech. Rep., 2021. [Online].
  Available:
  \url{https://en.wikipedia.org/wiki/Cartesian_product#:~:text=In%20mathematics%2C%20specifically%20set%20theory,and%20a%20set%20of%20columns.}
\BIBentrySTDinterwordspacing

\bibitem{Fischer:1971:00}
M.~J. Fischer and A.~R. Meyer, ``Boolean matrix multiplication and transitive
  closure,'' in \emph{Switching and Automata Theory, 1971., 12th Annual
  Symposium on}.\hskip 1em plus 0.5em minus 0.4em\relax IEEE, 1971, pp.
  129--131.

\bibitem{Anonymous:2021:01}
\BIBentryALTinterwordspacing
Anonymous, ``Kronecker delta,'' Wikipedia, Tech. Rep., 2021. [Online].
  Available: \url{https://en.wikipedia.org/wiki/Kronecker_delta}
\BIBentrySTDinterwordspacing

\bibitem{Golub:1996:00}
G.~Golub and C.~{Van Loan}, \emph{Matrix Computations}.\hskip 1em plus 0.5em
  minus 0.4em\relax Johns Hopkins University Press, 1996.

\bibitem{Kolda:2009:00}
T.~G. Kolda and B.~W. Bader, ``Tensor decompositions and applications,''
  \emph{SIAM review}, vol.~51, no.~3, pp. 455--500, 2009.

\bibitem{Kolda:2006:00}
T.~G. Kolda, ``Multilinear operators for higher-order decompositions.'' Sandia
  National Laboratories, Tech. Rep., 2006.

\bibitem{Pan:2014:00}
R.~Pan, ``Tensor transpose and its properties,'' \emph{arXiv preprint
  arXiv:1411.1503}, 2014.

\end{thebibliography}
\newpage

\appendix
\nomenclature[B]{$R$}{Set of System Resources}
\nomenclature[B]{$M$}{Set of Transformation Resources}
\nomenclature[B]{$B$}{Set of Independent Buffers}
\nomenclature[B]{$H$}{Set of Transportation Resources}
\nomenclature[C]{$r$}{Resource in set of system resources $R$}
\nomenclature[C]{$m$}{Machine in set of machines $M$}
\nomenclature[C]{$b$}{Independent buffer in set of independent buffers $B$}
\nomenclature[C]{$h$}{Transporter in set of transporters $H$}
\nomenclature[B]{$P$}{Set of System Processes}
\nomenclature[B]{$P_\mu$}{Set of Transformation Processes}
\nomenclature[B]{$P_\eta$}{Set of Transportation Processes}
\nomenclature[B]{$B_S$}{Set of Buffers}
\nomenclature[C]{$p_{\mu j}$}{Transformation process $j$ in set of transformation processes $P_\mu$}
\nomenclature[D]{$j$}{Index of transformation process $p_{\mu j}$ in set of transformation processes $P_\mu$}
\nomenclature[F]{\Cross}{Cartesian Product}
\nomenclature[C]{$p_{\eta u}$}{Transportation process $u$ in set of transportation processes $P_\eta$}
\nomenclature[C]{$b_{sy_1}$}{Origin buffer $y_1$ in set of buffers $B_S$}
\nomenclature[C]{$b_{sy_2}$}{Destination buffer $y_2$ in set of buffers $B_S$}
\nomenclature[D]{$u$}{Index of transportation process $p_{\eta u}$ in set of transportation processes $P_\eta$}
\nomenclature[D]{$y_1$}{Index of origin buffer $b_{sy_1}$ in set of buffers $B_S$}
\nomenclature[D]{$y_2$}{Index of destination buffer $b_{sy_2}$ in set of buffers $B_S$}
\nomenclature[F]{$\sigma()$}{The size of the set ().}
\nomenclature[B]{$P_{\bar{\eta}}$}{Set of Refined Transportation Processes}
\nomenclature[B]{$P_{\gamma}$}{Set of Holding Processes}
\nomenclature[C]{$p_{\gamma g}$}{Holding process $g$ in set of holding processes $P_{\gamma}$}
\nomenclature[D]{$g$}{Index of holding process $p_{\gamma g}$ in set of holding processes $P_{\gamma}$}
\nomenclature[C]{$p_{\bar{\eta} \varphi}$}{Refined transportation process $\varphi$ in set of refined transportation processes $P_{\bar{\eta}}$}
\nomenclature[D]{$\varphi$}{Index of refined transportation process $p_{\bar{\eta} \varphi}$ in $P_{\bar{\eta}}$}
\nomenclature[E]{$J_S$}{System Knowledge Base}
\nomenclature[D]{$w$}{Index of system process $p_w$ in set of system processes $P$}
\nomenclature[D]{$v$}{Index of physical resource $p_w$ in set of physical resources $R$}
\nomenclature[B]{${\cal E}$}{Set of System Actions}
\nomenclature[C]{$e_{wv}$}{System action $w,v$ in set of system actions ${\cal E}$}
\nomenclature[F]{$\odot$}{Matrix boolean multiplication}
\nomenclature[E]{$J_M$}{Transformation Knowledge Base}
\nomenclature[E]{$J_H$}{Transportation Knowledge Base}
\nomenclature[E]{$J_\gamma$}{Holding Knowledge Base}
\nomenclature[E]{$J_{\bar{H}}$}{Refined Transportation Knowledge Base}
\nomenclature[F]{$\otimes$}{Kronecker Product}
\nomenclature[E]{$\mathds{1}^n$}{Ones-vector of length n}
\nomenclature[E]{$K_S$}{System Constraints Matrix}
\nomenclature[E]{$K_M$}{Transformation Constraints Matrix}
\nomenclature[E]{$K_{\bar{H}}$}{Refined Transportation Constraints Matrix}
\nomenclature[E]{$DOF_S$}{Structural Degrees of Freedom}
\nomenclature[B]{${\cal E}_S$}{Set of Structural Degrees of Freedom}
\nomenclature[E]{$A_S$}{System Concept}
\nomenclature[F]{$\langle A , B \rangle_F$}{Frobenius Product of matrices $A$ and $B$}
\nomenclature[E]{$DOF_M$}{Transformation Degrees of Freedom}
\nomenclature[E]{$DOF_H$}{Transportation Degrees of Freedom}
\nomenclature[D]{$\chi$}{Index in $\left[1\dots \sigma(R)\sigma(P)\right]$}
\nomenclature[F]{$()^V$}{Shorthand for vectorization (i.e. vec())}
\nomenclature[E]{$A_{\rho}$}{Hetero-functional Adjacency Matrix}
\nomenclature[E]{$J_{\rho}$}{System Sequence Knowledge Base}
\nomenclature[E]{$K_\rho$}{System Sequence Constraints Matrix}
\nomenclature[B]{${\cal Z}$}{Set of System Activity Strings}
\nomenclature[C]{$z_{\chi_1, \chi_2}$}{String of two sequential activities $\chi_1$ and $\chi_2$ in set of strings ${\cal Z}$}

\subsection{Ontological Science Definitions}\label{sec:ontology}
The formal definitions of soundness, completeness, lucidity, and laconicity rely on ``Ullman's Triangle" in Figure \ref{fig:UllmansTriangle}.  
\liinesbigfig{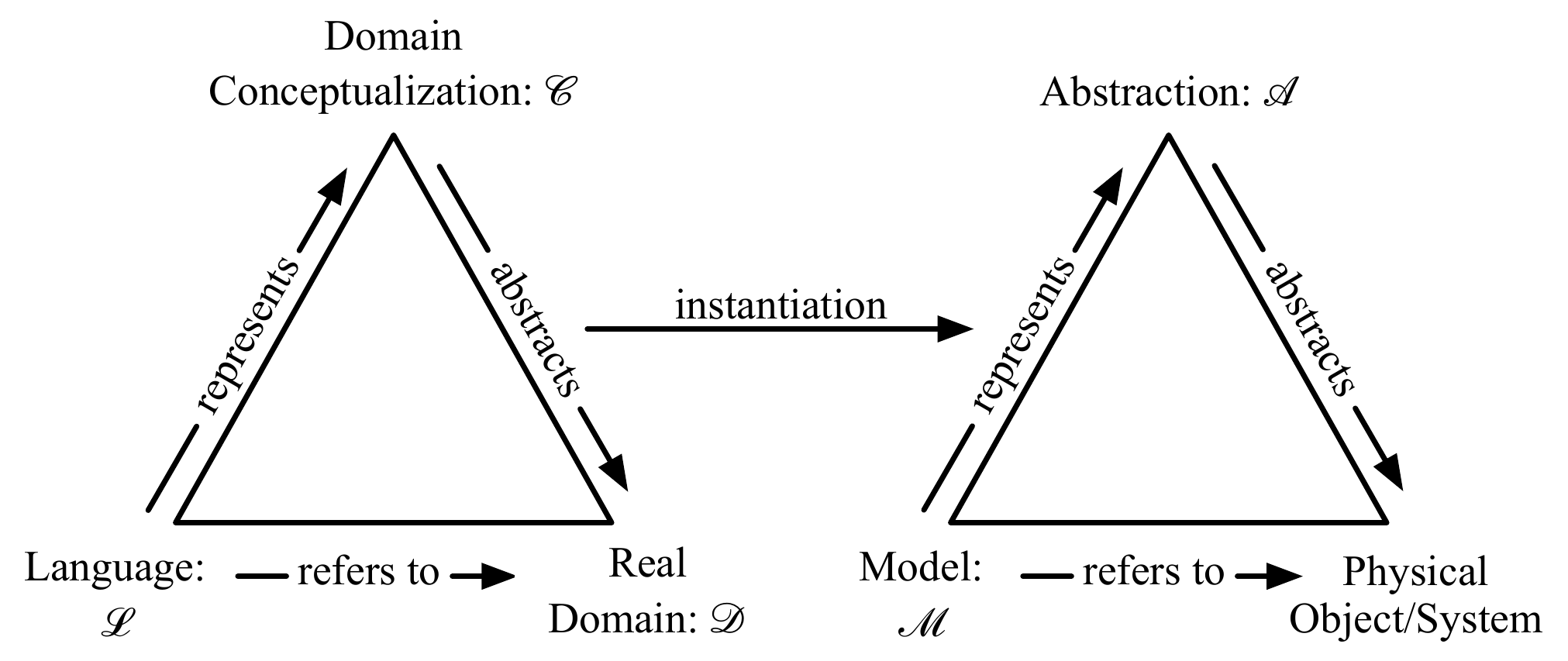}{Two Versions of Ullman's Triangle. On the left is the relationship between reality, the understanding of reality, and the description of reality.  On the right the instantiated version of the ontological definition\cite{Guizzardi:2007:00}.}{fig:UllmansTriangle}
\begin{defn}[Soundness\cite{Guizzardi:2005:00}]\label{defn:soundness}
A language ${\cal L}$ is sound w.r.t. a domain conceptualization ${\cal C}$ iff every modeling primitive in the language (${\cal M}$) has an interpretation in the domain abstraction ${\cal A}$. (The absence of soundness results in the excess of modeling primitives w.r.t. the domain abstractions as shown in Figure \ref{fig:Ontology}{\color{blue}.c} on lucidity.)
\end{defn}
\begin{defn}[Completeness\cite{Guizzardi:2005:00}]\label{defn:completeness}
A language ${\cal L}$ is complete w.r.t. a domain conceptualization ${\cal C}$ iff every concept in the domain abstraction ${\cal A}$ of that domain is represented in a modeling primitive of that language. (The absence of completeness results in one or more concepts in the domain abstraction not being represented by a modeling primitive, as shown in Figure \ref{fig:Ontology}{\color{blue}.d} on laconicity.)
\end{defn}
\begin{defn}[Lucidity\cite{Guizzardi:2005:00}]\label{defn:lucidity}
A language ${\cal L}$ is lucid w.r.t. a domain conceptualization ${\cal C}$ iff every modeling primitive in the language represents at most one domain concept in abstraction ${\cal A}$. (The absence of lucidity results in the overload of a modeling primitive w.r.t. two or more domain concepts as shown in Figure \ref{fig:Ontology}{\color{blue}.a} on soundness.)
\end{defn}
\begin{defn}[Laconicity\cite{Guizzardi:2005:00}]\label{defn:laconicity}
    A language ${\cal L}$ is laconic w.r.t. a domain conceptualization ${\cal C}$ iff every concept in the abstraction ${\cal A}$ of that domain is represented at most once in the model of that language. (The absence of laconicity results in the redundancy of modeling primitives w.r.t the domain abstractions as shown in Figure \ref{fig:Ontology}{\color{blue}.b} on completeness.)
\end{defn}

\subsection{Notation Conventions}\label{app:notations}
\noindent Several notation conventions are used throughout this work:
\begin{itemize}
\item All sets are indicated by a capital letter.  e.g. P -- the set of processes..  
\item All elements within a set are indicated by a lower case letter.  e.g. $p \in P$.
\item A subscript number indicates the position in an ordered set.  e.g. $p_i \in P$.
\item The $i^{th}$ elementary basis vector of size n is denoted by $e_i^n$.  
\item A matrix of ones of size $m\times n$ is denoted by $\mathds{1}^{m\times n}$.  
\item A matrix of zeros of size $m\times n$ is denoted by $\mathbf{0}^{m\times n}$.  
\item With the exception of elementary basis vectors, all vectors and matrices are indicated with a capital letter.  e.g. $J_H$.
\item All tensors are indicated with capital letters in calligraphic script.  e.g. $\cal{J}_H$.
\item All elements in vectors, matrices, and tensors are indicated with indices within parentheses.  e.g. $J_S(w,v)$.
\item A(:,i) denotes the $i^{th}$ column of A or equivalently the $i^{th}$ mode-1 fiber.  The : indicates all elements of the vector.  
\item A(i,:) denotes the $i^{th}$ row of A or equivalently the $i^{th}$ mode-2 fiber.  
\item {\cal A}(i,j,:) denotes the $i,j$ mode-3 fiber of $\cal A$.  
\item Given the presence of Booleans, real numbers and their operators, this work refrains from the use of Einstein's (shorthand) tensor notation where the sigma-notation $\sum$ is eliminated.   
\end{itemize}

\subsection{Hetero-functional Graph Theory Definitions}\label{sec:hfgt_dfn}

\begin{defn}[Transformation Resource\cite{Schoonenberg:2019:ISC-BK04}]\label{defn:M}
    A resource $r \in R$ is a transformation resource $m \in M$ iff it is capable of one or more transformation processes on one or more operands and it exists at a unique location in space. 
\end{defn}

\begin{defn}[Independent Buffer\cite{Schoonenberg:2019:ISC-BK04}]\label{defn:B}
    A resource $r \in R$ is an independent buffer $b \in B$ iff it is capable of storing one or more operands and is not able to transform them or transport them to another location and it exists at a unique location in space.
\end{defn}

\begin{defn}[Transportation Resource\cite{Schoonenberg:2019:ISC-BK04}]\label{defn:H}
    A resource $r \in R$ is a transportation resource $h \in H$ iff it is capable of transporting one or more operands between an origin and a distinct destination, without transforming these operands. 
\end{defn}

\begin{defn}[Buffer\cite{Schoonenberg:2019:ISC-BK04}]\label{defn:BS}
    A resource $r \in R$ is a buffer $b_s \in B_S$ iff it is capable of storing one or more operands at a unique location in space.  $B_S=M \cup B$. 
\end{defn}

\begin{defn}[Transformation Process\cite{Schoonenberg:2019:ISC-BK04}]\label{defn:Pmu}
    A process is a transformation process $p_{\mu j} \in P_{\mu}$ iff it is capable of transforming one or more properties of a set of operands into a distinct set of output properties in place. It's syntax is:
\begin{equation}
    \{\mbox{transitive verb, operands}\}\rightarrow\{\mbox{outputs}\}
\end{equation}
\end{defn}

\begin{defn}[Refined Transportation Process\cite{Schoonenberg:2019:ISC-BK04}]\label{defn:PHref}
    A process is a refined transportation process $p_{\bar{\eta} \varphi} \in P_{\bar{\eta}}$ iff it is capable of transporting one or more operands between an origin buffer $b_{sy_1} \in B_S$ to a destination buffer $b_{sy_2} \in B_S$ while it is realizing holding process $p_{\gamma g} \in P_\gamma$. 
    It's syntax is:  
    \begin{align}\nonumber
        \{\mbox{transport, operands, origin, destination,while transitive verb}\}\rightarrow \\  \{\mbox{outputs, destination}\}
    \end{align} 
\end{defn}

\begin{defn}[Transportation Process\cite{Schoonenberg:2019:ISC-BK04}]\label{defn:PH}
    A process is a transportation process $p_{\eta u} \in P_{\eta}$ iff it is capable of transporting one or more operands between an origin buffer $b_{sy_1} \in B_S$ to a destination buffer $b_{sy_2} \in B_S$ according to the following convention of indices\cite{Farid:2006:IEM-C02,Farid:2007:IEM-TP00,Farid:2008:IEM-J05,Farid:2008:IEM-J04,Farid:2015:ISC-J19}\footnote{Note that a ``storage process" is merely a transportation process with the same origin and destination.}: 
    \begin{equation}\label{eq:indexingtransport}
    	u = \sigma(B_S)(y_1 -1) + y_2
    \end{equation}
    It's syntax is:  
\begin{align}
    \{\mbox{transport, operands,} & \mbox{origin,destination}\}\rightarrow\{\mbox{outputs,destination}\}
\end{align}   
\end{defn}

\begin{defn}[Holding Process\cite{Schoonenberg:2019:ISC-BK04}]\label{defn:Pgamma}
    A process is a holding process $p_{\gamma g} \in P_\gamma$ iff it holds one or more operands during the transportation from one buffer to another.  In order to maintain the independence axiom and the mutual exclusivity of the system processes (Theorem \ref{thm:Ch4:MutualExclusivity}), holding processes are specified so as to distinguish between transportation processes that: 
\begin{itemize}
    \item Have \emph{different} operands,
    \item Hold a given operand in a \emph{given way}, or
    \item Change the \emph{state} of the operand.
\end{itemize}
\end{defn}

\begin{thm}[Mutual Exclusivity of System Processes\cite{Schoonenberg:2019:ISC-BK04}]\label{thm:Ch4:MutualExclusivity}
    A lucid representation of system processes as a domain conceptualization distinguishes between two system processes as modeling primitives with different sets of inputs and outputs.  
\end{thm}

\begin{defn}[Capability\cite{Farid:2006:IEM-C02,Farid:2007:IEM-TP00,Farid:2008:IEM-J05,Farid:2008:IEM-J04,Farid:2015:ISC-J19,Farid:2016:ISC-BC06}]\label{defn:capability}
An action $e_{wv} \in {\cal E}_S$ (in the SysML sense) defined by a system process $p_w \in P$ being executed by a resource $r_v \in R$.  It constitutes a subject + verb + operand sentence of the form: ``Resource $r_v$ does process $p_w$".  
\end{defn} 

\begin{defn}[The Negative Transformation Process-Operand Incidence Matrix $M_{L P_\mu}^-$]\label{def:negxFormProcOper}
A binary incidence matrix $M_{L P_\mu}^{-} \in \{0,1\}^{\sigma(L)\times \sigma(P_\mu)}$ whose element $M_{L P}^{-}(i,j)=1$ when the transformation system process $p_{\mu_j} \in P$ pulls operand $l_i \in L$ as an input.

\end{defn} 

\begin{defn}[The Negative Refined Transportation Process-Operand Incidence Matrix $M_{LP_{\bar{\eta}}}^-$]\label{def:negxPortRefProcOper}
A binary incidence matrix $M_{L P_{\bar{\eta}}}^{-} \in \{0,1\}^{\sigma(L)\times \sigma(P_{\bar{\eta}})}$ whose element $M_{L P}^{-}(i,\varphi)=1$ when the refined transportation process $p_\varphi \in P_{\bar{\eta}}$ pulls operand $l_i \in L$ as an input.  It is calculated directly from the negative holding process-operand incidence matrix $M_{LP_{\gamma}}^-$.  
\begin{alignat}{2}
M_{LP_{\bar{\eta}}}^-(i,\varphi)
&=\sum_{u=1}^{\sigma(P_\eta)} 
M_{LP_{\gamma}}^-(i,g) \cdot \delta_u \qquad 
\forall i \in \{1,\ldots \sigma(L)\}, \;
g \in \{1, \ldots, \sigma(P_\gamma)\},
u \in \{1, \ldots, \sigma(P_\eta)\},
\varphi=\sigma(P_\eta)(g-1)+u
\\
M_{LP_{\bar{\eta}}}^-(i,\varphi)
&= \sum_{y_1=1}^{\sigma(B_S)}
\sum_{y_2=1}^{\sigma(B_S)} 
M_{LP_{\gamma}}^-(i,g) \cdot \delta_{y_1}\cdot \delta_{y_2}
\quad 
\forall y_1,y_2 \in \{1, \ldots, \sigma(B_S)\},
\varphi=\sigma^2(B_S)(g-1)+\sigma(B_S)(y_1-1)+y_2 \\
M_{LP_{\bar{\eta}}}^-&= 
\sum_{u=1}^{\sigma(P_\eta)}
M_{LP_{\gamma}}^- \otimes e_u^{\sigma(P_\eta)T} \quad\;\;
=\sum_{y_1=1}^{\sigma(B_S)}
\sum_{y_2=1}^{\sigma(B_S)} 
M_{LP_{\gamma}}^- \otimes \left(e_{y_1}^{\sigma(B_S)}  \otimes e_{y_2}^{\sigma(B_S)}\right)^T\\
&=M_{LP_{\gamma}}^- \otimes \mathds{1}^{\sigma(P_\eta)T} \qquad\quad 
=M_{LP_{\gamma}}^- \otimes \left(\mathds{1}^{\sigma(B_S)}\otimes \mathds{1}^{\sigma(B_S)}\right)^T
\end{alignat}
\end{defn} 


\begin{defn}[The Negative Holding Process-Operand Incidence Matrix $M_{LP_{\gamma}}^-$]\label{defn:negHoldProcOper}
A binary incidence matrix $M_{LP_{\gamma}}^- \in \{0,1\}^{\sigma(L)\times \sigma(P_{\gamma})}$ whose element $M_{L P}^{-}(i,g)=1$ when the holding process $p_g \in P_{\gamma}$ pulls operand $l_i \in L$ as an input.  

\end{defn} 

\begin{defn}[The Positive Transformation Process-Operand Incidence Matrix $M_{L P_\mu}^+$]\label{def:posxFormProcOper}
A binary incidence matrix $M_{L P_\mu}^{+} \in \{0,1\}^{\sigma(L)\times \sigma(P_\mu)}$ whose element $M_{L P}^{+}(i,j)=1$ when the transformation system process $p_{\mu_j} \in P$ ejects operand $l_i \in L$ as an output.

\end{defn} 

\begin{defn}[The Positive Refined Transportation Process-Operand Incidence Matrix $M_{LP_{\bar{\eta}}}^+$]\label{def:posxPortRefProcOper}
A binary incidence matrix $M_{L P_{\bar{\eta}}}^{+} \in \{0,1\}^{\sigma(L)\times \sigma(P_{\bar{\eta}})}$ whose element $M_{L P}^{+}(i,\varphi)=1$ when the refined transportation process $p_\varphi \in P_{\bar{\eta}}$ ejects operand $l_i \in L$ as an output.  It is calculated directly from the negative holding process-operand incidence matrix $M_{LP_{\gamma}}^+$.
\begin{alignat}{2}
M_{LP_{\bar{\eta}}}^+(i,\varphi)
&=\sum_{u=1}^{\sigma(P_\eta)} 
M_{LP_{\gamma}}^+(i,g) \cdot \delta_u \qquad 
\forall i \in \{1,\ldots \sigma(L)\}, \;
g \in \{1, \ldots, \sigma(P_\gamma)\},
u \in \{1, \ldots, \sigma(P_\eta)\},
\varphi=\sigma(P_\eta)(g-1)+u
\\
M_{LP_{\bar{\eta}}}^+(i,\varphi)
&= \sum_{y_1=1}^{\sigma(B_S)}
\sum_{y_2=1}^{\sigma(B_S)} 
M_{LP_{\gamma}}^+(i,g) \cdot \delta_{y_1}\cdot \delta_{y_2}
\quad 
\forall y_1,y_2 \in \{1, \ldots, \sigma(B_S)\},
\varphi=\sigma^2(B_S)(g-1)+\sigma(B_S)(y_1-1)+y_2 \\
M_{LP_{\bar{\eta}}}^+&= 
\sum_{u=1}^{\sigma(P_\eta)}
M_{LP_{\gamma}}^+ \otimes e_u^{\sigma(P_\eta)T} \quad\;\;
=\sum_{y_1=1}^{\sigma(B_S)}
\sum_{y_2=1}^{\sigma(B_S)} 
M_{LP_{\gamma}}^+ \otimes \left(e_{y_1}^{\sigma(B_S)}  \otimes e_{y_2}^{\sigma(B_S)}\right)^T\\
&=M_{LP_{\gamma}}^+ \otimes \mathds{1}^{\sigma(P_\eta)T} \qquad\quad 
=M_{LP_{\gamma}}^+ \otimes \left(\mathds{1}^{\sigma(B_S)}\otimes \mathds{1}^{\sigma(B_S)}\right)^T
\end{alignat}
\end{defn} 

\begin{defn}[The Positive Holding Process-Operand Incidence Matrix $M_{LP_{\gamma}}^+$]\label{defn:posHoldProcOper}
A binary incidence matrix $M_{LP_{\gamma}}^+ \in \{0,1\}^{\sigma(L)\times \sigma(P_{\gamma})}$ whose element $M_{L P}^{+}(i,g)=1$ when the holding process $p_g \in P_{\gamma}$ ejects operand $l_i \in L$ as an output.  

\end{defn}

\subsection{Definitions of Set Operations}\label{sec:setDefn}
\begin{defn}[$\sigma$\mbox{()} Notation \cite{Schoonenberg:2019:ISC-BK04}]\label{defn:sigma}
returns the size of the set.  Given a set S with n elements, $n=\sigma(S)$.  
\end{defn}

\begin{defn}[Cartesian Product \Cross \cite{Anonymous:2021:00}]\label{defn:CartesProd}
Given three sets, A, B, and C, 
\begin{equation}
A \mbox{\Cross} B = \{(a,b)\in C  \quad \forall  a \in A \;\mbox{and}\; b \in B\}
\end{equation}
\end{defn}

\subsection{Definitions of Boolean Operations}\label{sec:boolDefn}
\noindent The conventional symbols of $\wedge$, $\vee$, and $\lnot$ are used to indicate the AND, OR, and NOT operations respectively.  
\begin{defn}[$\bigvee$ Notation]\label{defn:bigvee}
$\bigvee$ notation indicates a Boolean OR over multiple binary elements $a_i$.  
\begin{equation}
\bigvee_i^n a_i = a_1 \vee a_2 \vee \ldots \vee a_n
\end{equation}  
\end{defn}

\begin{defn}[Matrix Boolean Addition $\oplus$] \label{defn:boolAdd}
Given Boolean matrices $A,B,C \in \{0,1\}^{m\times n}$, $C=A\oplus B$ is equivalent to
\begin{equation}
C(i,j)=A(i,j) \vee B(i,j) \qquad \forall i\in\{1\ldots m\}, j\in\{1\ldots n\}
\end{equation}
\end{defn}

\begin{defn}[Matrix Boolean Scalar Multiplication $(\cdot$)]\label{defn:boolScalMult}
Given Boolean matrices $A,B,C \in \{0,1\}^{m\times n}$, $C=A\cdot B$ is equivalent to
\begin{equation}
C(i,j) = A(i,j) \wedge B(i,j) = A(i,j) \cdot B(i,j) \qquad \forall i\in\{1\ldots m\}, j\in\{1\ldots n\}
\end{equation}
\end{defn}

\begin{defn}[Matrix Boolean Multiplication $\odot$\cite{Farid:2007:IEM-TP00,Fischer:1971:00}]\label{defn:boolMult}
Given matrices $A\in \{0,1\}^{m\times n}$, $B \in \{0,1\}^{n\times p}$, and $C \in \{0,1\}^{m\times p}$, $C=A \times B=AB$ is equivalent to
\begin{equation}
C(i,k)=\bigvee_{i=1}^{n} A(i,j) \wedge B(j,k) = \bigvee_{i=1}^{n} A(i,j) \cdot B(j,k)  \qquad \forall i\in\{1\ldots m\}, k\in\{1\ldots p\}
\end{equation}
\end{defn}

\begin{defn}[Matrix Boolean Subtraction]\label{defn:boolSub}
Given Boolean matrices $A,B,C \in \{0,1\}^{m\times n}$, $C=A\ominus B$ is equivalent to
\begin{equation}
C(i,k)=A(i,j) \wedge \lnot B(i,j)) = A(i,j) \cdot \lnot B(i,j) \qquad \forall i\in\{1\ldots m\}, j\in\{1\ldots n\}
\end{equation}
\end{defn}

\subsection{Matrix Operations}\label{sec:matDefn}

\begin{defn}[Kronecker Delta Function $\delta_{ij}$\cite{Anonymous:2021:01}]\label{defn:kronDelta}
\begin{equation}
\delta_{ij}= 
\left\{
\begin{array}{cc}
1 & \mbox{if} \; i=j\\
0 & \mbox{if} \; i\neq j
\end{array}
\right.
\end{equation}
\end{defn}

\begin{defn}[Hadamard Product\cite{Golub:1996:00}]\label{defn:hadamard}
Given matrices $A,B,C \in \mathds{R}^{m\times n}$, $C=A \cdot B$ is equivalent to
\begin{equation}
C(i,j)=A(i,j) \cdot B(i,j)  \qquad \forall i\in\{1\ldots m\}, j\in\{1\ldots n\}
\end{equation}
\end{defn}

\begin{defn}[Matrix Product \cite{Golub:1996:00}]
Given matrices $A\in \mathds{R}^{m\times n}$, $B \in \mathds{R}^{n\times p}$, and $C \in \mathds{R}^{m\times p}$, $C=A \times B=AB$ is equivalent to
\begin{equation}
C(i,k)=\sum_{i=1}^{n} A(i,j) \cdot B(j,k)  \qquad \forall i\in\{1\ldots m\}, k\in\{1\ldots p\}
\end{equation}
\end{defn}

\begin{defn}[Kronecker Product \cite{Kolda:2009:00,Kolda:2006:00}]\label{defn:kron}
Given matrix $A \ \in \mathds{R}^{m\times n}$ and $B \ \in \mathds{R}^{p\times q}$, the Kronecker (kron) product denoted  by $C=A\otimes B$ is given by:
\begin{equation}
C=
\begin{bmatrix}
A(1,1)B & A(1,2)B & \ldots & A(1,n)B\\
A(2,1)B & A(2,2)B & \ldots & A(2,n)B\\
\vdots & \vdots & \ddots & \vdots \\
A(m,1)B & A(m,2)B & \ldots & A(m,n)B
\end{bmatrix}
\end{equation}
Alternatively, in scalar notation:
\begin{equation}
C(p(i-1)+k,q(j-1)+l)=a(i,j)\cdot b(k,l) \qquad \forall  i\in\{1\ldots m\}, j\in\{1\ldots n\}, k\in\{1\ldots p\} , l\in\{1\ldots q\}
\end{equation}
\end{defn}

\begin{defn}[Khatri-Rao Product \cite{Kolda:2009:00,Kolda:2006:00}]\label{defn:khatri-rao}
The Khatri-Rao Product is the ``column-wise Kronecker product".  Given matrix $A \ \in \mathds{R}^{m\times n}$ and $B \ \in \mathds{R}^{p\times n}$, the Khatri-Rao product denoted  by $C=A\circledast B$ is given by:
\begin{align}
C&=
\begin{bmatrix}
A(:,1)\otimes B(:,1) & A(:,2)\otimes B(:,2) & \ldots & A(:,n)\otimes B(:,n)\\
\end{bmatrix}\\
&=\left[
A\otimes \mathds{1}^{p}
\right]
\cdot
\left[
\mathds{1}^{m}
\otimes
B
\right]
\end{align}
Alternatively, in scalar notation:
\begin{equation}
C(p(i-1)+k,j)=a(i,j)\cdot b(k,j) \qquad \forall  i\in\{1\ldots m\}, j\in\{1\ldots n\}, k\in\{1\ldots p\} 
\end{equation}
If A and B are column vectors, the Kronecker and Khatri-Rao products are identical.  
\end{defn}

\subsection{Tensor Operations}\label{sec:tensorDefn}

\begin{defn}[Outer Product of Vectors \cite{Kolda:2009:00,Kolda:2006:00}]\label{defn:outerProduct}
Given two vectors $A_1 \in \mathds{R}^{m_1}$ and $A_2 \in \mathds{R}^{m_2}$, their outer product $B \in \mathds{R}^{m_1\times m_2}$ is denoted by 
\begin{align}
B=A_1 \circ A_2&=A_1A_2^T\\
B(i_1,i_2)&=A_1(i_1)\cdot A_2(i_2) \qquad \forall  i_1\in\{1\ldots m_1\}, i_2\in\{1\ldots m_2\} 
\end{align}
Given $n$ vectors,  $A_1 \in \mathds{R}^{m_1},  A_2 \in \mathds{R}^{m_2}, \ldots, A_n \in \mathds{R}^{m_n}$, their outer product ${\cal B} \in \mathds{R}^{m_1\times m_2 \times \ldots \times m_n}$ is denoted by ${\cal B}=A_1 \circ A_2 \circ \ldots  \circ A_n$ where 
\begin{align}
{\cal B}(i_1,i_2,\ldots,i_n)=A_1(i_1)\cdot A_2(i_2)\cdot\ldots\cdot A_n(i_n) \quad \forall  i_1\in\{1\ldots m_1\}, i_2\in\{1\ldots m_2\}, \ldots, i_n=\{1\ldots m_n\}
\end{align}
\end{defn}

\begin{defn}[Matricization ${\cal F}_M()$ \cite{Kolda:2009:00,Kolda:2006:00}]\label{defn:matricization}
Given an n$^{th}$ order tensor ${\cal A} \in \mathds{R}^{p_1\times p_2 \times \ldots \times p_n}$, and ordered sets $R=\{r_1,\ldots, r_L\}$ and $C=\{c_1,\ldots, c_M\}$ that are a partition of the n modes $N=\{1,\ldots, n\}$ (i.e. $R\cup C=N, R\cap C=\emptyset$), the matricization function ${\cal F}_M()$ outputs the matrix $A \in \mathds{R}^{J\times K}$
\begin{align}
A&={\cal F}_M({\cal A},R,C)\\
A(j,k)&={\cal A}(i_1,i_2,\ldots, i_n) \qquad \forall i_1\in \{1,\ldots, p_1\}, i_2\in \{1,\ldots, p_2\}, \ldots i_n\in \{1,\ldots, p_n\}
\end{align}
where 
\begin{alignat}{4}
j=1+\sum_{l=1}^L\left[(i_{r_l}-1)\prod_{l'=1}^{l-1} i_{r_{l'}}\right], \quad
&&k=1+\sum_{m=1}^M\left[(i_{c_m}-1)\prod_{m'=1}^{m-1} i_{c_{m'}}\right], \quad
&& J=\prod_{q\in R}p_q \quad 
&& K=\prod_{q\in C}p_q
\end{alignat}
For the sake of clarity, ${\cal F}_M()$ is implemented in MATLAB code:
\begin{lstlisting}
ATensor = rand(4,7,5,3); R = [4 1]; C = [2 3];
function AMatrix=matricize(ATensor,R,C);
P = size(ATensor); J = prod(P(R)); K = prod(P(C));
AMatrix = reshape(permute(ATensor,[R C]),J,K);  % Matricize 
\end{lstlisting}
\end{defn}

\begin{defn}[Tensorization \cite{Kolda:2009:00,Kolda:2006:00}]\label{defn:tensorization}
Given a matrix $A \in \mathds{R}^{J\times K}$, the dimensions $P=[p_1, p_2, \ldots, p_n]$ of a target 
n$^{th}$ order tensor ${\cal A} \in \mathds{R}^{p_1\times p_2 \times \ldots \times p_n}$, 
 and ordered sets $R=\{r_1,\ldots, r_L\}$ and $C=\{c_1,\ldots, c_M\}$ that are a partition of the n modes $N=\{1,\ldots, n\}$ (i.e. $R\cup C=N, R\cap C=\emptyset$), the tensorization function ${\cal F}_M^{-1}()$ outputs the n$^{th}$ order tensor ${\cal A}$.  
 \begin{align}
{\cal A}&={\cal F}_M^{-1}(A,P, R,C)\\
{\cal A}(i_1,i_2,\ldots, i_n)&=A(j,k) \qquad \forall i_1\in \{1,\ldots, p_1\}, i_2\in \{1,\ldots, p_2\}, \ldots i_n\in \{1,\ldots, p_n\}
\end{align}
where 
\begin{alignat}{4}
j=1+\sum_{l=1}^L\left[(i_{r_l}-1)\prod_{l'=1}^{l-1} i_{r_{l'}}\right], \quad
&&k=1+\sum_{m=1}^M\left[(i_{c_m}-1)\prod_{m'=1}^{m-1} i_{c_{m'}}\right], \quad
&& J=\prod_{q\in R}p_q \quad 
&& K=\prod_{q\in C}p_q
\end{alignat}
For the sake of clarity, ${\cal F}_M^{-1}()$ is implemented in MATLAB code:
\begin{lstlisting}
AMatrix = rand(12,35); P=[4,7,5,3]; R = [4 1]; C = [2 3];
function ATensor=tensorize(AMatrix,P,R,C);
ATensor = ipermute(reshape(AMatrix,[P(R) P(C)]),[R C]); % Tensorize
\end{lstlisting}
\end{defn}

\begin{defn}[Vectorization \cite{Golub:1996:00,Kolda:2009:00,Kolda:2006:00}]\label{defn:vec}
Vectorization denoted by $vec()$ or $()^V$ as a shorthand is a special case of matricization when the resulting matrix is simply a vector.   Formally, given an n$^{th}$ order tensor ${\cal A} \in \mathds{R}^{p_1\times p_2 \times \ldots \times p_n}$ and the dimensions $P=[p_1, p_2, \ldots, p_n]$, the vectorization function $vec()=()^V$ outputs the vector $A \in \mathds{R}^{J}$
\begin{align}
A&=vec({\cal A})={\cal A}^V\\
A(j)&={\cal A}(i_1,i_2,\ldots, i_n) \qquad \forall i_1\in \{1,\ldots, p_1\}, i_2\in \{1,\ldots, p_2\}, \ldots i_n\in \{1,\ldots, p_n\}
\end{align}
where 
\begin{alignat}{4}
j=1+\sum_{l=1}^n\left[(i_{l}-1)\prod_{l'=1}^{l-1} i_{l'}\right], \quad
&& J=\prod_{q=1}^n p_q \quad 
\end{alignat}
\end{defn}

\begin{defn}[Inverse Vectorization \cite{Golub:1996:00,Kolda:2009:00,Kolda:2006:00}]\label{defn:vec-1}
Inverse vectorization denoted by $vec^{-1}()$ is a special case of tensorization when the input matrix is simply a vector.  Formally, $A \in \mathds{R}^{J}$ and the dimensions $P=[p_1, p_2, \ldots, p_n]$ of a target 
n$^{th}$ order tensor ${\cal A} \in \mathds{R}^{p_1\times p_2 \times \ldots \times p_n}$, the inverse vectorization function $vec^{-1}()$ outputs the n$^{th}$ order tensor ${\cal A}$.  
 \begin{align}
 {\cal A}&=vec^{-1}(A,P)\\
{\cal A}(i_1,i_2,\ldots, i_n)&=A(j) \qquad \forall i_1\in \{1,\ldots, p_1\}, i_2\in \{1,\ldots, p_2\}, \ldots i_n\in \{1,\ldots, p_n\}
\end{align}
where 
\begin{alignat}{4}\label{Eq:VecIndices}
j=1+\sum_{l=1}^n\left[(i_{l}-1)\prod_{l'=1}^{l-1} i_{l'}\right], \quad
&& J=\prod_{q=1}^n p_q \quad 
\end{alignat}
Furthermore, the above definition of inverse vectorization can be applied to a q$^{th}$ dimensional slice of a tensor.  In such a case, 
\begin{align}
 {\cal B}&=vec^{-1}(A,P,r)\\
{\cal B}(k_1,\ldots, k_{r-1},i_1,\ldots,i_n,k_{r+1},\ldots, k_m)&={\cal A}(k_1,\ldots, k_{r-1},j, k_{r+1},\ldots, k_m)
\end{align}
where index convention in Equation \ref{Eq:VecIndices} applies.  
\end{defn}

\begin{defn}[Matrix and Tensor Transpose]\label{defn:transpose}
Given a matrix $A \in \mathds{R}^{{m_1}\times {m_2}}$, its matrix transpose $A^T \in \mathds{R}^{{m_2}\times {m_1}}$ is equivalent to:
\begin{equation}
A^T(j,i)=A(i,j) \qquad \forall i \in \{1\ldots m_1\}, j \in \{1\ldots m_2\}
\end{equation}
In this work, the generalization to tensors is a special case of the definition provided in \cite{Pan:2014:00}.  Given a tensor  ${\cal A} \in \mathds{R}^{{m_1}\times \ldots \times {m_n}}$, its tensor transpose ${\cal A}^T \in \mathds{R}^{{m_n}\times \ldots \times  {m_1}}$ is equivalent to:
\begin{equation}
{\cal A}^T(i_n,\ldots,i_1)={\cal A}(i_1,\ldots i_n) \qquad \forall i_1 \in \{1\ldots m_1\}, \ldots, i_n \in \{1\ldots m_n\}
\end{equation}
\end{defn}

\begin{defn}[N-Mode Matrix Product $\times_p$ \cite{Kolda:2009:00,Kolda:2006:00}]\label{defn:nModeProduct}
The N-mode matrix product is a generalization of the matrix product.   Given a tensor ${\cal A}\in \mathds{R}^{m_1\times m_2 \times \ldots \times m_p \times \ldots \times m_n}$, matrix $B \in \mathds{R}^{m_p \times q}$,  and ${\cal C}\in \mathds{R}^{m_1\times m_2 \times \ldots \times q \times \ldots \times m_n}$, the n-mode matrix product denoted by ${\cal C}={\cal A} \times_p B$ is equivalent to:
\begin{align}
&{\cal C}(i_1,i_2,\ldots, i_{p-1},j,i_{p+1},\ldots, i_n)=
\sum_{i_p=1}^{m_p} 
{\cal A}(i_1,i_2,\ldots, i_n)\cdot B(i_p,j) \\\nonumber
&\forall i_1\in \{1,\ldots, m_1\},  \ldots, i_{p-1}\in \{1,\ldots, m_{p-1}\}, i_{p+1}\in \{1,\ldots, m_{p+1}\} ,\ldots,  i_n\in \{1,\ldots, m_n\}, j \in \{1,\ldots,q\}
\end{align}
\end{defn}

\begin{defn}[N-Mode Boolean Matrix Product]\label{defn:nBoolModeProduct}
The N-mode Boolean matrix product is a generalization of the Boolean matrix product.   Given a tensor ${\cal A}\in \{0,1\}^{m_1\times m_2 \times \ldots \times m_p \times \ldots \times m_n}$, matrix $B \in \{0,1\}^{m_p \times q}$,  and ${\cal C}\in \{0,1\}^{m_1\times m_2 \times \ldots \times q \times \ldots \times m_n}$, the n-mode matrix product denoted by ${\cal C}={\cal A} \odot_p B$ is equivalent to:
\begin{align}
&{\cal C}(i_1,i_2,\ldots, i_{p-1},j,i_{p+1},\ldots, i_n)=
\bigvee_{i_p=1}^{m_p} 
{\cal A}(i_1,i_2,\ldots, i_n)\cdot B(i_p,j) \\\nonumber
&\forall i_1\in \{1,\ldots, m_1\},  \ldots, i_{p-1}\in \{1,\ldots, m_{p-1}\}, i_{p+1}\in \{1,\ldots, m_{p+1}\} ,\ldots,  i_n\in \{1,\ldots, m_n\}, j \in \{1,\ldots,q\}
\end{align}
\end{defn}

\end{document}